\documentclass[sigconf]{acmart}
\AtBeginDocument{%
  }

\AtBeginDocument{%
  }

\setcopyright{acmlicensed}
\copyrightyear{2027}
\acmYear{2027}
\acmDOI{} 

\acmConference[KDD '27]{Proceedings of the 33rd ACM SIGKDD Conference on Knowledge Discovery and Data Mining}{August 1--5, 2027}{San Jose, CA, USA}
\acmBooktitle{Proceedings of the 33rd ACM SIGKDD Conference on Knowledge Discovery and Data Mining (KDD '27), August 1--5, 2027, San Jose, CA, USA}
\acmISBN{978-1-4503-XXXX-X/2027/08}

\usepackage[utf8]{inputenc}   
\usepackage[T1]{fontenc}       
\usepackage{xcolor}            
\usepackage{booktabs}          
\usepackage{xspace}
\usepackage{amsfonts}          
\usepackage{amsmath}           
\usepackage{nicefrac}          
\usepackage{microtype}         
\usepackage{graphicx}          
\usepackage{multirow}
\usepackage{makecell}
\usepackage{algorithm}
\usepackage{algpseudocode}
\usepackage{subcaption}
\usepackage{enumitem}
\newcommand{\framework}{\textnormal{\texttt{AMP-OD}}\xspace}

\begin{document}

\title{\framework: Active Model Probing for Unsupervised\texorpdfstring{\\}{ }Outlier Model Selection--Extended Version}

\author{Hong-Phuc Phan}
\authornote{These authors contributed equally to this work.}
\affiliation{%
  \institution{Independent Researcher}
  \country{Vietnam}
}

\author{Tuan-Anh Vu}
\authornotemark[1]
\affiliation{%
  \institution{Department of Information Technology, University of Science, VNU-HCM}
  \country{Vietnam}
}

\author{Tung Kieu}
\authornotemark[1]
\authornote{Corresponding authors: \href{mailto:tungkvt@cs.aau.dk}{tungkvt@cs.aau.dk}, \href{mailto:ha.son@rmit.edu.vn}{ha.son@rmit.edu.vn}.}
\affiliation{%
  \institution{Department of Computer Science, Aalborg University}
  \country{Denmark}
}
\author{Son Ha Xuan}
\authornotemark[2]
\affiliation{%
  \institution{The Business School, RMIT University}
  \country{Vietnam}
}

\author{Sean Bin Yang}
\affiliation{%
  \institution{Department of Computer Science, Aalborg University}
  \country{Denmark}
}

\author{Bin Yang}
\affiliation{%
  \institution{Department of Computer Science, Aalborg University}
  \country{Denmark}
}

\author{Christian S. Jensen}
\affiliation{%
  \institution{Department of Computer Science, Aalborg University}
  \country{Denmark}
}

\begin{abstract}
Outlier detection is often deployed in fully unsupervised settings, where labels are unavailable and practitioners must choose a detector without direct performance feedback. Most existing unsupervised outlier model selection methods predict detector quality from static meta-features and historical performance on meta-datasets, committing before observing how any detector behaves on the target dataset, while internal-evaluation methods must run many candidate detectors. We propose \framework, an execution-conditioned probe-then-commit framework. \framework runs a small number of inexpensive probe detectors, uses their score patterns as target-dataset evidence, and then selects a final detector from a larger candidate pool. To interpret unlabeled probe responses, we introduce a Leave-One-Out Family Probe Posterior (LOO-FPP), which converts probe score patterns into gated detector-family evidence while reducing self-validation in pseudo-label construction. \framework further combines this evidence with historical reliability through a warm-start prior, a sequential probe/stop policy, and lower-confidence-bound arbitration. Extensive experiments and ablation studies show that detector responses provide useful evidence beyond static meta-features, with LOO-FPP as the key contributor.
\end{abstract}


\maketitle

\section{Introduction}
\label{sec:introduction}

Outlier detection is a fundamental machine learning problem with applications in fraud detection, medical diagnosis, network intrusion detection, and industrial fault monitoring~\citep{DBLP:journals/csur/ChandolaBK09,DBLP:journals/pieee/RuffKVMSKDM21}. 
In many real-world settings, ground-truth labels are unavailable, so outlier detection must be performed in a fully unsupervised manner. This makes deployment challenging: practitioners must choose an effective detector without any feedback on which instances are genuinely anomalous.

A central difficulty is that detector performance is highly dataset-dependent. Classical methods such as \texttt{Isolation Forest}~\citep{DBLP:conf/icdm/LiuTZ08}, \texttt{LOF}~\citep{DBLP:conf/sigmod/BreunigKNS00}, and \texttt{HBOS}~\citep{goldstein2012hbos}, as well as deep neural networks~\citep{DBLP:conf/icml/RuffGDSVBMK18,DBLP:conf/iclr/ZongSMCLCC18}, can behave very differently across domains, data modalities, outlier types, and hyperparameter or architectural choices~\citep{DBLP:conf/nips/DingZA22,DBLP:conf/nips/JiangHZHHWH023}. Large-scale evidence from \textbf{ADBench}~\citep{DBLP:conf/nips/HanHHJ022} confirms that no single detector consistently dominates. Thus, in practice, the key question is often not how to design another outlier detector, but how to select a suitable detector for the dataset at hand.

This problem is commonly studied as \emph{unsupervised outlier model selection} (UOMS): given several candidate detectors and a new unlabeled dataset, the goal is to select the detector expected to be most effective, although labels are unavailable to verify performance during selection~\citep{DBLP:journals/corr/abs-2104-01422}. Existing UOMS methods use detector performance on previous labeled datasets, often called \emph{meta-datasets}, to predict a suitable detector. Meta-feature methods such as \texttt{MetaOD}~\citep{DBLP:conf/nips/ZhaoRA21} compute static meta-features from the target dataset and commit before observing how any detector behaves on it, while internal-evaluation methods such as \texttt{ELECT}~\citep{DBLP:conf/icdm/0016ZA22} must run many candidate detectors to compute their internal performance measures.

This limitation matters because detector suitability is often revealed by its score distribution on the target dataset. Static meta-features summarize general properties such as dimensionality, sparsity, and simple statistics, but they cannot show whether a detector produces a meaningful outlier ranking. However, running a few inexpensive detectors can reveal whether their scores clearly separate suspicious instances, remain nearly uniform, or agree or conflict with one another. These response patterns provide direct target-dataset evidence that static UOMS methods cannot use.

To address this limitation, we propose \framework (\textbf{A}ctive \textbf{M}odel \textbf{P}robing for \textbf{O}utlier \textbf{D}etection), an execution-conditioned framework for UOMS. Before selecting the final detector, \framework runs a small number of inexpensive probe detectors, summarizes their score patterns as detector-response signatures, and uses them as additional evidence. Thus, \framework combines historical and meta-feature information with the observed behavior of a few inexpensive probe detectors on the target dataset before selecting one detector from a larger candidate pool.

Two challenges arise. First, probe responses remain unlabeled, so \framework cannot know directly which probe is correct. We address this with a \emph{Leave-One-Out Family Probe Posterior} (LOO-FPP), which groups detectors into families, e.g., isolation-based, density-based, neighborhood-based, or projection-based methods. LOO-FPP constructs consensus pseudo-labels from all but one family and evaluates the held-out family against them, preventing a family from validating labels that it helped create. Second, probe evidence can be uncertain when the probe pool misses the best detector type or when probe scores are weak and conflicting. \framework therefore combines probe evidence with historical reliability through a warm-start prior learned from meta-datasets, a probe/stop policy, and a lower confidence bound (LCB) arbitrator.

We evaluate \framework with held-out-dataset protocols (leave-one-dataset-out, and leave-one-base-dataset-out where a testbed contains variants of the same base dataset) on multiple outlier-detection testbeds, with main results reported on \textbf{ADBench-Classical} and \textbf{ADBench-Advanced}. For reproducibility, probe scores are precomputed in the benchmark implementation and \framework reveals cached probe signatures sequentially during decision-making; thus, the reported probe budget measures the amount of detector-response evidence used, while detector materialization cost is accounted for separately. Empirical results and ablations show that the execution-conditioned probing and LOO-FPP both contribute to the final performance.

In summary, we make these contributions:
(i) We address a key limitation of meta-feature UOMS methods by reformulating detector selection as a probe-then-commit process, where target-dataset detector responses are observed before final selection.
(ii) We introduce LOO-FPP, a leave-one-out pseudo-labeling strategy that converts unlabeled probe responses into gated evidence over detector families.
(iii) We propose \framework, which combines probe evidence with historical reliability through a warm-start prior, a sequential probe/stop policy, and LCB arbitration.
(iv) We evaluate \framework under held-out-dataset protocols on multiple testbeds and analyze when probing helps or fails due to limited probe-family coverage.
\section{Preliminaries}
\label{sec:preliminaries}

\paragraph{Outlier Detection.} 
Let $\mathcal{D}$ denote a space of datasets. A dataset
$D \in \mathcal{D}$ is represented by a feature matrix $\mathbf{X}=(\mathbf{x}_1,\ldots,\mathbf{x}_n)^\top \in \mathbb{R}^{n\times d}$, where $n$ is the number of instances and $d$ is the feature dimensionality. In the fully unsupervised setting, ground-truth labels are unavailable when scores are produced: although labels may exist for benchmark evaluation, they are hidden during selection.

An outlier detector $m$ is an outlier detection algorithm with specified hyperparameters. Given an unlabeled dataset $D$, the detector outputs an outlier-score vector
\(
  \mathbf{s}_m(D)=m(D)\in\mathbb{R}^{n},
\)
where a larger score indicates that the corresponding instance is more likely to be an outlier. 

\paragraph{Unsupervised Outlier Model Selection.}

Unsupervised outlier model selection (UOMS) studies how to choose a suitable outlier detector for a new unlabeled dataset from a finite candidate pool
\(
  \mathcal{M}=\{m_1,\ldots,m_{|\mathcal{M}|}\}.
\)
The selection target is usually defined by a supervised evaluation metric that can be computed only when ground-truth labels are available for offline evaluation. Given such labels, the oracle detector is
\begin{equation}
  m^\star(D)
  =
  \arg\max_{m\in\mathcal{M}} P(m,D),
  \label{eq:uoms}
\end{equation}
where $P(m,D)$ denotes a supervised performance measure, such as Average Precision (AP) or Area Under the ROC Curve (AUROC). However, in the UOMS setting the labels of the target dataset are unavailable during selection, so $P(m,D)$ cannot be queried.

A common UOMS setting assumes access to labeled meta-training datasets
\(
  \mathcal{T}=\{(D_i,\mathbf{y}_i)\}_{i=1}^{N_{\mathrm{train}}},
\)
where the labels are used to estimate historical detector performance on previous datasets. These previously labeled datasets are often referred to as \emph{meta-datasets}. The target dataset is held out from all training statistics and is presented without labels at selection time. Therefore, UOMS can be viewed as supervised meta-learning across labeled meta-datasets followed by unsupervised detector selection on a new unlabeled target dataset~\citep{DBLP:journals/air/VilaltaD02,DBLP:journals/pami/HospedalesAMS22,DBLP:conf/nips/ZhaoRA21}.

\section{Methodology}
\label{sec:methodology}

\begin{figure*}[t]
    \centering
    \includegraphics[width=0.95\linewidth]{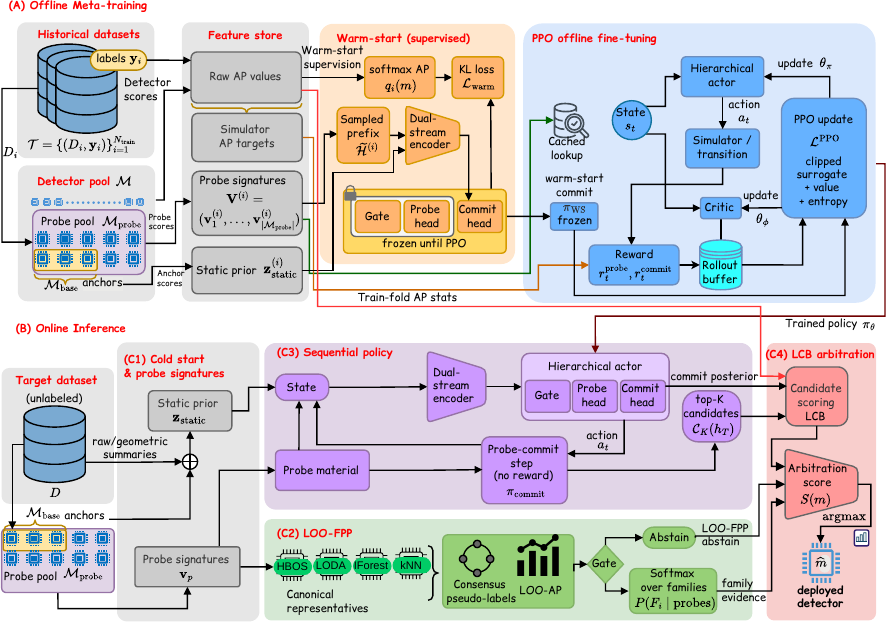}
    \Description[Overview of the framework]{Detailed architecture of the framework showing four main components: cold-start representation, LOO-FPP evidence, sequential policy, and posterior-aware LCB arbitration.}
    \caption{Overview of \framework: (\textbf{C1})~cold-start representation from anchor detectors, (\textbf{C2})~LOO-FPP label-free family evidence, (\textbf{C3})~sequential probe/stop policy, and (\textbf{C4})~posterior-aware LCB arbitration for final detector selection.}
    \label{fig:overview}
\end{figure*}

\framework is an execution-conditioned framework for unsupervised 
outlier model selection (UOMS). Given an unlabeled target dataset $D$ and 
a candidate detector pool $\mathcal{M}$, \framework first obtains 
lightweight detector responses from a small probe pool and then uses 
these responses as target-dataset evidence before selecting a final 
detector from $\mathcal{M}$. Figure~\ref{fig:overview} shows 
the full pipeline. The framework consists of four main components: 
\textbf{(C1)}~cold-start representation and probe signatures 
(Sec.~\ref{sec:method:step0}), \textbf{(C2)}~Leave-One-Out Family Probe Posterior (LOO-FPP) for 
label-free family evidence (Sec.~\ref{sec:method:loofpp}), 
\textbf{(C3)}~a sequential probe/stop policy 
(Sec.~\ref{sec:method:learning}), and 
\textbf{(C4)}~posterior-aware LCB arbitration 
(Sec.~\ref{sec:method:lcb}). This section presents the complete 
inference and learning pipeline.

\subsection{Execution-Conditioned Probe-then-Commit Formulation}
\label{sec:method:formulation}

Meta-feature UOMS methods select a detector before executing any candidate on the target dataset, while internal-evaluation methods such as \texttt{ELECT} execute many candidates. In contrast, \framework first acquires limited target-dataset evidence by probing a small set of inexpensive detectors, then decides when to stop and commit to a final detector. Related to active model selection and sequential information acquisition~\citep{DBLP:conf/nips/LiuXS024,DBLP:conf/icml/LiO21,DBLP:conf/icml/CovertQLKWL23}, we formulate this as a finite-horizon problem with two actions: \emph{probe}, which reveals a detector-response signature, and \emph{commit}, which stops probing and selects from the candidate pool.

We use three detector sets. The candidate pool $\mathcal{M}$ contains all detectors eligible for final selection. The probe pool $\mathcal{M}_{\mathrm{probe}}\subseteq\mathcal{M}$ contains inexpensive detectors whose responses can be observed before commitment. The anchor set $\mathcal{M}_{\mathrm{base}}\subseteq\mathcal{M}_{\mathrm{probe}}$ is used only to construct the initial cold-start representation. Because detectors may use different score scales, each detector score vector is independently min--max normalized:
\begin{equation}
    \displaystyle
  \widetilde{\mathbf{s}}_m(D)
  =
  \frac{
    \mathbf{s}_m(D)-\min(\mathbf{s}_m(D))
  }{
    \max(\mathbf{s}_m(D))-\min(\mathbf{s}_m(D))+\epsilon
  },
  \label{eq:method:score_norm}
\end{equation}
where $\epsilon>0$ avoids division by zero.

At step $t$, the state is
\(o_t =
  \bigl(
    \mathbf{z}_{\mathrm{static}},
    \mathcal{H}_t
  \bigr),
\)
where $\mathbf{z}_{\mathrm{static}}$ is the fixed cold-start representation, and $\mathcal{H}_t = \bigl( \mathbf{v}_{p_1},\ldots,\mathbf{v}_{p_t} \bigr)$ is the ordered history of revealed probe signatures. Each signature $\mathbf{v}_{p_j}$ summarizes the normalized score vector of probe detector $p_j$.

The action space is
\(
  \mathcal{A}
  =
  \mathcal{A}_{\mathrm{probe}}
  \cup
  \mathcal{A}_{\mathrm{commit}}.
  \label{eq:method:action_space}
\)
A probe action selects an unrevealed detector from $\mathcal{M}_{\mathrm{probe}}$ and appends its signature to $\mathcal{H}_t$; previously revealed probes are masked. A commit action stops the process and selects a detector from $\mathcal{M}$. The policy must commit within horizon $T_{\max}$. In the reported configuration, anchor detectors are excluded from the charged sequential action mask.

Let $a_t\in\mathcal{A}$ denote the action selected at decision step $t$.
A probe action incurs cost $-\lambda c(a_t)$, while a commit action receives terminal utility $u(\widehat{m}_T,D)=\operatorname{AP}(\widehat{m}_T,D)$, the supervised detector quality evaluated on cached meta-training folds inside the simulator. Let $\mathcal{I}_T$ be the set of probe steps before commitment. The policy objective is:
\begin{equation}
  J(\pi_\theta)
  =
  \mathbb{E}_{\pi_\theta}
  \left[
    u(\widehat{m}_T,D)
    -
    \sum_{t\in\mathcal{I}_T}
    \lambda c(a_t)
  \right]
  \label{eq:method:sequential_objective}
\end{equation}
Eq.~\ref{eq:method:sequential_objective} states the high-level objective; in Sec.~\ref{sec:method:learning}, probe and commit rewards (Eqs.~\ref{eq:reward_prelim}--\ref{eq:method:commit_reward}) instantiate the same probe-budget tradeoff in the offline simulator used for Proximal Policy Optimization (PPO) updates.
In our benchmark implementation, probe scores are precomputed for reproducibility and the policy is charged only for revealing cached signatures; detector materialization cost is reported separately. Thus, the objective captures the central tradeoff in \framework: acquiring more detector-response evidence can sharpen the final selection, but each additional probe consumes budget, so the policy must balance the statistical value of further evidence against its acquisition cost.




\subsection{Cold Start and Probe Signatures}
\label{sec:method:step0}

\framework uses three fast anchor detectors---\texttt{HBOS},
\texttt{LODA}, and \texttt{IForest}, collectively $\mathcal{M}_{\mathrm{base}}$---
to build a $25$-dimensional static representation
$\mathbf{z}_{\mathrm{static}}\in\mathbb{R}^{25}$ from their responses
and raw-data summaries; these anchors are not counted as charged sequential probes.

The sequential policy then operates on a fixed ten-detector probe pool $\mathcal{M}_{\mathrm{probe}}$ listed in Appx.~\ref{app:method_probe_pool}. Each probe response is compressed from the normalized score vector $\widetilde{\mathbf{s}}_p(D)$ into a $29$-dimensional signature
\(
  \mathbf{v}_p\in\mathbb{R}^{29},
\)
containing score statistics, tail information, concentration measures, detector-family identity, and probe identity. Coordinate-level feature definitions are given in Appx.~\ref{app:method_features}.

\subsection{Leave-One-Out Family Probe Posterior}
\label{sec:method:loofpp}

Probe responses must be interpreted without ground-truth labels. \framework uses LOO-FPP to convert unlabeled probe responses into evidence over detector families, where each family groups detectors based on a similar detection principle.

Let $\mathcal{G}$ contain one representative normalized score vector for each compared family or canonical probe. For each held-out representative $g\in\mathcal{G}$, LOO-FPP forms consensus pseudo-labels from the remaining representatives:
\begin{equation}
  \bar{\mathbf{s}}_{\setminus g}
  =
  \frac{1}{|\mathcal{G}|-1}
  \sum_{g'\neq g}
  \mathbf{s}_{g'},
  \qquad
  \widehat{\mathbf{y}}_{\setminus g}
  =
  \mathbb{I}
  \left[
    \bar{\mathbf{s}}_{\setminus g}
    \ge
    Q_{0.95}
    \bigl(
      \bar{\mathbf{s}}_{\setminus g}
    \bigr)
  \right]
  \label{eq:method:loo_pseudo}
\end{equation}
Here $\mathbb{I}[\cdot]$ denotes the elementwise indicator, and $Q_{0.95}(\cdot)$ denotes the 95th percentile of a score vector (a scalar threshold).
The held-out representative is then evaluated against pseudo-labels it did not help construct via the leave-one-out average precision (LOO-AP):
\(
  \operatorname{LOO\text{-}AP}(g) = \operatorname{AP}\!\left(\widehat{\mathbf{y}}_{\setminus g},\,\mathbf{s}_g\right),
\)
where $\operatorname{AP}$ denotes standard average precision between binary pseudo-labels and real-valued outlier scores.
This leave-one-out design reduces the self-validation problem of naive consensus scoring.


\paragraph{Evidence Gate.}
LOO-FPP contributes to the final decision only when probe responses are sufficiently discriminative. We evaluate a canonical four-probe set,
\(
\{\texttt{HBOS}, \texttt{LODA}, \texttt{IForest}, \texttt{kNN}\},
\)
and activate the gate when the resulting LOO-AP scores either have a large global spread or a high-quality winner with a clear margin:
\begin{equation}
      \max_g \operatorname{LOO\text{-}AP}(g)
  -
  \min_g \operatorname{LOO\text{-}AP}(g)
  \ge
  \delta_{\mathrm{range}},
  \label{eq:method:gate_range}
\end{equation}
or if the best representative has both high quality and a clear margin:
\begin{equation}
  \operatorname{LOO\text{-}AP}(g^{(1)})
  \ge
  \delta_{\mathrm{quality}}
  \wedge
  \operatorname{LOO\text{-}AP}(g^{(1)})
  -
  \operatorname{LOO\text{-}AP}(g^{(2)})
  \ge
  \delta_{\mathrm{margin}},
  \label{eq:method:gate_margin}
\end{equation}
where $g^{(1)}$ and $g^{(2)}$ are the top two representatives. Threshold values are given in Appx.~\ref{app:loofpp_details}.



\paragraph{Family Evidence.}
Let $F_i$ denote a detector family index in the softmax. When the gate fires, \framework averages probe responses within each detector family and converts the resulting LOO-AP scores into a posterior-like family-evidence distribution:
\begin{equation}
  P(F_i\mid \mathrm{probes})
  =
  \frac{
    \exp\!\left(
      \operatorname{LOO\text{-}AP}(F_i)/\tau
    \right)
  }{
    \sum_j
    \exp\!\left(
      \operatorname{LOO\text{-}AP}(F_j)/\tau
    \right)
  }
  \label{eq:method:fpp_posterior}
\end{equation}
This is a calibrated family-evidence score, not a Bayesian posterior. If the gate does not fire, LOO-FPP abstains; otherwise, the family evidence is passed to the final arbitration stage. The temperature $\tau$ is specified in Appx.~\ref{app:loofpp_details}.

\subsection{Sequential Policy and Learning}
\label{sec:method:learning}

\framework encodes the fixed cold-start representation and the revealed probe history with a dual-stream architecture. A multi-layer perceptron (MLP) maps $\mathbf{z}_{\mathrm{static}}$ to $h_{\mathrm{static}}$, while a Transformer encoder~\citep{DBLP:conf/nips/VaswaniSPUJGKP17}
maps the probe history $\mathcal{H}_t$ to 
$h_{\mathrm{dynamic}}(\mathcal{H}_t)$ using learned position and 
probe-identity embeddings. The two streams are normalized and fused:
\begin{equation}
      \displaystyle
  h_t
  =
  f_{\mathrm{fuse}}
  \left(
    \frac{h_{\mathrm{static}}}{\|h_{\mathrm{static}}\|_2},
    \frac{h_{\mathrm{dynamic}}(\mathcal{H}_t)}
         {\|h_{\mathrm{dynamic}}(\mathcal{H}_t)\|_2}
  \right).
  \label{eq:method:fused_state}
\end{equation}
Here, $f_{\mathrm{fuse}}(\cdot)$ denotes concatenation and layer normalization of the pre-normalized streams.

The fused state drives a hierarchical actor-critic policy. A gate decides whether to probe or commit; the probe head selects the next unmasked probe; and the commit head produces a posterior over candidate detectors. The commit logits use a residual form,
\(
  \ell_{\mathrm{commit}}(h_t)
  =
  f_{\mathrm{base}}(h_{\mathrm{static}})
  +
  f_{\mathrm{refine}}(h_{\mathrm{dynamic}}),
  \label{eq:method:commit_logits}
\)
so dynamic probe evidence refines the cold-start prior instead of replacing it. Here, $f_{\mathrm{base}}(\cdot)$ and $f_{\mathrm{refine}}(\cdot)$ are MLP units.

\paragraph{Warm-start Training.}
Training begins with a supervised warm-start from historical AP values on meta-training datasets. For dataset $D_i$, we sample a short probe prefix $\widetilde{\mathcal{H}}^{(i)}$ and define
\begin{equation}
  q_i(m) \propto \exp(\operatorname{AP}(m,D_i)/\tau_{\mathrm{ws}}).
\end{equation}
The commit head is trained by
\begin{equation}
  \mathcal{L}_{\mathrm{warm}}
  =
  \mathbb{E}_{D_i}
  \left[
    \operatorname{KL}
    \left(
      q_i(\cdot)
      \,\|\,
      \pi_{\mathrm{commit}}
      \bigl(
        \cdot
        \mid
        \mathbf{z}_{\mathrm{static}}^{(i)},
        \widetilde{\mathcal{H}}^{(i)}
      \bigr)
    \right)
  \right],
  \label{eq:method:warmstart}
\end{equation}
where $\operatorname{KL}(\cdot\,\|\,\cdot)$ denotes the Kullback--Leibler (KL) divergence.
This warm-start phase provides a strong model-quality prior before sequential policy optimization.

\paragraph{PPO Fine-tuning.}
After warm-starting, \framework fine-tunes the probe/stop policy with PPO in an offline simulator built from cached training-fold observations and detector AP values. Probe actions receive the cost penalty
\begin{equation}
  r_t^{\mathrm{probe}} = -\lambda\,c(a_t),
  \label{eq:reward_prelim}
\end{equation}
where $\lambda > 0$ and $c(a_t)$ is the normalized probe cost. A commit action receives a shaped terminal reward defined by improvement over the frozen warm-start baseline:
\begin{equation}
  r_T^{\mathrm{commit}}
  =
  w(n_{\mathrm{probe}})
  \left[
    \operatorname{AP}(\widehat{m}_T,D)    -
    \operatorname{AP}
    \left(
      m_{\mathrm{ws}}(p_{\min}),D
    \right)
  \right],
  \label{eq:method:commit_reward}
\end{equation}
where $m_{\mathrm{ws}}(p_{\min})$ is the warm-start commit using the first $p_{\min}$ probe signatures, and $w(n_{\mathrm{probe}})$ is a depth-dependent multiplier. PPO learns whether additional probe evidence improves over the warm-start prior while paying for probe acquisition; auxiliary objectives and thresholds (\textit{e.g.}, tying $p_{\min}$ to warm-start probing) are summarized in Appx.~\ref{app:method_training}. Benchmark pseudocode is listed in Appx.~\ref{app:method_algorithms}.

\subsection{Posterior-Aware LCB Arbitration}
\label{sec:method:lcb}

The neural commit posterior proposes candidates, but the final decision is made by conservative arbitration. At terminal state $h_T$, we form $\mathcal{C}_K(h_T)$ from the top-$K$ posterior candidates, augmented with the best training-fold detector from each family. For each candidate $m$, \framework computes
\(
  \displaystyle
  \operatorname{LCB}(m)
  =
  \overline{\operatorname{AP}}(m)
  -
  \alpha_\delta s_\rho(m)
\)
and
\(
  \displaystyle
  \alpha_\delta
  =
  \sqrt{
    \frac{
      2\log(3|\mathcal{M}|/\delta)
    }{
      N_{\mathrm{train}}
    }
  },
\)
where $\overline{\operatorname{AP}}(m)$ and $s_\rho(m)$ are the training-fold mean AP and normalized AP standard deviation (subscript $\rho$ indicates per-fold row-normalization), computed strictly within the training fold.
$N_{\mathrm{train}}$ is the number of training datasets used to form those AP statistics, $\delta\in(0,1)$ is a confidence parameter under the union-bound construction that defines $\alpha_\delta$, and $|\mathcal{M}|$ is the candidate-pool size; concrete values are given in Appx.~\ref{app:loofpp_details}.
The final score combines historical reliability, the neural posterior, and gated LOO-FPP family evidence:
\begin{equation}
  S(m)
  =
  (1-\gamma)
  \left[
    (1-\omega)
    \widetilde{\operatorname{LCB}}(m)
    +
    \omega
    \widetilde{\pi}(m\mid h_T)
  \right]
  +
  \gamma
  P(\mathcal{F}(m)\mid \mathrm{probes}),
  \label{eq:method:blend_score}
\end{equation}
where $\widetilde{\operatorname{LCB}}$ and $\widetilde{\pi}$ are min--max normalized within $\mathcal{C}_K(h_T)$, $\omega\in[0,1]$ controls their mixture inside the $(1-\gamma)$ term, $P(\mathcal{F}(m)\mid \mathrm{probes})$ denotes the softmax family-evidence masses from Eq.~\ref{eq:method:fpp_posterior}, and $\gamma\in[0,1]$ weights that family-evidence pathway.
If the LOO-FPP gate abstains, we set $\gamma=0$. The selected detector is:
\begin{equation}
  \widehat{m}
  =
  \arg\max_{m\in\mathcal{C}_K(h_T)}
  S(m)
  \label{eq:method:final_selection}
\end{equation}
Hyperparameters are given in Appx.~\ref{app:loofpp_details}. The formal derivation of the arbitration rule is provided in Appx.~\ref{app:method_lcb}. The overall complexity analysis is summarized in Appx.~\ref{app:method_complexity}.

\paragraph{Optional Family-first Route.}
With high gate confidence, \framework may restrict selection to the winning detector family. This route is accepted only if the proposed family set $\mathcal{R}$ is competitive with the best overall candidate under training-fold AP:
\(
  B(\mathcal{R}) \ge \delta_{\mathrm{guard}} B_{\mathrm{all}},
  \label{eq:method:family_guard}
\)
where $B(\mathcal{R})$ is the best training-fold AP within $\mathcal{R}$ and $B_{\mathrm{all}}$ is the best training-fold AP over all candidates. If the guard fails, \framework uses the soft arbitration score in Eq.~\ref{eq:method:blend_score}. The value of $\delta_{\mathrm{guard}}$ is given in Appx.~\ref{app:loofpp_details}.

\subsection{Probe Reachability and Observation Limits}
\label{sec:method:reachability}

Execution-conditioned selection helps only when the probe pool exposes information relevant to the best detector family. Let $\mathcal{F}(m)$ denote the family of detector $m$, and let
\(
  F^\star(D)=\mathcal{F}(m^\star(D))
\)
be the oracle family on dataset $D$. The available probe observations are
\(
  \mathcal{O}_{\mathrm{probe}}(D)
  =
  \{
    \widetilde{\mathbf{s}}_p(D)
    :
    p\in\mathcal{M}_{\mathrm{probe}}
  \}.
\)
We call $D$ \emph{probe-reachable} if $F^\star(D)$, or a closely related family, is represented in the probe pool and LOO-FPP separates it from plausible alternatives. We call $D$ \emph{probe-blind} if the probe responses cannot distinguish the oracle family from others.
This diagnostic states an observation limit: no probe-based selector can recover evidence absent from $\mathcal{O}_{\mathrm{probe}}(D)$. Thus, \framework exploits probe evidence when it is discriminative and relies more on historical reliability when the signal is weak or probe-blind.

\paragraph{Inference Summary.}
On a new unlabeled dataset, \framework builds $\mathbf{z}_{\mathrm{static}}$ from anchors and raw-data summaries, sequentially probes signatures via the learned policy, estimates gated family evidence with LOO-FPP, and selects the detector using posterior-aware LCB arbitration.
\section{Experiments}
\label{sec:experiments}

\subsection{Experimental Setup}
\label{sec:exp:setup}

\paragraph{Datasets.}
We evaluate \framework on two testbeds with a total of 137 datasets: \textbf{ADBench-Classical} (47 datasets), the primary benchmark for diverse tabular unsupervised model selection, whose full dataset statistics are listed in Appx.~\ref{app:datasets}, and \textbf{ADBench-Advanced} (90 datasets). \textbf{ADBench-Advanced} is designed to test whether a selector can recover when static dataset descriptors are misleading. Starting from inlier samples of 15 \textbf{ADBench-Classical} datasets, we inject three controlled outlier types: dependency violations, stronger dependency violations, and tight extreme clusters. We then apply two deception mechanisms, noise padding and feature permutation, which perturb static meta-features while preserving the underlying outlier geometry. Because only these pre-execution descriptors are corrupted while the true outlier structure stays intact, the best detector remains recoverable by \emph{executing} probes; the benchmark thus rewards execution-conditioned evidence rather than posing an impossible task. Such mismatch between static meta-features and detector behavior is common in practice---noisy or irrelevant feature dimensions, permuted or opaque encodings---so \textbf{ADBench-Advanced} tests robustness to realistic distribution shift rather than acting as an artificial adversary. This yields 90 datasets with the same 297-detector candidate pool, evaluated with grouped splits (Sec.~\ref{sec:exp:setup}). The construction details for \textbf{ADBench-Advanced} are provided in Appx.~\ref{app:adbenchadv_construction}.

\paragraph{Evaluation protocol.}
Experiments on \textbf{ADBench-Classical} follow a leave-one-dataset-out (LODO) protocol, training on the remaining testbed datasets and evaluating on the held-out target. \textbf{ADBench-Advanced} consists of six variants of each of 15 base datasets, so plain LODO would leave near-duplicates of the target in meta-training; we therefore evaluate it with leave-one-base-dataset-out (grouped) splits, in which all six variants of the target's base dataset are excluded from meta-training (84 training datasets per fold). Labels are reserved solely for offline evaluation, never for selection. Stratified sampling preserves outlier ratios across three random seeds. The implementation is available at \url{https://github.com/ph-phuc/AMP-OD}.

\paragraph{Detector pools.}
All methods use the same full candidate pool $\mathcal{M}$ of 297 unsupervised detectors, summarized in Appx.~\ref{app:method_model_pool}. \framework uses a smaller ten-detector probe pool listed in Appx.~\ref{app:method_probe_pool}. The first three probes (\texttt{HBOS}, \texttt{LODA}, \texttt{IForest}) also serve as cold-start anchors and are not counted as charged sequential probes in the reported configuration.


\paragraph{Baselines.}
We compare \framework with representative baselines from five groups: reference controls, standalone detectors, score aggregation methods, UOMS methods, and active probing methods without reinforcement learning (RL). All baselines use the same splits (LODO on \textbf{ADBench-Classical}, grouped on \textbf{ADBench-Advanced}) and the same 297-detector candidate pool whenever applicable. The most direct comparisons are the model-selection baselines: \texttt{GlobalBest} and \texttt{MetaOD}~\citep{DBLP:conf/nips/ZhaoRA21}, which select from meta-features and history without executing candidates on the target dataset, and \texttt{ELECT}~\citep{DBLP:conf/icdm/0016ZA22}, an internal-evaluation method that executes candidate detectors to compute its internal performance measures. 
Active probing baselines share \framework's probe pool and budget but omit its learned policy and conservative arbitration (details in Appx.~\ref{app:method_baselines}).

\paragraph{Evaluation Metrics.}
\label{sec:exp:metrics}

We use three evaluation metrics: \textbf{Average Precision} (AP), \textbf{pool rank}, and \textbf{AUROC}. Higher AP and AUROC are better, while a lower pool rank is better. 
Statistical significance is assessed via one-sided Wilcoxon signed-rank tests on paired per-dataset results ($\alpha=0.05$). See Appx.~\ref{app:eval_metrics} for metric and aggregation details.

\subsection{Main Results}
\label{sec:exp:results}

Table~\ref{tab:main_results} reports aggregate AP, mean pool rank, and AUROC on \textbf{ADBench-Classical} and \textbf{ADBench-Advanced}. \framework achieves the highest AP and the best mean pool rank on both testbeds, while remaining competitive in AUROC. These results show that execution-conditioned probing improves over standalone detectors, naive aggregation methods, and meta-feature UOMS baselines. The gains over standalone detectors and aggregation are largest on \textbf{ADBench-Advanced}, where static meta-features are deliberately misleading; against the strongest baseline there, \texttt{ELECT}, the margin is $+0.026$ AP (Table~\ref{tab:significance}).

\begin{table}[t]
\caption{Main comparison. AP: Average Precision
  ($\uparrow$). Rank: pool rank ($\downarrow$; pool size 297).
  AUROC: area under ROC curve ($\uparrow$).
  Point estimates are seed-averaged; per-seed standard deviations are
  reported in Table~\ref{tab:main_results_full}.
  Significance tests use seed-averaged per-dataset AP.
  \textbf{ADBench-Advanced} uses leave-one-base-dataset-out splits (all variants of the target's base dataset are excluded from meta-training); this affects only the methods that learn from training datasets (\texttt{GlobalBest}, \texttt{MetaOD}, \texttt{ELECT}, \framework); the other methods do not use training datasets, so their entries are unchanged.
  }
  \label{tab:main_results}
  \centering
  \fontsize{7.5pt}{7.5pt}\selectfont
  \setlength{\tabcolsep}{2.0pt}
  \begin{tabular}{lcccccc}
    \toprule
    & \multicolumn{3}{c}{\textbf{ADBench-Classical}}
    & \multicolumn{3}{c}{\textbf{ADBench-Advanced}} \\
    \cmidrule(lr){2-4}\cmidrule(lr){5-7}
    Method
      & AP\,$\uparrow$
      & Rank\,$\downarrow$
      & AUROC\,$\uparrow$
      & AP\,$\uparrow$
      & Rank\,$\downarrow$
      & AUROC\,$\uparrow$ \\
    \midrule
    \multicolumn{7}{@{}l}{\textit{Reference controls}} \\[1pt]
    \texttt{Oracle}
      & \textcolor{violet}{0.5930}
      & \textcolor{violet}{1.0}    
      & \textcolor{violet}{0.8580} 
      & \textcolor{violet}{0.7396}
      & \textcolor{violet}{1.0}    
      & \textcolor{violet}{0.9524} \\
    \texttt{RandomSelect}
      & 0.3128  & 145.5 & 0.6851 
      & 0.3049  & 145.6 & 0.7494  \\
    \addlinespace[3pt]
    \multicolumn{7}{@{}l}{\textit{Standalone detectors}} \\[1pt]
    \texttt{IForest}
      & \underline{0.3863} & 104.2 & 0.7679 
      & 0.3830 & 124.7 & \underline{0.8621} \\
    \texttt{kNN}
      & 0.3546 & 122.3 & 0.7238
      & 0.2773 & 113.8 & 0.6706 \\
    \texttt{LOF}
      & 0.3410 & 134.2 & 0.7156
      & 0.4064 & \underline{74.7} & 0.7983 \\
    \texttt{DAGMM}
      & 0.3112 & 173.3 & 0.6989
      & 0.2851 & 187.2 & 0.7325 \\
    \texttt{DeepSVDD}
      & 0.2076 & 204.5 & 0.5929
      & 0.1772 & 184.4 & 0.6875 \\
    \texttt{RandNet}
      & 0.3368 & 142.2 & 0.7164
      & 0.4028 & 133.1 & 0.8020 \\
    \texttt{RDA}
      & 0.2774 & 156.4 & 0.7160
      & 0.3283 & 110.6 & 0.6788 \\
    \texttt{DTE-C}
      & 0.3465 & 141.0 & 0.7465
      & 0.3787 & 106.7 & 0.8028 \\
    \texttt{TCCM}
      & 0.3151 & 149.5 & 0.7059
      & 0.2686 & 162.0 & 0.6999 \\
    \texttt{LUNAR}
      & 0.2765 & 166.5 & 0.6188
      & 0.3475 & 102.4 & 0.7544 \\
    \addlinespace[3pt]
    \multicolumn{7}{@{}l}{\textit{Aggregation baselines}} \\[1pt]
    \texttt{LSCP}
      & 0.3517 & 133.5 & 0.7287
      & 0.3806 & 98.4 & 0.8282 \\
    \texttt{Full-Pool Ensemble}
      & 0.3127 & 151.2 & 0.7213
      & 0.2372 & 173.3 & 0.6655 \\
    \addlinespace[3pt]
    \multicolumn{7}{@{}l}{\textit{Model selection}} \\[1pt]
    \texttt{GlobalBest}
      & 0.3686 & \underline{101.4} & \underline{0.7698} 
      & 0.4133 & 81.4 & 0.8007 \\
    \texttt{ELECT}
      & 0.3728 & 115.6 & 0.7592 
      & \underline{0.4586} & 106.8 & 0.8442  \\
    \texttt{MetaOD}
      & 0.3434 & 141.7 & 0.7027
      & 0.4117 & 122.5 & 0.8023 \\
    \midrule
    \framework \textit{(ours)}
      & \textbf{0.4146} & \textbf{82.7} & \textbf{0.7820}
      & \textbf{0.4848} & \textbf{68.6} & \textbf{0.8693} \\
    \bottomrule
  \end{tabular}%
\end{table}

\subsection{Dataset-Level and Outlier-Type Analysis}
\label{sec:exp:datasetlevel}

One-sided Wilcoxon signed-rank tests in Appx.~\ref{app:eval_metrics} show significant improvements over every baseline on both testbeds, also after Holm correction over the 16 baselines. On \textbf{ADBench-Advanced}, \framework outperforms \texttt{MetaOD} and \texttt{ELECT} on $63/90$ and $57/90$ datasets, respectively.

\begin{table}[t]
\centering
\small
\setlength{\tabcolsep}{4.5pt}
\caption{AP on \textbf{ADBench-Advanced} by outlier type (leave-one-base-dataset-out splits), versus the
two learned UOMS baselines (mean over 15 base datasets $\times$ 3 seeds; 15 datasets per row).
\framework is best on the four dependency combinations; on tight-cluster, \texttt{MetaOD} (noise) and
\texttt{ELECT} (permute) are best. \textbf{Bold}: best per row. The full 17-method breakdown is in
Appx.~\ref{app:per_dataset_results}.}
\label{tab:outlier_type}
\begin{tabular}{llccc}
\toprule
Outlier type & Deception & \texttt{MetaOD} & \texttt{ELECT} & \framework \\
\midrule
Dependency          & Noise   & 0.1697 & 0.2648 & \textbf{0.3444} \\
Dependency          & Permute & 0.2729 & 0.2735 & \textbf{0.4040} \\
Dep.-Strong         & Noise   & 0.1666 & 0.2747 & \textbf{0.2806} \\
Dep.-Strong         & Permute & 0.2464 & 0.2577 & \textbf{0.4275} \\
Tight-Cluster       & Noise   & \textbf{0.7915} & 0.7765 & 0.5860 \\
Tight-Cluster       & Permute & 0.8230 & \textbf{0.9041} & 0.8663 \\
\midrule
\textbf{All (90)}   &         & 0.4117 & 0.4586 & \textbf{0.4848} \\
\bottomrule
\end{tabular}
\end{table}

Table~\ref{tab:outlier_type} breaks these \textbf{ADBench-Advanced} results down by outlier type against the two learned UOMS baselines, \texttt{MetaOD} and \texttt{ELECT}. \framework improves over both on all four \emph{dependency} combinations, where the deception mechanisms most corrupt static meta-features. On \emph{tight-cluster}, \texttt{MetaOD} is best under noise and \texttt{ELECT} under permutation. The advantage of \framework over these two learned selectors therefore comes from the dependency regimes.

\begin{figure}[t]
  \centering
  \includegraphics[width=\linewidth]{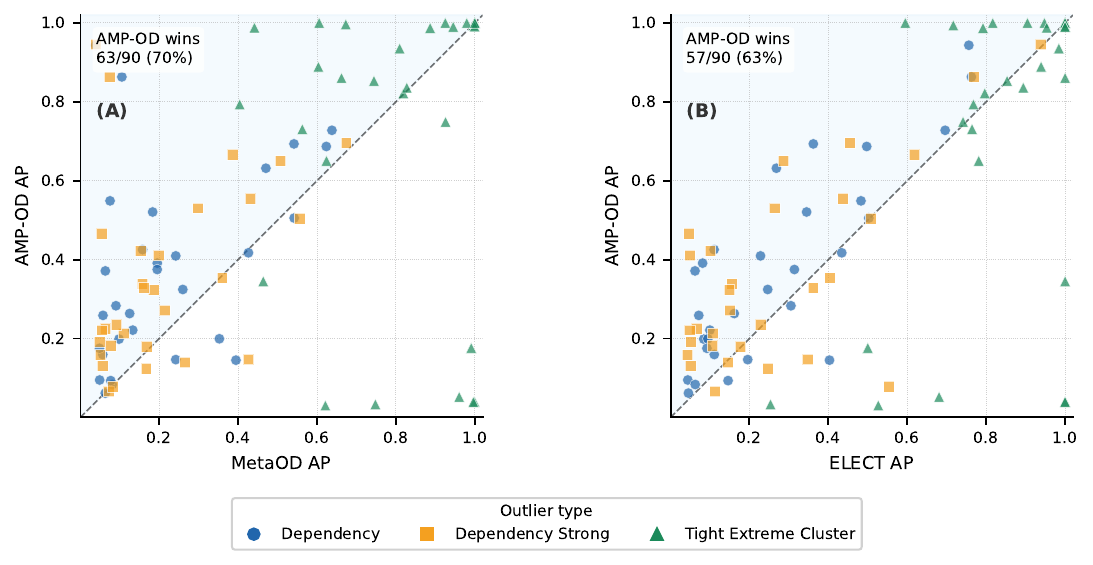}
  \caption{Dataset-level AP scatter on \textbf{ADBench-Advanced}. Panel~(A): \framework vs.\ \texttt{MetaOD}; Panel~(B): \framework vs.\ \texttt{ELECT}. Each point is one dataset; points above the diagonal are datasets where \framework achieves higher AP. Leave-one-base-dataset-out splits. Gains concentrate on the \emph{dependency} types, where static meta-features are least reliable; on \emph{tight-extreme-cluster} the results are mixed.}
  \Description{Two scatter plots compare dataset-level average precision of AMP-OD against MetaOD and ELECT on ADBench-Advanced. Most points lie above the diagonal (63 of 90 against MetaOD, 57 of 90 against ELECT), with the largest gains in dependency datasets.}
  \label{fig:adv_scatter}
\end{figure}

Figure~\ref{fig:adv_scatter} shows this at the per-dataset level: most points lie above the diagonal, with the largest gains on the \emph{dependency} datasets (Table~\ref{tab:outlier_type}). \framework does not win everywhere---static baselines stay competitive where their descriptors happen to align with the target geometry---which is consistent with our observation-limit analysis: probing helps most when probe responses expose family evidence that static descriptors miss. Full per-dataset tables are deferred to Appx.~\ref{app:per_dataset_results}.

\subsection{Ablation Study}
\label{sec:exp:ablation}

\begin{table}[t]
  \caption{Component ablations for \framework\ (3 seeds; LODO on \textbf{ADBench-Classical}, leave-one-base-dataset-out on \textbf{ADBench-Advanced}).
  AP: mean Average Precision ($\uparrow$; per-seed standard deviations in Table~\ref{tab:ablation_app}).
  Rank: mean pool rank ($\downarrow$; pool size $=297$).
  Probes: mean number of probes before commit.
  $\Delta$AP: signed gap vs.\ Full \framework (\textbf{bold}: reference row). $^\dagger$\,Probe counts for $-$LOO-FPP and $-$LCB match Full \framework by construction: these ablations share the full PPO stopping policy and modify only the commit-time evidence aggregation.}
  \label{tab:ablation}
  \centering
  \fontsize{7.5pt}{7.5pt}\selectfont
  \setlength{\tabcolsep}{0.9pt}
  \begin{tabular}{lcccccccc}
    \toprule
    & \multicolumn{4}{c}{\textbf{ADBench-Classical}}
    & \multicolumn{4}{c}{\textbf{ADBench-Advanced}} \\
    \cmidrule(lr){2-5}\cmidrule(lr){6-9}
    Variant
      & AP\,$\uparrow$
      & Rank\,$\downarrow$
      & Probes
      & $\Delta$AP
      & AP\,$\uparrow$
      & Rank\,$\downarrow$
      & Probes
      & $\Delta$AP \\
    \midrule
    \textbf{Full AMP-OD}
      & \textbf{0.4146} & \textbf{82.7} & \textbf{3.72} & $\phantom{-}0.000$
      & \textbf{0.4848} & \textbf{68.6} & \textbf{4.12} & $\phantom{-}0.000$ \\
    \midrule
    \texttt{WS-only} (no PPO)
      & 0.4031 & 89.2 & 5.50 & $-0.011$
      & 0.4848 & 68.6 & 4.12 & $\phantom{-}0.000$ \\
    \addlinespace[1pt]
    $-$\,LOO-FPP$^\dagger$
      & 0.3856 & 95.0 & 3.72 & $-0.029$
      & 0.4356 & 71.5 & 4.12 & $-0.049$ \\
    \addlinespace[1pt]
    $-$\,LCB Prior$^\dagger$
      & 0.3935 & 97.1 & 3.72 & $-0.021$
      & 0.4918 & 63.4 & 4.12 & $+0.007$ \\
    \addlinespace[1pt]
    \texttt{Static-Only}
      & 0.3991 & 90.6 & 0.00 & $-0.015$
      & 0.4578 & 71.2 & 0.00 & $-0.027$ \\
    \addlinespace[1pt]
    $\lambda{=}0$ (no probe cost)
      & 0.4040 & 88.5 & 3.93 & $-0.011$
      & 0.4848 & 68.6 & 4.12 & $\phantom{-}0.000$ \\
    \bottomrule
  \end{tabular}
\end{table}

To assess component contributions, we disable key \framework mechanisms under the same protocols (LODO on \textbf{ADBench-Classical}, leave-one-base-dataset-out on \textbf{ADBench-Advanced}). Table~\ref{tab:ablation} summarizes results on \textbf{ADBench-Classical} and \textbf{ADBench-Advanced}. On \textbf{ADBench-Classical} every ablated variant degrades AP, rank, or probe efficiency. Removing LOO-FPP causes the largest drop, showing label-free family evidence is the most critical component. Removing the LCB prior hurts on \textbf{ADBench-Classical}, but not on \textbf{ADBench-Advanced} ($+0.007$, not significant). \textsc{Static-Only} performs worse than the full model, confirming that target-dataset probe responses provide information beyond static meta-features. On \textbf{ADBench-Classical}, warm-start-only (\textsc{WS-only}) and $\lambda{=}0$ show the learned policy and probe-cost regularization control additional evidence acquisition; on \textbf{ADBench-Advanced} both make the same selections as Full \framework.


\begin{figure}[t]
  \centering
  \includegraphics[width=\linewidth]{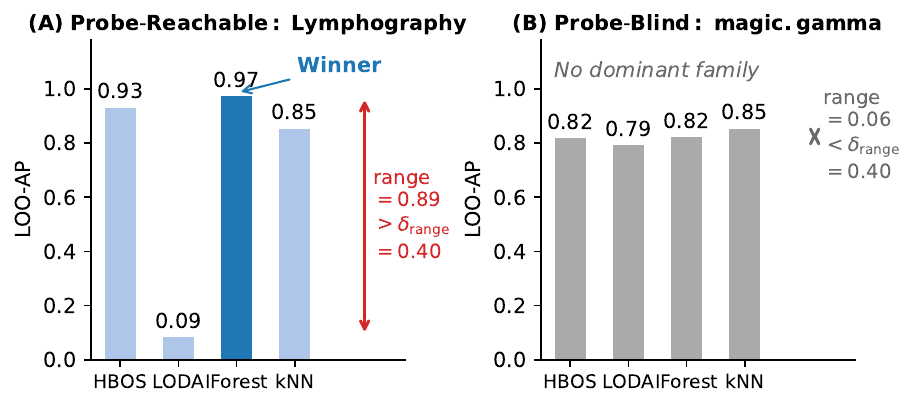}
  \caption{\textbf{LOO-FPP gate contrast.} LOO-AP profiles for the four
  canonical probe families on two datasets. \textbf{(A) Probe-reachable}
  (\textbf{Lymphography}): \texttt{IForest} tops the profile ($0.975$) while
  \texttt{LODA} collapses ($0.085$)---a range of $0.89 \gg
  \delta_{\text{range}}=0.40$---so the gate fires the family override
  (AP $=0.976$). \textbf{(B) Probe-blind} (\textbf{magic.gamma}): all families
  cluster tightly (range $=0.06 < \delta_{\text{range}}$), so the gate abstains
  and falls back to LCB-prior blending (AP $=0.706$).}
  \Description{Two side-by-side LOO-AP profiles for HBOS, LODA, IForest, and kNN. On Lymphography, IForest is highest and LODA is lowest, producing a range of 0.89 that exceeds the 0.40 gate threshold. On magic.gamma, all four values lie within a range of 0.06, below the threshold.}
  \label{fig:gate_contrast}
\end{figure}

Among all components, LOO-FPP is the most impactful: removing it reduces mean AP by $-0.029$ on \textbf{ADBench-Classical} and $-0.049$ on \textbf{ADBench-Advanced} (mean pool rank $68.6\to71.5$); on \textbf{ADBench-Advanced} the difference is not significant over datasets ($p=0.24$), because LOO-FPP helps strongly on the probe-reachable datasets and costs a little elsewhere. To see \emph{when} LOO-FPP helps, we stratify test datasets post hoc by whether the gate is open and the winning probe reaches LOO-AP $\geq\delta_{\mathrm{commit}}=0.75$ (\emph{probe-reachable}) or not (\emph{probe-blind}); this stratum is for analysis only and is not a deployed route. The gate itself fires on $45\%$ of \textbf{ADBench-Classical} and $57\%$ of \textbf{ADBench-Advanced} datasets; $17\%$ and $29\%$ are probe-reachable. On this probe-reachable subset, LOO-FPP lifts AP from $0.521$ to $0.542$ (Classical) and from $0.346$ to $0.573$ (Advanced), i.e., $+2.1$ and $+22.7$ pp. Figure~\ref{fig:gate_contrast} shows the mechanism: on \textbf{Lymphography} the LOO-AP profile spreads widely (range $0.89 \gg \delta_{\text{range}}=0.40$) and the gate fires a family override, whereas on \textbf{magic.gamma} all families cluster (range $0.06$) and the gate abstains. LOO-FPP is thus an interpretable, evidence-gated mechanism rather than a black box.
\begin{sloppypar}

The remaining components each earn their place. \textsc{Static-Only}---committing from static meta-features without any probing---loses $-0.015$ AP on \textbf{ADBench-Classical} and $-0.027$ on \textbf{ADBench-Advanced} (rank $68.6\to71.2$; $p=0.07$), consistent with probe responses adding information once static descriptors are corrupted. Dropping the probe-cost penalty ($\lambda{=}0$) changes little: $-0.011$ AP on \textbf{ADBench-Classical}, and on \textbf{ADBench-Advanced} the selections are identical to Full \framework. Removing the LCB prior more than doubles the seed-level AP standard deviation ($\pm0.018$ vs.\ $\pm0.007$) and costs $-0.021$ AP on \textbf{ADBench-Classical}, showing that conservative arbitration stabilizes the final commit against noisy probe evidence; on \textbf{ADBench-Advanced}, removing it does not hurt ($+0.007$, not significant). Architecturally, the supervised warm-start is essential: initializing PPO from scratch costs $-0.049$ AP on \textbf{ADBench-Classical} and roughly doubles the seed standard deviation, confirming it as a necessary tractability mechanism for the $297$-way action space. Replacing the Transformer probe encoder with a flat MLP is milder in mean AP ($-0.014$) but destabilizes the learned stopping rule (probe-count std $2.36$ vs.\ $0.11$), so order-sensitive temporal modeling matters for a consistent stopping criterion.

\end{sloppypar}

Overall, the ablation results support the intended division of roles in \framework: LOO-FPP interprets unlabeled probe responses, dynamic probing supplies target-dataset evidence, the learned policy controls evidence acquisition, and LCB arbitration stabilizes the final decision. The family cross-entropy (FamCE) auxiliary loss and the full gate-threshold sensitivity sweep are analyzed in Appx.~\ref{app:exp:ablation}.

\subsection{Value of the Learned Policy}
\label{sec:exp:policy_value}

We further evaluate whether the learned probe/stop policy provides benefits beyond fixed-budget probing. We compare \framework with controlled baselines that share the same trained commit head but replace the PPO stopping rule with fixed probe budgets and simple probe orders (Table~\ref{tab:policy_value_main}). \framework achieves the best AP on both testbeds (tied with \textsc{WS-only} on \textbf{ADBench-Advanced}) and better accuracy--cost tradeoffs than both max-budget (\textsc{Greedy Max}, $k{=}7$) and static-budget (\textsc{Fixed-K}/\textsc{Random}, $k{=}4$) heuristics---note that \textsc{Fixed-K} trails by $3.3$/$6.4$ AP points at a nearly identical probe count. The main benefit is adaptive depth calibration: \framework stops early when additional probes are likely to add noise, and probes further when the target dataset requires more evidence, whereas the ablated \textsc{WS-only} variant over-probes on \textbf{ADBench-Classical} (on \textbf{ADBench-Advanced} it makes the same selections as Full \framework). Full per-variant statistics are in Appx.~\ref{app:exp:policy_value}.

\begin{table}[t]
\centering
\small
\setlength{\tabcolsep}{4pt}
\caption{Value of the learned policy (3 seeds; \textbf{ADBench-Advanced} on leave-one-base-dataset-out splits). Fixed-budget baselines
share \framework's trained commit head but replace the PPO stopping rule.
\framework attains the best AP on both testbeds (tied with \textsc{WS-only} on Advanced) while spending an
\emph{adaptive} probe count (per-testbed mean; fixed-budget baselines have
$\sigma{=}0$). Full breakdown (rank, std, $\Delta$AP) in
Appx.~\ref{app:exp:policy_value}. \textbf{Bold}: best AP.}
\label{tab:policy_value_main}
\begin{tabular}{@{}lccc@{}}
\toprule
Variant & Probes & AP (Cls.) & AP (Adv.) \\
\midrule
\textbf{Full \framework}          & 3.7/4.1 & \textbf{0.4146} & \textbf{0.4848} \\
\textsc{WS-only} (no PPO)          & 5.5/4.1 & 0.4031 & \textbf{0.4848} \\
\textsc{Greedy Max} ($k{=}7$)      & 7.0     & 0.3795 & 0.4236 \\
\textsc{Fixed-K} ($k{=}4$)         & 4.0     & 0.3812 & 0.4208 \\
\textsc{Random} ($k{=}4$)          & 4.0     & 0.3825 & 0.4271 \\
\bottomrule
\end{tabular}
\end{table}

\subsection{Computational Cost}
\label{sec:exp:cost}

Execution-conditioned selection introduces additional overhead because \framework observes probe responses before committing to a detector. We therefore measure per-dataset selection cost and compare it with representative static and aggregation-based baselines. All timings report \emph{selection overhead only}; the final execution of the chosen detector is common to all methods and excluded from the comparison.

\begin{table}[t]
  \centering
  \small
  \setlength{\tabcolsep}{6pt}
  \caption{Wall-clock \emph{selection} overhead per test dataset (deployment mode: from raw data, all 47 \textbf{ADBench-Classical} datasets, single core, Intel Core Ultra~7 265K). The final detector execution is common to all methods and excluded. Full cost accounting in Appx.~\ref{app:exp:cost}.}
  \label{tab:timing}
  \begin{tabular}{@{}lrr@{}}
    \toprule
    Method & Median (ms) & p75 (ms) \\
    \midrule
    \framework  & 1{,}915 & 3{,}186 \\
    \texttt{MetaOD}  & 400 & 450 \\
    \texttt{ELECT}$^{\dagger}$   & ${\approx}47{,}000$ & ${\approx}48{,}000$ \\
    \texttt{GlobalBest} & $<1$ & $<1$ \\
    \texttt{Full-Pool Ens.} & ${\approx}83{,}000$ & ${>}130{,}000$ \\
    \bottomrule
  \end{tabular}
  \\[3pt]
  {\small $^{\dagger}$\,\texttt{ELECT}'s ${\approx}47$\,s is from~\citep{DBLP:conf/icdm/0016ZA22}; our algorithm-only overhead with pre-cached internal performance measures is ${\approx}6.5$\,s.}
\end{table}

\begin{figure}[t]
  \centering
  \includegraphics[width=\linewidth]{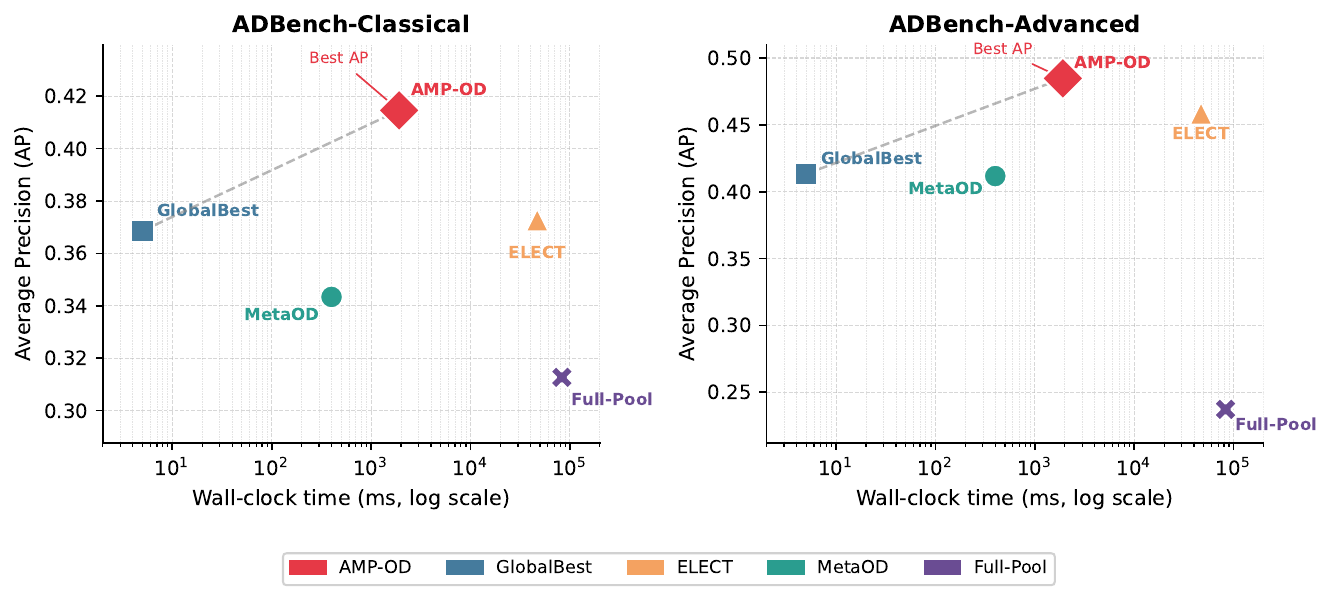}
  \caption{Cost--accuracy trade-off on both testbeds. Each marker is a selection
  method (colors match the legend), positioned by its deployment selection
  overhead (wall-clock time, log scale, $x$-axis) and mean Average Precision
  ($y$-axis); the dashed line traces the Pareto frontier, and markers toward the
  upper-left are preferable. \framework is placed at its measured median of ${\approx}1.9$\,s (Table~\ref{tab:timing}); \textbf{ADBench-Advanced} uses leave-one-base-dataset-out splits. \framework (red diamond) sits at the top of both
  panels and on the frontier, whereas the internal-evaluation and full-pool
  baselines fall far to the lower-right. Times are selection overhead only; full
  cost accounting is in Appx.~\ref{app:exp:cost}.}
  \Description{Two cost--accuracy plots compare selection time on a logarithmic horizontal axis with average precision on the vertical axis for the two ADBench testbeds.}
  \label{fig:pareto}
\end{figure}

In benchmark mode, where probe scores are precomputed for reproducible held-out-dataset evaluation, \framework's policy and arbitration add only ${\approx}35$\,ms per dataset. In deployment mode, where probes run on raw data, its median selection overhead is ${\approx}1.9$\,s (Table~\ref{tab:timing}): LOO-FPP reads the scores of all ten probe-pool detectors at cold start, so this is the cost of running those ten, roughly $5\times$ the cost of \texttt{MetaOD}'s four landmarkers, with higher AP on both testbeds. Against detector-heavy internal-evaluation methods \framework is far cheaper: \texttt{ELECT} must run ${\sim}50$ detectors to form its internal measures, and \framework is ${\approx}25\times$ faster than its reported deployment time, and more accurate on both testbeds; it also avoids the ${\approx}83$\,s cost of ensembling all 297 detectors. Taken together, these comparisons position \framework at a favorable cost--accuracy operating point, visualized in Figure~\ref{fig:pareto}; full measurement details are provided in Appx.~\ref{app:exp:cost}.

\subsection{Discussion}
\label{sec:exp:discussion}

The results show that execution-conditioned evidence improves unsupervised outlier model selection, especially under distribution shift. The largest gains come from LOO-FPP, which interprets unlabeled probe responses; LCB arbitration, which avoids over-commitment to noisy signals or unreliable detector families, helps on \textbf{ADBench-Classical} ($-0.021$ without it) but not on \textbf{ADBench-Advanced}. PPO fine-tuning gives smaller gains on \textbf{ADBench-Classical}, indicating that its main role is to calibrate when to stop probing. \framework has the highest mean AP on both \textbf{ADBench} testbeds and is significantly better than every baseline, including \texttt{ELECT}, also when no variant of the target's base dataset is seen in meta-training. These findings support the central claim that detector responses provide useful target-dataset evidence when combined with label-free evaluation and conservative final selection. Beyond accuracy, these gains come at modest cost: running ten inexpensive probe detectors keeps selection overhead at about two seconds, a few times that of static meta-feature methods and far below internal-evaluation baselines such as \texttt{ELECT}, making execution-conditioned selection practical for deployment rather than a purely offline analysis tool. Taken together, the results suggest that a small amount of \emph{targeted} probing---interpreted through label-free family evidence and committed conservatively---is a favorable operating point for unsupervised model selection under distribution shift, capturing most of the benefit of observing detector behavior without the cost of exhaustively running the candidate pool.
\section{Related Work}
\label{sec:related}

\paragraph{Unsupervised Outlier Detection and Model Selection.}
Outlier detection has been studied widely in fully unsupervised settings, with methods spanning proximity-based detectors such as \texttt{kNN}~\citep{DBLP:conf/icpr/HautamakiKF04}, \texttt{LOF}~\citep{DBLP:conf/sigmod/BreunigKNS00}, and \texttt{COF}~\citep{DBLP:conf/pakdd/TangCFC02}; isolation and ensemble methods such as \texttt{IForest}~\citep{DBLP:conf/icdm/LiuTZ08}, \texttt{LODA}~\citep{DBLP:journals/ml/Pevny16}, and \texttt{HBOS}~\citep{goldstein2012hbos}; distribution-based methods such as \texttt{COPOD}~\citep{DBLP:conf/icdm/LiZBIH20} and \texttt{ECOD}~\citep{DBLP:journals/tkde/LiZHBIC23}; boundary- and angle-based methods such as \texttt{OCSVM}~\citep{DBLP:conf/nips/ScholkopfWSSP99} and \texttt{ABOD}~\citep{DBLP:conf/kdd/KriegelSZ08}; and deep models such as \texttt{DeepSVDD}~\citep{DBLP:conf/icml/RuffGDSVBMK18} and \texttt{DAGMM}~\citep{DBLP:conf/iclr/ZongSMCLCC18}. Surveys are provided by Chandola et al.~\citep{DBLP:journals/csur/ChandolaBK09} and Ruff et al.~\citep{DBLP:journals/pieee/RuffKVMSKDM21}. Benchmarks and sensitivity studies show strong dataset and configuration dependence~\citep{DBLP:conf/nips/HanHHJ022,DBLP:conf/nips/DingZA22,DBLP:conf/nips/JiangHZHHWH023}, motivating UOMS. Internal criteria such as Excess-Mass and Mass-Volume curves~\citep{DBLP:journals/corr/Goix16} can be unreliable across datasets~\citep{DBLP:journals/tkdd/MarquesCSZ20,DBLP:journals/corr/abs-2104-01422}. \texttt{MetaOD}~\citep{DBLP:conf/nips/ZhaoRA21}, \texttt{ELECT}~\citep{DBLP:conf/icdm/0016ZA22}, \texttt{HyPer}~\citep{DBLP:conf/kdd/Ding0A24}, and robust anomaly meta-learning~\citep{DBLP:conf/aistats/KumagaiITF23} transfer historical-task information, while \texttt{AutoMS} selects a detector with false-discovery control in clean-reference novelty detection~\citep{DBLP:conf/nips/ZhangJRZD22}. \framework instead uses unlabeled target-dataset probe responses before selection.

\paragraph{Label-Free Evaluation and Pseudo-Labeling.}
Because labels are unavailable in unsupervised outlier detection, prior work estimates detector quality using surrogate signals~\citep{DBLP:journals/csur/PangSCH21,DBLP:journals/pieee/RuffKVMSKDM21}. Internal criteria such as Excess-Mass and Mass-Volume curves evaluate score functions geometrically~\citep{DBLP:journals/corr/Goix16}, but their reliability is limited across diverse datasets~\citep{DBLP:journals/tkdd/MarquesCSZ20}. Another line constructs pseudo-labels or robust scores from detector agreement and diversity, as in feature bagging~\citep{DBLP:conf/kdd/LazarevicK05}, outlier ensembles~\citep{DBLP:journals/sigkdd/AggarwalS15,DBLP:journals/tkdd/RayanaA16}, \texttt{LSCP}~\citep{DBLP:conf/sdm/ZhaoNHL19}, and sequential ensemble selectors~\citep{DBLP:conf/icdm/RayanaZA16}. However, naive consensus can be circular because a detector may help create the pseudo-labels used to evaluate it. \framework addresses this with LOO-FPP, which constructs pseudo-labels from all but one detector family and evaluates the held-out family, turning unlabeled probe responses into family-level evidence while avoiding self-validation.

\paragraph{Meta-Learning and Sequential Decision-Making.}
Algorithm selection and meta-learning are well studied in supervised machine learning~\citep{DBLP:journals/air/VilaltaD02}, with AutoML systems such as \texttt{Auto-Sklearn}~\citep{DBLP:conf/nips/FeurerKESBH15} using meta-features for model choice. Outlier selection is harder because test labels are unavailable and performance varies across datasets~\citep{DBLP:journals/sigkdd/ZimekCS13}. \texttt{AutoOD}~\citep{DBLP:conf/icde/LiCZZJCH21} and \texttt{T-AutoOD}~\citep{DBLP:conf/ijcai/VuKAS24} make static choices or target instance-level scoring. Active model selection and testing~\citep{NEURIPS2020_0c72cb7e,DBLP:conf/icml/KossenFGR21,DBLP:conf/nips/LiuXS024}, active feature acquisition~\citep{DBLP:conf/icml/LiO21,DBLP:conf/icml/CovertQLKWL23}, and RL~\citep{DBLP:journals/corr/SchulmanWDRK17} gather information sequentially, typically for supervised targets. Unlike these settings, \framework observes neither labels nor task losses: each query returns only an unlabeled detector-response pattern. It therefore bridges one-shot meta-selection and detector ensembles by updating a historical prior from execution feedback before committing to one deployable model.
\vspace{2pt}
\section{Conclusion}
\label{sec:conclusion}

We presented \framework, an execution-conditioned framework for unsupervised outlier model selection. Rather than relying only on static meta-features, \framework uses the responses of inexpensive probe detectors as target-dataset evidence before selecting the final detector. Its core component, LOO-FPP, converts unlabeled probe responses into gated family-level evidence, while lower-confidence-bound arbitration combines this evidence with historical reliability. Experiments under held-out-dataset evaluation (leave-one-dataset-out, and leave-one-base-dataset-out on \textbf{ADBench-Advanced}) show that detector responses can improve unsupervised model selection. Ablation studies further identify LOO-FPP as the most important component and show the benefit of dynamic probing; conservative final selection helps on \textbf{ADBench-Classical}. Future work will explore adaptive probe-pool construction and extend the framework to streaming and resource-constrained deployment settings.

\newpage

\bibliographystyle{ACM-Reference-Format}
\bibliography{references}

\appendix
\counterwithin{figure}{section}
\counterwithin{table}{section}
\section{Appendix}

\subsection{Probe Pool and Detector Families}
\label{app:method_probe_pool}

\framework uses a fixed probe pool $\mathcal{M}_{\text{probe}}$ for evidence acquisition and a larger benchmark-dependent candidate pool $\mathcal{M}$ for final commitment. The probe pool is intentionally small, diverse, and inexpensive: it spans eight algorithmic families while remaining short enough for sequential decision-making. Table~\ref{tab:app_probe_pool} lists the exact pool used in the reported configuration.

\begin{table}[ht]
  \caption{Probe pool used by \framework. Detectors marked ``anchor'' are also used
  in the static prior initialization stage. In the implementation, these three
  anchor detectors are evaluated before the sequential Markov decision process (MDP) begins, so their
  scores are already cached and do not consume the sequential probing budget.
  Because this anchor information is already embedded in the static prior, the
  sequential policy only chooses from probes that remain available under the
  action mask at each step.}
  \label{tab:app_probe_pool}
  \centering
  \fontsize{7.5pt}{7.5pt}\selectfont
  \begin{tabular}{ccll}
    \toprule
    Index & Detector & Family & Role \\
    \midrule
    $1$ & HBOS & Histogram & probe, anchor \\
    $2$ & LODA & Projection & probe, anchor \\
    $3$ & IForest & Tree & probe, anchor \\
    $4$ & kNN(largest, $k=5$) & Distance & probe \\
    $5$ & kNN(mean, $k=10$) & Distance & probe \\
    $6$ & LOF(euclidean, $k=10$) & Density & probe \\
    $7$ & LOF(manhattan, $k=20$) & Density & probe \\
    $8$ & OCSVM(RBF) & Boundary & probe \\
    $9$ & ABOD($k=5$) & Angle & probe \\
    $10$ & COF($k=5$) & Connectivity & probe \\
    \bottomrule
  \end{tabular}
\end{table}

The three anchor detectors are exactly the first three entries of Table~\ref{tab:app_probe_pool}. They provide the cold-start prior and are not treated as part of the sequential budget. All later sequential actions are drawn from the same $10$-detector probe pool, subject to masking that prevents reusing a previously executed probe. In the reported implementation, these three anchor probes are additionally masked out from sequential actions from the start of each episode, so they are used only in the cold-start stage.

\subsection{Static and Dynamic Feature Definitions}
\label{app:method_features}

All score-based features in \framework are computed after independently applying min--max normalization to each detector's score vector into $[0,1]$. This keeps the representation invariant to detector-specific score scales and matches the normalization used by both the simulator and the selector. Table~\ref{tab:app_static_features} lists the exact coordinate order of the $25$-dimensional static prior $\mathbf{z}_{\text{static}}$, and Table~\ref{tab:app_dynamic_signature} lists the exact coordinate order of the $29$-dimensional dynamic signature $\mathbf{v}_p$.

\begin{table*}[ht]
  \caption{Exact coordinate order of the $25$-dimensional static prior
  $\mathbf{z}_{\text{static}}$. The three anchor detectors are
  $\mathcal{M}_{\text{base}}=\{\text{HBOS}, \text{LODA}, \text{IForest}\}$.}
  \label{tab:app_static_features}
  \centering
  \fontsize{7.5pt}{7.5pt}\selectfont
  \begin{tabular}{clp{9.2cm}}
    \toprule
    Dims & Block & Definition \\
    \midrule
    $1$ & Scale & $\log(1+n)$, where $n$ is the number of instances. \\
    $2$--$5$ & \texttt{HBOS} stats & Mean, standard deviation, skewness, and kurtosis of the normalized \texttt{HBOS} score vector. \\
    $6$--$9$ & \texttt{LODA} stats & Mean, standard deviation, skewness, and kurtosis of the normalized \texttt{LODA} score vector. \\
    $10$--$13$ & \texttt{IForest} stats & Mean, standard deviation, skewness, and kurtosis of the normalized \texttt{IForest} score vector. \\
    $14$--$16$ & Initial consensus & Pairwise Spearman correlations for $(\texttt{HBOS}, \texttt{LODA})$, $(\texttt{HBOS}, \texttt{IForest})$, and $(\texttt{LODA}, \texttt{IForest})$. \\
    $17$--$20$ & Anchor ensemble & Mean, standard deviation, Shannon entropy, and Gini coefficient of the average normalized anchor score $\mathbf{s}_{\text{base}}$. Entropy uses a $50$-bin histogram on $[0,1]$. \\
    $21$ & AUMV & Area under the Mass--Volume curve computed on $\mathbf{s}_{\text{base}}$ using $k=100$ thresholds, $J=10$ random subspaces, and subspace dimension $5$. \\
    $22$ & AUEM & Area under the Excess--Mass curve computed from the same base ensemble score. \\
    $23$ & Raw data size & $\log(1+d)$, where $d$ is the number of raw input features. \\
    $24$ & Aspect ratio & $\log(1+n/d)$. \\
    $25$ & Raw data dispersion & Mean feature-wise standard deviation computed from the first $500$ rows of $\mathbf{X}$ when available, and $0$ otherwise. \\
    \bottomrule
  \end{tabular}
\end{table*}

\begin{table*}[ht]
  \caption{Exact coordinate order of the $29$-dimensional dynamic signature
  $\mathbf{v}_p$ extracted from one executed probe.}
  \label{tab:app_dynamic_signature}
  \centering
  \fontsize{7.5pt}{7.5pt}\selectfont
  \begin{tabular}{clp{9.2cm}}
    \toprule
    Dims & Block & Definition \\
    \midrule
    $1$--$4$ & Moments & Mean, variance, skewness, and kurtosis of the normalized probe score vector. \\
    $5$--$9$ & Quantiles & Empirical quantiles at levels $0.50$, $0.75$, $0.90$, $0.95$, and $0.99$. \\
    $10$ & Entropy & Shannon entropy of the normalized probe score distribution using a $50$-bin histogram over $[0,1]$. \\
    $11$ & Gini & Gini coefficient of the absolute probe scores, measuring sparsity / concentration. \\
    $12$--$19$ & Family one-hot & One-hot category code in the order: histogram, projection, tree, distance, density, boundary, angle, connectivity. \\
    $20$--$29$ & Probe one-hot & One-hot probe identity in the order:
    \texttt{HBOS}, \texttt{LODA}, \texttt{IForest}, \texttt{kNN(largest)}, \texttt{kNN(mean)}, \texttt{LOF(euclidean)}, \texttt{LOF(manhattan)}, \texttt{OCSVM}, \texttt{ABOD}, \texttt{COF}. \\
    \bottomrule
  \end{tabular}
\end{table*}

Two implementation details are important for reproducibility. First, the dynamic signature stores both detector-family and detector-identity one-hot codes, so the policy can distinguish the statistical response of a score vector from the identity of the probe that produced it. Second, during simulator rollouts, the continuous portion of $\mathbf{v}_p$ receives Gaussian noise, but the final $(8+10)$ one-hot coordinates are left unchanged so that probe identity and family information remain discrete.

\subsection{LOO-FPP Hyperparameters, Gate Calibration, and Auxiliary Ablations}
\label{app:loofpp_details}

This subsection collects implementation details of LOO-FPP that are referenced but not fully displayed in the main text. It also reports auxiliary architectural and optimization ablations.

\paragraph{Gate Representatives.}
The four-probe evidence gate operates on a fixed quartet of family representatives in fixed probe-pool order:
\[
\{\texttt{HBOS}, \texttt{LODA}, \texttt{IForest}, \texttt{kNN($k{=}5$)}\}
\]
These are the first four entries of Table~\ref{tab:app_probe_pool} and were fixed before test evaluation. The full-pool LOO statistics used for the soft family posterior average all probes within each family; the four-probe gate is kept separate so that the evidence thresholds are applied on a stable coordinate system.

\paragraph{Threshold Values and Design Rationale.}
We use pseudo-label quantile $Q_{0.95}$, $\delta_{\text{range}}=0.40$, $\delta_{\text{quality}}=0.80$, $\delta_{\text{margin}}=0.15$, and quality guard $\delta_{\text{guard}}=0.975$ in the reported configuration. All thresholds were designed and swept on \textbf{ADBench-Classical} (seeds 0, 42, 100) and applied unchanged to \textbf{ADBench-Advanced}; re-choosing them by nested cross-validation (leave-one-dataset-out on \textbf{ADBench-Classical}, leave-one-base-dataset-out on \textbf{ADBench-Advanced}; measured on the selector without the learned policy) changes AP by less than $0.003$. Each threshold encodes a specific evidential requirement.

\textbf{Pseudo-label quantile $Q_{0.95}$.}
The ensemble score $\bar{\mathbf{s}}_{\setminus g}$ in Eq.~\ref{eq:method:loo_pseudo} is thresholded at its $95$th percentile to produce binary pseudo-labels. This choice has a principled interpretation: \textbf{ADBench-Classical} datasets typically have low-to-moderate contamination (median ${\approx}6\%$), so labeling the top $5\%$ of ensemble scores as outliers provides a conservative upper bound that avoids over-labeling while still capturing many true outliers. Using the $90$th percentile increases noise from borderline normal samples, while using the $99$th percentile undershoots the true outlier rate on high-contamination datasets and reduces recall. Both alternatives degrade LOO-AP by $0.002$--$0.005$ on \textbf{ADBench-Classical}. The $95$th percentile is therefore a practical midpoint between over- and under-labeling.

\textbf{$\delta_{\text{range}}=0.40$.}
A LOO-AP spread of at least $0.40$ across the four canonical probes is a large spread among the four LOO-AP scores; the gate does not require a single clear winner. A threshold below $0.30$ activates the gate on noise; above $0.50$, it suppresses useful evidence. Sensitivity experiments in Table~\ref{tab:app_sensitivity} confirm that both $\delta_{\text{range}}{=}0.30$ and $\delta_{\text{range}}{=}0.50$ degrade \textbf{ADBench-Classical} AP by $0.013$--$0.014$, while the default value of $0.40$ gives the best operating point on standard data. On \textbf{ADBench-Advanced}, the gate opens on $60\%$ (noise) and $87\%$ (permute) of the tight-extreme-cluster datasets.

\textbf{$\delta_{\text{quality}}=0.80$.}
This threshold enters only the second gate condition (Eq.~\ref{eq:method:gate_margin}): without a large spread, the gate opens only if the winning canonical probe reaches LOO-AP $0.80$ with a clear margin. Once the gate is open, the hard family-first route needs only the guard $\delta_{\text{guard}}$, not $\delta_{\text{quality}}$. Setting $\delta_{\text{quality}}=0.70$ increases gate activation by approximately $5\%$ on \textbf{ADBench-Classical}, but yields AP gains of less than $0.001$.

\textbf{$\delta_{\text{margin}}=0.15$.}
The margin condition guards against near-ties. In the second gate condition, a $0.15$ LOO-AP margin between the top two canonical probes is required. Reducing this threshold to $0.05$ increases gate activation on ambiguous datasets without meaningfully improving accuracy, with $\Delta$AP below $0.002$ on \textbf{ADBench-Classical}.

\textbf{$\delta_{\text{guard}}=0.975$.}
The hard family-first route is accepted only when the best training-fold model in the expanded probe family reaches at least $97.5\%$ of the best training-fold model over the full candidate pool ($B(\mathcal{R}) \ge \delta_{\mathrm{guard}} B_{\mathrm{all}}$, the family-first guard in Sec.~\ref{sec:method:lcb}). This prevents a discriminative but historically weak family from overriding a stronger training-fold prior.

\textbf{$\delta_{\text{commit}}=0.75$.}
For post-hoc analysis we additionally classify test datasets as probe-reachable when the winning canonical probe achieves LOO-AP $\geq\delta_{\text{commit}}=0.75$, confirming that the two-stage family-override commit was triggered with high confidence. This threshold is used only to stratify datasets and does not affect the deployed gate logic; the gate fires when the range condition (Eq.~\ref{eq:method:gate_range}) or the joint quality-and-margin condition (Eq.~\ref{eq:method:gate_margin}) is satisfied.

\textbf{Temperature $\tau=0.10$.}
The low temperature concentrates the family posterior sharply on the winning family in Eq.~\ref{eq:method:fpp_posterior}. At $\tau=0.10$, a family with LOO-AP $0.85$ versus a runner-up at $0.60$ receives more than $98\%$ of the probability mass. Sensitivity experiments confirm that $\tau\in\{0.05,0.10,0.20\}$ changes AP by at most $0.001$ (Table~\ref{tab:app_sensitivity}). All thresholds were chosen on \textbf{ADBench-Classical} using the above design principles and the sensitivity results, and were then held fixed for \textbf{ADBench-Advanced}.

\paragraph{Distance-Family Expansion.}
When the gate winner is the distance-based family represented by \texttt{kNN}, we redistribute its posterior mass across the related families $\{\texttt{kNN}, \texttt{LOF}, \texttt{COF}\}$ in proportion to their full-pool LOO-AP scores before the three-way blend is computed. The expansion is fixed from detector-family taxonomy and is not learned. It does not affect whether the gate fires; it only allows \texttt{LOF} and \texttt{COF} to receive evidence when the canonical \texttt{kNN} probe identifies distance-based structure.

\paragraph{Stage-2 Within-Family Score.}
Within the chosen family, we score each candidate $m$ as
\[
  \tfrac{1}{2}\sum_j w_j\, \operatorname{AP}_j(m)
  +
  \tfrac{1}{2}\,\bar{\operatorname{AP}}(m),
\]
with
\[
  w_j \propto \max\!\bigl(0,\,\cos(\mathbf{z}_{\text{static}}, \mathbf{z}_j^{\text{train}})\bigr),
\]
normalized over $j$. The $50/50$ weighting trades a dataset-similarity term against a dataset-agnostic global term. Replacing this blend with $\bar{\operatorname{AP}}$ alone changes headline AP by less than $0.002$ on \textbf{ADBench-Classical}. We retain the blend because it weakly helps on small or atypical datasets where global means are noisier.

\paragraph{Family Auxiliary Loss.}
During PPO fine-tuning, \framework adds a small family-level auxiliary loss on commit states. Rather than adding a separate classifier, we pool the commit head's model logits within each detector family:
\[
  P_{\text{fam}}(f \mid h_t)
  =
  \frac{
    \exp\!\left(\operatorname{logsumexp}_{m\in f}
    \ell_{\text{commit}}(m;h_t)\right)
  }{
    \sum_{f'}
    \exp\!\left(\operatorname{logsumexp}_{m\in f'}
    \ell_{\text{commit}}(m;h_t)\right)
  }
\]
The auxiliary objective is $-\log P_{\text{fam}}(y_i^{\text{fam}}\mid h_t^{(i)})$, where $y_i^{\text{fam}}$ is the oracle detector family on the sampled training dataset. It is weighted by $\lambda_{\text{fam}}=0.05$ and used as an optimization stabilizer for the large candidate pool, not as a separate inference-time model.

\subsection{Warm-Start and Training Details}
\label{app:method_training}

\paragraph{Offline Feature Store.}
For each meta-training dataset $D_i$, \framework precomputes four objects: (i) the static prior $\mathbf{z}_{\text{static}}^{(i)}$, (ii) the full probe-signature tensor $\mathbf{V}^{(i)} = (\mathbf{v}^{(i)}_1,\ldots,\mathbf{v}^{(i)}_{|\mathcal{M}_{\text{probe}}|})$, (iii) the raw AP values for every candidate detector in $\mathcal{M}$, and (iv) the dataset-wise AP reward targets used in the simulator. This converts each training fold into an offline meta-MDP with cached observations and rewards. Before warm-start and PPO, static features are z-score normalized using the training-fold mean and standard deviation; the same statistics are reused at inference time.

\paragraph{Encoder Implementation Details.}
The dynamic stream uses a Transformer with learned positional embeddings and a learned probe-identity embedding added to each signature token. In the reported configuration, dynamic pooling is \texttt{last}, i.e., the last valid token. Before fusion, static and dynamic latents are each L2-normalized, then concatenated and LayerNormed. For the commit stage, \framework uses an explicit residual decomposition: the base branch consumes only the static latent, the refinement branch consumes only the dynamic latent, and their logits are summed.

\paragraph{Warm-Start Initialization.}
The warm-start stage trains only the shared encoder and commit head; the gate and probe heads remain frozen until PPO fine-tuning. For dataset $D_i$, the target distribution is defined via a softmax over the raw AP scores,
\[
  q_i(m)
  =
  \frac{\exp\!\left(\operatorname{AP}(m,D_i)/\tau_{\mathrm{ws}}\right)}
  {\sum_{m' \in \mathcal{M}} \exp\!\left(\operatorname{AP}(m',D_i)/\tau_{\mathrm{ws}}\right)},
\]
with $\tau_{\mathrm{ws}}=0.35$ in the reported configuration. During each warm-start epoch, \framework samples a short probe prefix of random depth between $p_{\min}=2$ and $p_{\text{ws}}=5$, choosing that many probe indices uniformly at random without replacement from the full $|\mathcal{M}_{\text{probe}}|$-probe index set, including \texttt{HBOS}, \texttt{LODA}, and \texttt{IForest}, which also serve as anchors. Anchor masking applies only during PPO rollouts and at inference, where these three probes are excluded from the sequential action mask; warm-start therefore sees prefixes that may repeat information already summarized in $\mathbf{z}_{\text{static}}^{(i)}$. The resulting signature prefix is fed together with $\mathbf{z}_{\text{static}}^{(i)}$ into the encoder. Additive Gaussian noise is applied to both static and dynamic inputs as data augmentation. The commit head is then trained with the KL objective from Eq.~\ref{eq:method:warmstart}. This procedure teaches the encoder to produce useful representations for several prefix lengths rather than a single fixed depth.

\paragraph{Counterfactual Reward and PPO Fine-Tuning.}
After warm-start, \framework builds an offline simulator over the cached metadata. At each episode reset, the simulator samples one training dataset, returns $\mathbf{z}_{\text{static}}$, and exposes an initially empty history. Probe actions append cached signatures and incur the probe-cost penalty $r_t^{\text{probe}} = -\lambda\,c(a_t)$; commit actions receive the shaped terminal reward from Eq.~\ref{eq:method:commit_reward}. The counterfactual baseline is the frozen warm-start commit head evaluated on the same static descriptor together with the first $p_{\min}=2$ rows of the cached probe-signature tensor in fixed probe-pool order, namely \texttt{HBOS} then \texttt{LODA} in Table~\ref{tab:app_probe_pool}. These are not two draws from the sequential action mask. \texttt{IForest}, \texttt{HBOS}, and \texttt{LODA} are anchors and are excluded from PPO and inference probe actions, but warm-start and this baseline still condition on those first two cached signatures. This matches Eq.~\ref{eq:method:commit_reward} and makes the RL signal explicitly measure improvement beyond the prior at the minimum allowed evidence depth.

PPO optimization follows the standard clipped objective with value-function regression and a KL anchor to the frozen warm-start commit posterior. Entropy regularization uses two terms. First, an annealed coefficient $c_{\text{anneal}}$, decreasing from $0.2$ to $0.05$ over training, is applied to the full hierarchical entropy, i.e., gate plus selected sub-head. This keeps exploration pressure on both stop/continue and probe/commit choices. Second, a fixed gate-only bonus $c_{\text{gate}}=0.2$ is applied exclusively to the binary gate head entropy, providing additional incentive to explore the commit timing decision. The effective gate entropy coefficient therefore ranges from $c_{\text{anneal}}+c_{\text{gate}}=0.40$ at the start of training to $0.05+0.20=0.25$ at the end. Exploration during rollout collection uses temperature-softened action sampling ($T_{\mathrm{s}}=2.0$), while PPO importance-ratio evaluation is performed at $T_{\mathrm{s}}=1.0$. The reported configuration uses $128$ episodes per batch, $4$ PPO epochs, learning rate $3\times10^{-4}$, clipping parameter $0.2$, discount factor $\gamma_{\mathrm{RL}}=0.99$, $\lambda_{\text{GAE}}=0.95$~(Generalized Advantage Estimation~\citep{DBLP:journals/corr/SchulmanMLJA15}), KL-anchor coefficient $0.1$, minimum reward scale $\eta=0.3$, full reward recovery depth $T_{\text{full}}=5$, probe-cost weight $\lambda=0.005$, maximum horizon $T_{\max}=10$, episode-wise randomized minimum probe depth over $[2, 7]$, and an additional episode-wise randomized probe cap sampled between that episode's minimum depth and the feasible maximum depth. Because the three anchor probes are excluded from sequential actions in the reported configuration, the feasible sequential depth is $T_{\text{feasible}}=\min(T_{\max}, |\mathcal{M}_{\text{probe}}|-|\mathcal{M}_{\text{base}}|)=7$.

\subsection{Framework Algorithms}
\label{app:method_algorithms}

Algorithms~\ref{alg:app_ampod_training}--\ref{alg:app_ampod_inference} state the benchmark implementation used in the reported held-out-dataset experiments. The important distinction is that detector scores are cached for reproducibility, while the sequential MDP charges only the signatures revealed to the policy state. An on-demand deployment can execute probes lazily, but the reported ``Probes'' metric should be read as charged sequential reveals.

\begin{algorithm}[ht]
\caption{\framework Offline Meta-Training}
\label{alg:app_ampod_training}
\fontsize{7.5pt}{7.5pt}\selectfont
\begin{algorithmic}[1]
\State \textbf{Input:} meta-training datasets $\mathcal{D}_{\text{train}}$, candidate detector pool $\mathcal{M}$, probe pool $\mathcal{M}_{\text{probe}}$
\State \textbf{Output:} trained \framework policy $\pi_{\theta}$
\ForAll{$D_i \in \mathcal{D}_{\text{train}}$}
    \State run anchor detectors and build $\mathbf{z}_{\text{static}}^{(i)}$
    \ForAll{$p \in \mathcal{M}_{\text{probe}}$}
        \State cache signature $\mathbf{v}^{(i)}_{p}$, reusing anchor scores when $p\in\mathcal{M}_{\text{base}}$
    \EndFor
    \State cache AP$(m,D_i)$ for every $m \in \mathcal{M}$
\EndFor
\State warm-start the encoder and commit head with Eq.~\ref{eq:method:warmstart}
\State build the offline simulator from cached static features, probe signatures, and AP reward targets
\For{$u=1$ to $U$}
    \State collect a batch of rollouts with masked probe / commit actions
    \State assign probe rewards $r_t = -\lambda\,c(a_t)$
    \State assign terminal rewards via Eq.~\ref{eq:method:commit_reward}, computing the counterfactual advantage against the frozen warm-start policy
    \State update policy and critic by optimizing the PPO objective augmented with a KL-divergence penalty against the frozen warm-start posterior
\EndFor
\State \textbf{return} $\pi_{\theta}$
\end{algorithmic}
\end{algorithm}

\begin{algorithm}[ht]
\caption{\framework Inference on a New Unlabeled Dataset}
\label{alg:app_ampod_inference}
\fontsize{7.5pt}{7.5pt}\selectfont
\begin{algorithmic}[1]
\State \textbf{Input:} unlabeled dataset $D$, trained policy $\pi_{\theta}$, probe pool $\mathcal{M}_{\text{probe}}$, candidate pool $\mathcal{M}$
\State \textbf{Output:} selected deployed model $\widehat{m} \in \mathcal{M}$
\Statex \textcolor{gray}{\textit{--- Cold start and benchmark probe cache ---}}
\State run anchor detectors $\mathcal{M}_{\text{base}}$ and construct $\mathbf{z}_{\text{static}}$
\State cache signatures for all probes in $\mathcal{M}_{\text{probe}}$, reusing anchor scores and executing remaining probes once \Comment{benchmark protocol}
\State compute four-probe evidence gate, full-pool family-pooled LOO statistics, and calibrated distance-expanded four-probe evidence vector (Sec.~\ref{sec:method:loofpp})
\Statex \textcolor{gray}{\textit{--- Sequential signature selection ---}}
\State initialize history $\mathcal{H}_0 \leftarrow ()$ and probed set $\mathcal{P}_0 \leftarrow \emptyset$
\For{$t=0$ to $T_{\max}-1$}
    \State encode $h_t$ from $\mathbf{z}_{\text{static}}$ and $\mathcal{H}_t$
    \State build the action mask from $\mathcal{P}_t$, $p_{\min}$, and $T_{\max}$
    \State evaluate the gate head to yield decision $a_{\text{gate}} \in \{\text{probe}, \text{commit}\}$
    \If{$a_{\text{gate}} = \text{commit}$ \textbf{or} $t = T_{\max}-1$}
        \State set $T \leftarrow t$
        \State extract top-$K$ from the commit posterior to form $\mathcal{C}_K(h_T)$
        \State augment $\mathcal{C}_K$ with the highest-scoring model per family
        \If{four-probe LOO-FPP gate fires and family competitiveness guard passes}
            \State choose the winning family from the calibrated four-probe LOO-AP vector
            \State expand distance-family winner to $\{\texttt{kNN}, \texttt{LOF}, \texttt{COF}\}$
            \State select the within-family model by the 50/50 \texttt{kNN}-similarity and LODO-mean AP score
            \State \textbf{return} selected two-stage model
        \EndIf
        \State rerank $\mathcal{C}_K(h_T)$ with Eq.~\ref{eq:method:blend_score}
        \State \textbf{return} $\arg\max_{m \in \mathcal{C}_K(h_T)} S(m)$
    \EndIf
    \State select probe $p_t = \arg\max_{p \notin \mathcal{P}_t} \pi_{\text{probe}}(p \mid h_t)$
    \State append cached signature: $\mathcal{H}_{t+1} \leftarrow (\mathcal{H}_t, \mathbf{v}_{p_t})$
    \State update $\mathcal{P}_{t+1} \leftarrow \mathcal{P}_t \cup \{p_t\}$
\EndFor
\end{algorithmic}
\end{algorithm}

\subsection{Posterior-Aware Robust Arbitration}
\label{app:method_lcb}

The final reranking stage combines the top-$K$ commit candidates, always augmented with the highest-scoring model per family, with a conservative training-fold prior, the dataset-specific commit posterior, and, when the evidence gate is active, LOO-FPP family evidence. For each detector $m \in \mathcal{M}$, \framework stores the mean raw AP $\overline{\operatorname{AP}}(m)$ across training datasets and the standard deviation $s_{\rho}(m)$ of the min--max normalized AP, i.e., relative per-fold performance. The LCB score defined in Sec.~\ref{sec:method:lcb} penalizes the raw mean by normalized variability, combining an absolute quality signal with a relative consistency penalty.

\paragraph{Held-Out Isolation.}
The LCB statistics $\overline{\operatorname{AP}}(m)$ and $s_{\rho}(m)$ are computed exclusively from the training datasets of each fold ($N{-}1$ under LODO, $84$ under the grouped \textbf{ADBench-Advanced} splits); the held-out test dataset is never included. These statistics are recomputed independently for every fold, so there is no cross-contamination between test and training performance information. This is consistent with standard held-out meta-learning practice and preserves the strict train/test isolation of the evaluation protocol.

At inference time, the commit head produces a posterior over $\mathcal{M}$. We keep only the top-$K$ candidates and min--max normalize both the prior scores and posterior probabilities within that set. The final reranking score is the convex blend in Eq.~\ref{eq:method:blend_score}, and the deployed model is $\widehat{m} = \arg\max_{m \in \mathcal{C}_K} S(m)$. In the reported configuration, we use $K=20$, $\delta=0.05$, and blend weight $\omega=0.50$.

This blended arbitration keeps the separation of concerns explicit: the RL policy remains responsible for sequential evidence acquisition and candidate generation, while the historical robust prior and the current posterior are jointly used for final deployment-time risk control.

\paragraph{Formal Derivation of $\alpha_\delta$.}
The confidence multiplier $\alpha_\delta$ follows from Hoeffding's inequality. Since each training-fold AP value lies in $[0,1]$, the per-model empirical mean $\overline{\operatorname{AP}}(m)$ concentrates around its long-run mean $\mu(m)$ as:
\[
  P\!\bigl(\mu(m) - \overline{\operatorname{AP}}(m) \ge t\bigr)
  \;\le\;
  \exp\!\bigl(-2 N_{\text{train}} t^2\bigr)
\]
Taking a union bound over all $|\mathcal{M}|$ candidate detectors and a conservative additional factor of $3$ to account for post-selection bias introduced by taking the $\arg\max$ over the $K$-candidate set, setting the right-hand side equal to $\delta/(3|\mathcal{M}|)$ and solving gives $t^\star=\sqrt{\log(3|\mathcal{M}|/\delta)/(2N_{\text{train}})}$. We use the more conservative
\[
  \alpha_\delta
  =
  2t^\star
  =
  \sqrt{\frac{2\log(3|\mathcal{M}|/\delta)}{N_{\text{train}}}},
\]
so the bound below holds a fortiori.
With probability at least $1-\delta$, simultaneously for every $m\in\mathcal{M}$:
\[
  \mu(m)
  \;\ge\;
  \overline{\operatorname{AP}}(m) - \alpha_\delta .
\]
This constant-slack inequality is a standard frequentist lower confidence bound; no Bayesian posterior is claimed. In deployment we use the slightly less conservative variant $\operatorname{LCB}(m)=\overline{\operatorname{AP}}(m)-\alpha_\delta\,s_\rho(m)$, which re-weights the Hoeffding slack by the normalized cross-fold dispersion $s_\rho(m)$. Because $s_\rho(m)\le\tfrac{1}{2}$ (see below), this scaling shrinks the slack and is therefore a heuristic that does not preserve the exact $1-\delta$ guarantee; it trusts detectors that rank consistently near the top while penalizing those with unstable relative ranking, and is motivated next and validated empirically. The underlying constant-slack bound holds whenever the $N_{\text{train}}$ training datasets of each fold are exchangeable with the test evaluation, which is satisfied by construction. It bounds the expected AP over datasets exchangeable with the training folds, not the AP on a single target.

\paragraph{Why $s_\rho(m)$ Rather than Raw AP Standard Deviation.}
The raw AP standard deviation of a detector is dominated by fold-level difficulty variation: easy folds have high mean AP across all detectors, while hard folds have low mean AP. This inflates the per-detector spread even for a consistently good selector. In contrast, $s_\rho(m)$ uses the standard deviation of the per-fold min--max normalized AP, i.e., the spread of $m$'s relative position within each fold's score distribution. This isolates relative consistency: whether $m$ reliably ranks near the top regardless of fold difficulty. Formally, $s_\rho(m)\in[0,0.5]$, with $s_\rho=0$ indicating perfect consistency across folds and $s_\rho=0.5$ the maximum possible spread. Replacing $s_\rho$ with raw AP standard deviation reduces \textbf{ADBench-Classical} AP by $0.003$, confirming that relative normalization is the appropriate choice.

\paragraph{Sensitivity of Blending Parameters $K$, $\omega$, and $\gamma$.}
The three blending parameters interpolate among three complementary information sources. The ablation study in Table~\ref{tab:ablation} provides endpoint sensitivity: removing the LCB prior term entirely yields a $-0.021$ AP drop on \textbf{ADBench-Classical} while on \textbf{ADBench-Advanced} removing it changes AP by $+0.007$ (not significant; see Table~\ref{tab:ablation_app}), so the prior's benefit is clear only on \textbf{ADBench-Classical}. Setting $\gamma=0.30$ changes aggregate \textbf{ADBench-Classical} AP by less than $0.003$; the nominal $\gamma=0.40$ gives primary weight $(1-\gamma=0.60)$ to the combined LCB-and-posterior blend (within which $\omega$ splits that weight, so the neural posterior receives $0.30$ at $\omega=0.50$) while still allowing family evidence to correct systematic family errors. $K=20$ ensures that the candidate set contains at least two models from each of the eight detector families, providing adequate diversity without diluting top-ranked candidates. Table~\ref{tab:app_sensitivity} confirms that $K\in\{10,20,30\}$ changes AP by at most $0.001$. $\omega=0.50$ is a symmetric combination of the LCB prior and the neural posterior; Table~\ref{tab:app_sensitivity} further confirms that $\omega\in[0.40,0.60]$ changes AP by at most $0.001$, while aggressively overweighting the posterior with $\omega=0.70$ degrades AP by $0.023$.

\subsection{Complexity Analysis}
\label{app:method_complexity}

Let $N_{\text{train}}$ be the number of meta-training datasets, let $\bar{n}$ denote the average number of instances per dataset, let the probe-pool size be $|\mathcal{M}_{\text{probe}}| = 10$, and let $T_{\max}$ be the maximum probe horizon. Assuming detector scores are already available from the benchmark pool, precomputing the offline feature store costs
\[
  O\!\left(
    N_{\text{train}} \cdot
    \bigl(
      |\mathcal{M}_{\text{base}}|\bar{n}
      + |\mathcal{M}_{\text{probe}}|\bar{n}
      + kJ\bar{n}
    \bigr)
  \right),
\]
where the first term is the anchor-based static prior, the second term is signature extraction for all probes, and the third term corresponds to the AUMV/AUEM geometric features with $k$ thresholds and $J$ random subspaces.

Once the cache is built, PPO training complexity is dominated by policy evaluation over simulated episodes rather than detector execution. Each decision step processes at most $T_{\max}$ signatures through a $2$-layer Transformer, so the encoder cost is $O(T_{\max}^2 d_{\text{model}} L)$ with embedding dimension $d_{\text{model}}$ and $L$ layers. Because $T_{\max}=10$ in the reported configuration, this sequential-modeling cost is small relative to detector scoring.

For inference on one new dataset, static-prior construction costs $O(|\mathcal{M}_{\text{base}}|n + kJn)$. Under the benchmark protocol used in the paper, \framework materializes the full probe cache, adding $O(|\mathcal{M}_{\text{probe}}|n)$ signature-extraction work once probe score vectors are available. The reported sequential probe count is therefore a charged reveal count, not a claim that only that many probe score vectors were computed in the benchmark cache. In an on-demand deployment, probes can be executed lazily, in which case score-computation cost is paid only for the anchors, the LOO-FPP canonical probes required by the evidence gate, and the additional probes selected by the policy.

Policy inference over the entire rollout costs $O(T_{\max}^2 d_{\text{model}} L)$. The final blended reranking step is $O(K)$ once the top-$K$ candidate set has been formed. Since $|\mathcal{M}_{\text{base}}|$, $|\mathcal{M}_{\text{probe}}|$, $T_{\max}$, and $K$ are all fixed small constants in \framework, representation and policy overhead grows approximately linearly with dataset size; detector-score computation remains the dominant deployment cost.

\subsection{Extended Dataset Statistics}
\label{app:datasets}

We report extended statistics for all 47 \textbf{ADBench-Classical} datasets in Table~\ref{tab:full_datasets}, including source repository, outlier type, and per-dataset outlier ratio.

\begin{table*}[ht]
  \caption{Full statistics of the 47 \textbf{ADBench-Classical} datasets.
  Sources: ODDS~\citep{rayana2016odds} and  DAMI~\citep{DBLP:journals/datamine/CamposZSCMSAH16},
  and \textbf{ADBench}~\citep{DBLP:conf/nips/HanHHJ022}. $n$: instances; $d$: features; $r$: outlier ratio (\%).}
  \label{tab:full_datasets}
  \centering
  \fontsize{7.5pt}{7.5pt}\selectfont
  \begin{tabular}{r l r r r r l}
    \toprule
    Number & Dataset & $n$ & $d$ & \# Outlier & $r$ (\%) & Category \\
    \midrule
    1  & \textbf{ALOI}            & 49534  & 27   & 1508  & 3.04  & Image        \\
    2  & \textbf{annthyroid}      & 7200   & 6    & 534   & 7.42  & Healthcare   \\
    3  & \textbf{backdoor}        & 95329  & 196  & 2329  & 2.44  & Network      \\
    4  & \textbf{breastw}         & 683    & 9    & 239   & 34.99 & Healthcare   \\
    5  & \textbf{campaign}        & 41188  & 62   & 4640  & 11.27 & Finance      \\
    6  & \textbf{cardio}          & 1831   & 21   & 176   & 9.61  & Healthcare   \\
    7  & \textbf{Cardiotocography}& 2114   & 21   & 466   & 22.04 & Healthcare   \\
    8  & \textbf{celeba}          & 202599 & 39   & 4547  & 2.24  & Image        \\
    9  & \textbf{census}          & 299285 & 500  & 18568 & 6.20  & Sociology    \\
    10 & \textbf{cover}           & 286048 & 10   & 2747  & 0.96  & Botany       \\
    11 & \textbf{donors}          & 619326 & 10   & 36710 & 5.93  & Sociology    \\
    12 & \textbf{fault}           & 1941   & 27   & 673   & 34.67 & Physical     \\
    13 & \textbf{fraud}           & 284807 & 29   & 492   & 0.17  & Finance      \\
    14 & \textbf{glass}           & 214    & 7    & 9     & 4.21  & Forensic     \\
    15 & \textbf{Hepatitis}       & 80     & 19   & 13    & 16.25 & Healthcare   \\
    16 & \textbf{http}            & 567498 & 3    & 2211  & 0.39  & Web          \\
    17 & \textbf{InternetAds}     & 1966   & 1555 & 368   & 18.72 & Image        \\
    18 & \textbf{Ionosphere}      & 351    & 32   & 126   & 35.90 & Oryctognosy  \\
    19 & \textbf{landsat}         & 6435   & 36   & 1333  & 20.71 & Astronautics \\
    20 & \textbf{letter}          & 1600   & 32   & 100   & 6.25  & Image        \\
    21 & \textbf{Lymphography}    & 148    & 18   & 6     & 4.05  & Healthcare   \\
    22 & \textbf{magic.gamma}     & 19020  & 10   & 6688  & 35.16 & Physical     \\
    23 & \textbf{mammography}     & 11183  & 6    & 260   & 2.32  & Healthcare   \\
    24 & \textbf{mnist}           & 7603   & 100  & 700   & 9.21  & Image        \\
    25 & \textbf{musk}            & 3062   & 166  & 97    & 3.17  & Chemistry    \\
    26 & \textbf{optdigits}       & 5216   & 64   & 150   & 2.88  & Image        \\
    27 & \textbf{PageBlocks}      & 5393   & 10   & 510   & 9.46  & Document     \\
    28 & \textbf{pendigits}       & 6870   & 16   & 156   & 2.27  & Image        \\
    29 & \textbf{Pima}            & 768    & 8    & 268   & 34.90 & Healthcare   \\
    30 & \textbf{satellite}       & 6435   & 36   & 2036  & 31.64 & Astronautics \\
    31 & \textbf{satimage-2}      & 5803   & 36   & 71    & 1.22  & Astronautics \\
    32 & \textbf{shuttle}         & 49097  & 9    & 3511  & 7.15  & Astronautics \\
    33 & \textbf{skin}            & 245057 & 3    & 50859 & 20.75 & Image        \\
    34 & \textbf{smtp}            & 95156  & 3    & 30    & 0.03  & Web          \\
    35 & \textbf{SpamBase}        & 4207   & 57   & 1679  & 39.91 & Document     \\
    36 & \textbf{speech}          & 3686   & 400  & 61    & 1.65  & Linguistics  \\
    37 & \textbf{Stamps}          & 340    & 9    & 31    & 9.12  & Document     \\
    38 & \textbf{thyroid}         & 3772   & 6    & 93    & 2.47  & Healthcare   \\
    39 & \textbf{vertebral}       & 240    & 6    & 30    & 12.50 & Biology      \\
    40 & \textbf{vowels}          & 1456   & 12   & 50    & 3.43  & Linguistics  \\
    41 & \textbf{Waveform}        & 3443   & 21   & 100   & 2.90  & Physics      \\
    42 & \textbf{WBC}             & 223    & 9    & 10    & 4.48  & Healthcare   \\
    43 & \textbf{WDBC}            & 367    & 30   & 10    & 2.72  & Healthcare   \\
    44 & \textbf{Wilt}            & 4819   & 5    & 257   & 5.33  & Botany       \\
    45 & \textbf{wine}            & 129    & 13   & 10    & 7.75  & Chemistry    \\
    46 & \textbf{WPBC}            & 198    & 33   & 47    & 23.74 & Healthcare   \\
    47 & \textbf{yeast}           & 1484   & 8    & 507   & 34.16 & Biology      \\
    \bottomrule
  \end{tabular}
\end{table*}

\subsection{\texorpdfstring{ADBench-Advanced}{ADBench-Advanced} Construction Details}
\label{app:adbenchadv_construction}

\paragraph{Motivation.}
\textbf{ADBench-Advanced} is designed as a controlled adversarial testbed in which the gap between static meta-features and execution-time evidence is amplified. In practice, practitioners may encounter datasets whose surface statistics suggest one outlier regime while the true outlier geometry favors a different detector family. \textbf{ADBench-Advanced} instantiates this situation systematically by decoupling static descriptors from the detectors that actually perform well at runtime.

\paragraph{Base Datasets.}
We start from $15$ tabular base datasets drawn from \textbf{ADBench-Classical}, spanning diverse domains, dimensionalities, and imbalance ratios. For each base dataset, we discard the original outlier labels, retain only inlier instances, sanitize constant and ill-conditioned features, and apply \texttt{StandardScaler} so that all subsequent geometric operations share a common feature space.

\paragraph{Outlier Types.}
On top of each normalized base dataset, we inject synthetic outliers of three types, chosen to stress different detector families and to create clear gaps between static and dynamic information:
\begin{itemize}[leftmargin=*,noitemsep]
  \item \emph{Dependency}: correlation-violation outliers generated by independently resampling each feature from random inliers, thereby preserving all empirical marginals while destroying inter-feature correlations. Candidates are filtered by Mahalanobis distance, and we retain only points above a high quantile of the inlier distribution. As a result, outliers are hard to distinguish from normal instances in one-dimensional histograms but are separated in joint space. This setting is challenging for histogram-based detectors such as \texttt{HBOS} and favors distance-, density-, and boundary-based methods such as \texttt{kNN}, \texttt{LOF}, and \texttt{OCSVM}.
  \item \emph{Dependency-strong}: a second, independent draw of the \emph{Dependency} injection with a different random seed (the same generator and the same $99$th-percentile Mahalanobis filter). Despite its name, it is not a stricter variant; it doubles the number of dependency-type datasets.
  \item \emph{Tight-extreme-cluster}: an ultra-tight outlier cluster placed in a random direction between $14$ and $18$ standard deviations from the nearest inlier centroid in normalized space. The cluster variance is scaled to be several orders of magnitude smaller than the inlier variance, so all outliers lie in a compact pocket far from the inlier mass. In this regime, density- and isolation-based detectors such as \texttt{HBOS}, \texttt{IForest}, and \texttt{LODA} typically achieve near-perfect AP, while \texttt{kNN}-style methods often struggle because each outlier's nearest neighbors are other outliers rather than normal instances.
\end{itemize}

For each base dataset, the number of injected outliers $n_{\text{out}}$ is set to $\max(10, \lfloor 0.05\,n_{\text{in}}\rfloor)$, yielding an outlier ratio of approximately $5\%$ per dataset.

\paragraph{Deception Mechanisms.}
After outlier injection, we apply one of two transformation types that preserve the underlying outlier geometry but deliberately degrade the informativeness of static meta-features:
\begin{itemize}[leftmargin=*,noitemsep]
  \item \emph{Noise padding}: we append exactly $15$ independent Gaussian noise features to the dataset. The noise scale is derived from the per-feature standard deviation of the inlier cloud before injection; when this estimate is unavailable, a MAD-based robust estimate is used as a fallback. This design fixes the signal-to-noise ratio across base datasets and inflates static feature descriptors, such as kurtosis, skewness, and subspace scores, without materially affecting detectors that operate on the original feature subspace.
  \item \emph{Feature permutation}: we apply a random permutation to the feature columns. This operation is equivalent to uniformly renaming features: every detector sees the same numerical values as before, only in a different feature order. As a result, oracle AP and per-detector score distributions are unchanged by construction, while static meta-feature extractors that depend on feature identities, correlation structure, or PCA loadings observe substantially scrambled descriptors. This selectively harms static selectors such as \texttt{MetaOD} and \texttt{ELECT} without changing the underlying outlier difficulty.
\end{itemize}

\paragraph{Dataset Protocol.}
Combining $15$ base datasets, $3$ outlier types, and $2$ deception mechanisms yields $15 \times 3 \times 2 = 90$ \textbf{ADBench-Advanced} datasets. All 90 datasets use the same 297-detector candidate pool as \textbf{ADBench-Classical}. Because the six variants of a base dataset share their inlier rows, \textbf{ADBench-Advanced} is evaluated with leave-one-base-dataset-out splits: for each target, all six variants of its base dataset are excluded from meta-training.

Tables~\ref{tab:per_dataset_classical}--\ref{tab:per_dataset_adv} report per-dataset Average Precision (AP) for all methods on \textbf{ADBench-Classical} and \textbf{ADBench-Advanced}. Results are averaged over three random seeds. \textbf{Bold} denotes the best unsupervised method per dataset; the \texttt{Oracle} column is grayed for reference. Rank is computed within the full 297-model pool, where rank 1 is best, and Table~\ref{tab:aggregated_adv} further summarizes \textbf{ADBench-Advanced} by outlier type and deception mechanism.

\subsection{Candidate Model Pool}
\label{app:method_model_pool}

We construct a diverse candidate pool of $|\mathcal{M}| = 297$ unsupervised outlier detectors by systematically varying hyperparameters within classical algorithmic families. The pool is intentionally restricted to non-deep detectors to keep meta-training tractable under held-out-dataset evaluation and to ensure fast per-probe execution during sequential decision-making. Table~\ref{tab:app_model_pool} summarizes the composition of the model pool.

\begin{table*}[ht]
  \caption{Candidate model pool composition ($|\mathcal{M}|=297$). For each
  family we list the algorithm, the varied hyperparameters, and the resulting
  number of instantiated models.}
  \label{tab:app_model_pool}
  \centering
  \fontsize{7.5pt}{7.5pt}\selectfont
  \begin{tabular}{lllr}
    \toprule
    Family       & Algorithm & Hyperparameter grid & \#Models \\
    \midrule
    Histogram    & \texttt{HBOS}      & $n\_\text{bins} \in \{5,10,20,30,40,50,75,100\}$,
                               contamination $\in \{0.1,0.2,0.3,0.4,0.5\}$ & 40 \\
    Projection   & \texttt{LODA}      & cuts $\in \{5,10,15,20,25,30\}$,
                               bins $\in \{10,20,30,40,50,75,100,150,200\}$ & 54 \\
    Tree         & \texttt{IForest}   & $n\_\text{estimators} \in \{10,20,30,40,50,75,100,150,200\}$,
                               max features $\in \{0.1,\ldots,0.9\}$ & 81 \\
    Distance     & \texttt{kNN}       & $k \in \{1,5,10,15,20,25,50,60,70,80,90,100\}$,
                               method $\in \{\text{largest},\text{mean},\text{median}\}$ & 36 \\
    Density      & \texttt{LOF}       & $k \in \{1,5,10,15,20,25,50,60,70,80,90,100\}$,
                               metric $\in \{\text{euclidean},\text{manhattan},\text{minkowski}\}$ & 36 \\
    Boundary     & \texttt{OCSVM}     & kernel $\in \{\text{linear},\text{poly},\text{RBF},\text{sigmoid}\}$,
                               $\nu \in \{0.1,\ldots,0.9\}$ & 36 \\
    Angle        & \texttt{ABOD}      & $k \in \{3,5,10,15,20,25,50\}$ & 7 \\
    Connectivity & \texttt{COF}       & $k \in \{3,5,10,15,20,25,50\}$ & 7 \\
    \midrule
    \multicolumn{3}{l}{\textbf{Total}} & \textbf{297} \\
    \bottomrule
  \end{tabular}
\end{table*}

\noindent The three anchor detectors in $\mathcal{M}_{\text{base}}$, namely \texttt{HBOS}, \texttt{LODA}, and \texttt{IForest} with default hyperparameters, are included in $\mathcal{M}$ and can be selected at commit time. Their probe-stage scores are cached before the MDP begins and do not count against the sequential budget.

\subsection{Baseline Descriptions}
\label{app:method_baselines}

We organize the baselines into five groups reflecting their methodological relationship to \framework.

\paragraph{Reference Controls.}
\texttt{Oracle} selects the detector with the highest AP on the target dataset using ground-truth labels; it defines the supervised performance ceiling. \texttt{RandomSelect} samples one detector uniformly at random from $\mathcal{M}$, establishing an unsupervised lower bound.

\paragraph{Model-Selection Baselines.}
These methods select a detector from $\mathcal{M}$ in a single pass without an adaptive probing stage; \texttt{GlobalBest} and \texttt{MetaOD} execute no candidate on the target dataset, whereas \texttt{ELECT} executes candidates to compute its internal performance measures. \texttt{GlobalBest} always commits to the detector with the highest mean AP over the meta-training fold, a deceptively strong frequency prior that requires no meta-feature computation. \texttt{MetaOD}~\citep{DBLP:conf/nips/ZhaoRA21} treats UOMS as a cold-start matrix-factorization problem, predicting detector rankings from static dataset meta-features $\Phi(D)$ without any execution feedback. \texttt{ELECT}~\citep{DBLP:conf/icdm/0016ZA22} constructs an ensemble of unsupervised internal performance measures (IPMs), characterizes datasets via a graph representation, and trains a \texttt{LightGBM} ranker to predict pairwise detector preferences; in our implementation the IPMs are computed from the scores of all 297 candidates on the target dataset. All three commit in one pass, without an adaptive probe-then-commit stage.

\paragraph{Standalone Detector Baselines.}
These methods bypass model selection entirely: they always apply one fixed detector, regardless of dataset characteristics. \texttt{IForest}~\citep{DBLP:conf/icdm/LiuTZ08} is the best-performing \texttt{Isolation Forest} configuration from the candidate pool; \texttt{kNN}~\citep{DBLP:conf/icpr/HautamakiKF04} is the best-performing $k$-nearest-neighbor configuration. \texttt{LOF}~\citep{DBLP:conf/sigmod/BreunigKNS00} is a fixed Local Outlier Factor detector ($k=10$, Euclidean metric) included as a representative density-based comparator. \texttt{DAGMM}~\citep{DBLP:conf/iclr/ZongSMCLCC18} jointly learns a compression network and a Gaussian mixture model in the latent space to produce density-based outlier scores. \texttt{DeepSVDD}~\citep{DBLP:conf/icml/RuffGDSVBMK18} maps normal data into a compact hypersphere via a neural encoder and uses distance from the learned center as the outlier score. \texttt{RandNet}~\citep{DBLP:conf/sdm/ChenSAT17} ensembles multiple autoencoders with randomized connectivity and uses aggregate reconstruction error as the outlier score. \texttt{RDA}~\citep{DBLP:conf/kdd/ZhouP17} decomposes data into a reconstructible normal component and a sparse outlier component via a robust deep autoencoder, using residual magnitude as the outlier score. \texttt{DTE-C}~\citep{DBLP:conf/iclr/LivernocheJHR24} is the categorical-parametric variant of the Diffusion Time Estimation framework; it is evaluated here under the same fully unsupervised protocol as the other standalone baselines. \texttt{TCCM}~\citep{DBLP:journals/corr/abs-2510-18328} is a flow-matching-based tabular outlier detector designed for scalability and robustness; it is included as a recent strong deep baseline for tabular data. \texttt{LUNAR}~\citep{DBLP:conf/aaai/GoodgeHNN22} generalizes local-neighborhood methods such as \texttt{kNN} and \texttt{LOF} into a learnable GNN framework, producing outlier scores from graph-structured representations of each test instance's neighborhood.

\paragraph{Aggregation Baselines.}
These baselines aggregate scores from multiple detectors rather than selecting one. \texttt{Full-Pool Ensemble} averages scores over the entire candidate pool $\mathcal{M}$; it serves as a ``just average everything'' control. \texttt{LSCP}~\citep{DBLP:conf/sdm/ZhaoNHL19} performs instance-wise locally selective combination: it identifies the most competent base detectors in the local neighborhood of each test instance and combines them via \texttt{AOM}/\texttt{MOA}-style aggregation.

\paragraph{Active Non-RL Baselines.}
These baselines are allowed to execute probes from $\mathcal{M}_{\text{probe}}$ before committing, but they use a fixed probe-ordering rule instead of a learned policy. \texttt{RandomProbing} selects probes uniformly at random up to $T_{\max}$ and then commits using the same trained \framework commit head; it isolates whether probe choice matters beyond merely being allowed to probe. \texttt{GreedyProbing} always executes probes in cheapest-first order, i.e., \texttt{HBOS} $\to$ \texttt{LODA} $\to$ \texttt{IForest} $\to \cdots$, and then commits with the same commit head; it tests whether \framework learns informative probe selection rather than simply cheap-first ordering. Both baselines share the same probe pool, maximum budget $T_{\max}=10$, and commit head as \framework; only the probe-selection policy differs.

\paragraph{Ablations of \framework}
\begin{itemize}[leftmargin=*, noitemsep]
  \item \textbf{$-$LOO-FPP:} The LOO-FPP evidence component is removed from the commit-time aggregation; the agent commits using the neural posterior blended with the LCB prior, with no label-free family-override stage. This isolates the contribution of LOO-FPP in extracting family evidence from probe responses.
  \item \textbf{$-$LCB Prior:} The lower confidence bound prior over detector families is removed from the commit-time reranking. The agent commits using the neural posterior alone, without the conservative family anchor.
  \item \textbf{\textsc{Static-Only}:} The dynamic stream and all probing actions are removed. The commit head acts directly on $\mathbf{z}_{\text{static}}$ after warm-start, with no RL fine-tuning. This isolates the contribution of sequential dynamic evidence.
  \item \textbf{$\lambda=0$:} The per-probe cost penalty is removed from the reward signal ($\lambda=0$ in Eq.~\ref{eq:reward_prelim}). This evaluates whether the cost-aware budget objective is necessary for efficient early stopping.
  \item \textbf{\textsc{WS-only}:} PPO fine-tuning is omitted entirely; the model commits using only the warm-start posterior, with no adaptive stopping rule. This isolates the contribution of the RL policy over the supervised initialization.
\end{itemize}

\begin{table*}[htbp]
  \caption{Pairwise significance of \framework vs.\ baselines
    (per-dataset mean AP; one-sided Wilcoxon signed-rank test, $\alpha=0.05$).
    $\Delta$AP: mean AP difference. W/T/L: wins/ties/losses per dataset.
    \textbf{ADBench-Advanced}: leave-one-base-dataset-out splits. With Holm correction over the 16 baselines,
    every comparison stays significant on both testbeds (largest adjusted $p$: $0.042$ on \textbf{ADBench-Classical}, $0.040$ on \textbf{ADBench-Advanced}).}
  \label{tab:significance}
  \centering
  \fontsize{7.5pt}{7.5pt}\selectfont
  \setlength{\tabcolsep}{4pt}
  \begin{tabular}{@{}l r r r r c r r r r c@{}}
    \toprule
    & \multicolumn{5}{c}{\textbf{ADBench-Classical}}
    & \multicolumn{5}{c}{\textbf{ADBench-Advanced}} \\
    \cmidrule(lr){2-6}\cmidrule(lr){7-11}
    Baseline & $\Delta$AP & W & T & L & $p$ & $\Delta$AP & W & T & L & $p$ \\
    \midrule
    \texttt{IForest}         & $+0.0283$ & 27 & 1 & 19 & $0.0078$ & $+0.1018$ & 70 & 2 & 18 & $0.0000$ \\
    \texttt{kNN} ($k{=}10$)  & $+0.0600$ & 28 & 1 & 18 & $0.0417$ & $+0.2075$ & 62 & 0 & 28 & $0.0000$ \\
    \texttt{LOF} (Euclidean) & $+0.0736$ & 27 & 1 & 19 & $0.0182$ & $+0.0785$ & 57 & 4 & 29 & $0.0132$ \\
    \texttt{DAGMM}           & $+0.1034$ & 41 & 0 & 6 & $0.0000$ & $+0.1997$ & 81 & 0 & 9 & $0.0000$ \\
    \texttt{DeepSVDD}        & $+0.2069$ & 40 & 0 & 7 & $0.0000$ & $+0.3076$ & 79 & 0 & 11 & $0.0000$ \\
    \texttt{GlobalBest}      & $+0.0460$ & 29 & 3 & 15 & $0.0044$ & $+0.0715$ & 51 & 10 & 29 & $0.0144$ \\
    \texttt{RandomSelect}    & $+0.1018$ & 36 & 0 & 11 & $0.0000$ & $+0.1799$ & 73 & 0 & 17 & $0.0000$ \\
    \texttt{RandNet}         & $+0.0778$ & 36 & 0 & 11 & $0.0001$ & $+0.0821$ & 61 & 5 & 24 & $0.0001$ \\
    \texttt{RDA}             & $+0.1371$ & 32 & 0 & 15 & $0.0000$ & $+0.1565$ & 47 & 0 & 43 & $0.0365$ \\
    \texttt{LSCP}             & $+0.0629$ & 29 & 0 & 18 & $0.0096$ & $+0.1043$ & 66 & 1 & 23 & $0.0000$ \\
    \texttt{Full-Pool Ensemble} & $+0.1019$ & 36 & 0 & 11 & $0.0000$ & $+0.2476$ & 77 & 1 & 12 & $0.0000$ \\
    \texttt{TCCM}            & $+0.0995$ & 30 & 1 & 16 & $0.0001$ & $+0.2162$ & 68 & 0 & 22 & $0.0000$ \\
    \texttt{DTE}-C           & $+0.0681$ & 32 & 1 & 14 & $0.0004$ & $+0.1061$ & 64 & 0 & 26 & $0.0003$ \\
    \texttt{LUNAR}           & $+0.1381$ & 35 & 0 & 12 & $0.0000$ & $+0.1373$ & 57 & 0 & 33 & $0.0036$ \\
    \texttt{ELECT}            & $+0.0417$ & 34 & 1 & 12 & $0.0001$ & $+0.0263$ & 57 & 4 & 29 & $0.0019$ \\
    \texttt{MetaOD}           & $+0.0711$ & 39 & 1 & 7 & $0.0000$ & $+0.0731$ & 63 & 4 & 23 & $0.0000$ \\
    \bottomrule
  \end{tabular}
\end{table*}

\subsection{Evaluation Protocol, Aggregation, and Statistical Significance}
\label{app:eval_metrics}
\label{app:eval_metrics:seeds}
\paragraph{Aggregation Across Seeds.}
Unless otherwise noted, all reported results use three random seeds: 0, 42, and 100. Aggregation follows a two-step protocol to avoid conflating seed variance with dataset-level variance:
\begin{enumerate}[leftmargin=*, noitemsep]
  \item \textbf{Per-seed mean}: for each seed $s$, we compute the mean metric across all datasets in the testbed,
  \[
  \bar{\mu}_s = \frac{1}{|\mathcal{D}|}\sum_{D \in \mathcal{D}} \operatorname{Metric}(D, s)
  \]
  \item \textbf{Cross-seed statistics}: we report
  \[
  \operatorname{Mean} = \frac{1}{3}\sum_s \bar{\mu}_s
  \]
  and
  \[
  \operatorname{Std} = \operatorname{std}(\bar{\mu}_0, \bar{\mu}_{42}, \bar{\mu}_{100})
  \]
  as the sample standard deviation with Bessel's correction ($n=3$).
\end{enumerate}

\paragraph{Statistical Significance.}
Table~\ref{tab:significance} reports pairwise Wilcoxon signed-rank test results comparing \framework against all baselines on both testbeds. Pairwise significance tests use a one-sided Wilcoxon signed-rank test (\framework $>$ baseline, $\alpha = 0.05$). The test is applied to per-dataset mean AP averaged across the three seeds, giving one paired observation per dataset. W/T/L counts report the number of datasets where \framework wins, ties, or loses against the baseline after averaging across seeds. A tie is counted when the rounded-to-4-decimal-place mean AP values are equal. Reported $p$-values are per-baseline and are used as descriptive evidence. With Holm correction over the 16 baselines, all comparisons stay significant on both testbeds (largest adjusted $p$: $0.042$ on \textbf{ADBench-Classical}, $0.040$ on \textbf{ADBench-Advanced}).


\subsection{Main Results with Per-Seed Standard Deviations}
\label{app:main_results_full}
Table~\ref{tab:main_results_full} reports the full version of the main comparison
(Table~\ref{tab:main_results}) with per-seed standard deviations across the three
seeds $\{0, 42, 100\}$. The main-text table omits the standard-deviation columns to
fit the two-column layout; all point estimates are identical, and the standard
deviations here confirm that \framework's improvements are stable across seeds
(e.g., AP std $\leq 0.007$ on both testbeds), consistent with the paired
significance tests in Table~\ref{tab:significance}.

\begin{table*}[t]
  \caption{Full version of the main comparison (Table~\ref{tab:main_results})
  with per-seed standard deviations. AP: Average Precision, mean$_{\pm\text{std}}$
  ($\uparrow$). Rank: mean pool rank$_{\pm\text{std}}$ ($\downarrow$; pool size
  297). AUROC: area under ROC curve, mean$_{\pm\text{std}}$ ($\uparrow$).
  Statistics are over three seeds $\{0, 42, 100\}$ (per-seed testbed means).
  \textbf{Bold}: best; \underline{underline}: second best;
  \textcolor{violet}{\texttt{Oracle}} is the label-aware upper bound.
  Point estimates are identical to Table~\ref{tab:main_results}. \textbf{ADBench-Advanced} uses leave-one-base-dataset-out splits (all variants of the target's base dataset are excluded from meta-training); this affects only the methods that learn from training datasets (\texttt{GlobalBest}, \texttt{MetaOD}, \texttt{ELECT}, \framework); the other methods do not use training datasets, so their entries are unchanged.}
  \label{tab:main_results_full}
  \centering
  \footnotesize
  \setlength{\tabcolsep}{4pt}
  \begin{tabular}{lcccccc}
    \toprule
    & \multicolumn{3}{c}{\textbf{ADBench-Classical}}
    & \multicolumn{3}{c}{\textbf{ADBench-Advanced}} \\
    \cmidrule(lr){2-4}\cmidrule(lr){5-7}
    Method
      & AP\,$\uparrow$
      & Rank\,$\downarrow$
      & AUROC\,$\uparrow$
      & AP\,$\uparrow$
      & Rank\,$\downarrow$
      & AUROC\,$\uparrow$ \\
    \midrule
    \multicolumn{7}{@{}l}{\textit{Reference controls}} \\[1pt]
    \texttt{Oracle}
      & \textcolor{violet}{$0.5930_{\pm 0.0000}$}
      & \textcolor{violet}{$1.0_{\pm 0.0}$}
      & \textcolor{violet}{$0.8580_{\pm 0.0000}$}
      & \textcolor{violet}{$0.7396_{\pm 0.0000}$}
      & \textcolor{violet}{$1.0_{\pm 0.0}$}
      & \textcolor{violet}{$0.9524_{\pm 0.0000}$} \\
    \texttt{RandomSelect}
      & $0.3128_{\pm 0.0593}$ & $145.5_{\pm 20.2}$ & $0.6851_{\pm 0.0478}$
      & $0.3049_{\pm 0.0055}$ & $145.6_{\pm 2.4}$ & $0.7494_{\pm 0.0170}$ \\
    \addlinespace[3pt]
    \multicolumn{7}{@{}l}{\textit{Standalone detectors}} \\[1pt]
    \texttt{IForest}
      & $\underline{0.3863}_{\pm 0.0016}$ & $104.2_{\pm 1.1}$ & $0.7679_{\pm 0.0025}$
      & $0.3830_{\pm 0.0053}$ & $124.7_{\pm 2.3}$ & $\underline{0.8621}_{\pm 0.0015}$ \\
    \texttt{kNN}
      & $0.3546_{\pm 0.0048}$ & $122.3_{\pm 2.3}$ & $0.7238_{\pm 0.0017}$
      & $0.2773_{\pm 0.0090}$ & $113.8_{\pm 2.0}$ & $0.6706_{\pm 0.0061}$ \\
    \texttt{LOF}
      & $0.3410_{\pm 0.0103}$ & $134.2_{\pm 4.6}$ & $0.7156_{\pm 0.0090}$
      & $0.4064_{\pm 0.0011}$ & $\underline{74.7}_{\pm 0.9}$ & $0.7983_{\pm 0.0035}$ \\
    \texttt{DAGMM}
      & $0.3112_{\pm 0.0090}$ & $173.3_{\pm 6.5}$ & $0.6989_{\pm 0.0064}$
      & $0.2851_{\pm 0.0055}$ & $187.2_{\pm 2.4}$ & $0.7325_{\pm 0.0070}$ \\
    \texttt{DeepSVDD}
      & $0.2076_{\pm 0.0101}$ & $204.5_{\pm 1.5}$ & $0.5929_{\pm 0.0133}$
      & $0.1772_{\pm 0.0031}$ & $184.4_{\pm 1.9}$ & $0.6875_{\pm 0.0058}$ \\
    \texttt{RandNet}
      & $0.3368_{\pm 0.0017}$ & $142.2_{\pm 2.5}$ & $0.7164_{\pm 0.0026}$
      & $0.4028_{\pm 0.0007}$ & $133.1_{\pm 1.3}$ & $0.8020_{\pm 0.0016}$ \\
    \texttt{RDA}
      & $0.2774_{\pm 0.0159}$ & $156.4_{\pm 9.3}$ & $0.7160_{\pm 0.0096}$
      & $0.3283_{\pm 0.0160}$ & $110.6_{\pm 5.7}$ & $0.6788_{\pm 0.0036}$ \\
    \texttt{DTE-C}
      & $0.3465_{\pm 0.0040}$ & $141.0_{\pm 1.6}$ & $0.7465_{\pm 0.0043}$
      & $0.3787_{\pm 0.0050}$ & $106.7_{\pm 2.8}$ & $0.8028_{\pm 0.0077}$ \\
    \texttt{TCCM}
      & $0.3151_{\pm 0.0091}$ & $149.5_{\pm 12.3}$ & $0.7059_{\pm 0.0058}$
      & $0.2686_{\pm 0.0034}$ & $162.0_{\pm 6.1}$ & $0.6999_{\pm 0.0162}$ \\
    \texttt{LUNAR}
      & $0.2765_{\pm 0.0040}$ & $166.5_{\pm 5.5}$ & $0.6188_{\pm 0.0046}$
      & $0.3475_{\pm 0.0085}$ & $102.4_{\pm 1.6}$ & $0.7544_{\pm 0.0195}$ \\
    \addlinespace[3pt]
    \multicolumn{7}{@{}l}{\textit{Aggregation baselines}} \\[1pt]
    \texttt{LSCP}
      & $0.3517_{\pm 0.0061}$ & $133.5_{\pm 2.5}$ & $0.7287_{\pm 0.0073}$
      & $0.3806_{\pm 0.0065}$ & $98.4_{\pm 0.9}$ & $0.8282_{\pm 0.0077}$ \\
    \texttt{Full-Pool Ensemble}
      & $0.3127_{\pm 0.0000}$ & $151.2_{\pm 0.0}$ & $0.7213_{\pm 0.0000}$
      & $0.2372_{\pm 0.0000}$ & $173.3_{\pm 0.0}$ & $0.6655_{\pm 0.0000}$ \\
    \addlinespace[3pt]
    \multicolumn{7}{@{}l}{\textit{Model selection}} \\[1pt]
    \texttt{GlobalBest}
      & $0.3686_{\pm 0.0000}$ & $\underline{101.4}_{\pm 0.0}$ & $\underline{0.7698}_{\pm 0.0000}$
      & $0.4133_{\pm 0.0000}$ & $81.4_{\pm 0.0}$ & $0.8007_{\pm 0.0000}$ \\
    \texttt{ELECT}
      & $0.3728_{\pm 0.0039}$ & $115.6_{\pm 8.0}$ & $0.7592_{\pm 0.0054}$
      & $\underline{0.4586}_{\pm 0.0174}$ & $106.8_{\pm 5.3}$ & $0.8442_{\pm 0.0049}$ \\
    \texttt{MetaOD}
      & $0.3434_{\pm 0.0220}$ & $141.7_{\pm 8.6}$ & $0.7027_{\pm 0.0127}$
      & $0.4117_{\pm 0.0107}$ & $122.5_{\pm 7.3}$ & $0.8023_{\pm 0.0292}$ \\
    \midrule
    \framework \textit{(ours)}
      & $\mathbf{0.4146}_{\pm 0.0068}$ & $\mathbf{82.7}_{\pm 2.9}$ & $\mathbf{0.7820}_{\pm 0.0024}$
      & $\mathbf{0.4848}_{\pm 0.0129}$ & $\mathbf{68.6}_{\pm 5.0}$ & $\mathbf{0.8693}_{\pm 0.0131}$ \\
    \bottomrule
  \end{tabular}
\end{table*}

\subsection{Full Experimental Results and Dataset-Level Analysis}
\label{app:per_dataset_results}

\begin{sloppypar}

For completeness, we report the full dataset-level results for
\textbf{ADBench-Classical} in Table~\ref{tab:per_dataset_classical}, the
aggregated outlier-type/deception breakdown for \textbf{ADBench-Advanced} in
Table~\ref{tab:aggregated_adv}, and the full per-dataset results for
\textbf{ADBench-Advanced} in Table~\ref{tab:per_dataset_adv}.
\end{sloppypar}

\paragraph{Dataset-level analysis on ADBench-Advanced.}
The per-dataset AP scatter is shown in the main text (Figure~\ref{fig:adv_scatter}): \framework outperforms \texttt{MetaOD} on $63/90$ datasets and \texttt{ELECT} on $57/90$ datasets (leave-one-base-dataset-out splits), with gains over both concentrated on the \emph{dependency} cases, where static meta-features are distorted by the adversarial construction; on \emph{tight-extreme-cluster}, \texttt{ELECT} is better under both deceptions and \texttt{MetaOD} under noise. \framework does not dominate every dataset: static methods remain competitive when their meta-features or internal performance measures are aligned with the target dataset. This supports our observation-limit argument---probing is most useful when probe responses expose detector-family evidence that static descriptors miss, and the benefit shrinks when static information is already sufficient or the probe pool is uninformative.

\subsection{Ablation Study}
\label{app:exp:ablation}

To understand the contribution of each component in \framework, we perform ablations by selectively disabling key mechanisms under the same protocols (LODO on \textbf{ADBench-Classical}, leave-one-base-dataset-out on \textbf{ADBench-Advanced}). Table~\ref{tab:ablation} summarizes the results across \textbf{ADBench-Classical} and \textbf{ADBench-Advanced}. On \textbf{ADBench-Classical}, every variant yields a degradation in AP, rank, or both. There, the largest drops arise from removing LOO-FPP, followed by the LCB prior and the dynamic stream, while \textsc{WS-only} and $\lambda=0$ produce smaller losses. On \textbf{ADBench-Advanced}, removing LOO-FPP ($-0.049$) or the dynamic stream ($-0.027$) costs AP, removing the LCB prior does not ($+0.007$), and \textsc{WS-only} and $\lambda=0$ make the same selections as Full \framework. Additional MLP-encoder, FamCE, and warm-start-initialization ablations are reported further below in this section.

\begin{table*}[t]
  \caption{Component ablations for \framework\ (3 seeds; LODO on \textbf{ADBench-Classical}, leave-one-base-dataset-out on \textbf{ADBench-Advanced}).
  AP: mean Average Precision $\pm$ std ($\uparrow$).
  Rank: mean pool rank ($\downarrow$; pool size $=297$).
  Probes: mean number of probes before commit.
  $\Delta$AP: signed gap vs.\ Full \framework (\textbf{bold}: reference row).
  $^\dagger$\,Probe counts for $-$LOO-FPP and $-$LCB match Full \framework by construction:
  these ablations share the full PPO stopping policy and modify only the
  commit-time evidence aggregation.}
  \label{tab:ablation_app}
  \centering
  \footnotesize
  \setlength{\tabcolsep}{3.5pt}
  \begin{tabular}{@{}l r@{\,}r@{\,}r@{\,}r@{\,}r r@{\,}r@{\,}r@{\,}r@{\,}r@{}}
    \toprule
    & \multicolumn{5}{c}{\textbf{ADBench-Classical}}
    & \multicolumn{5}{c}{\textbf{ADBench-Advanced}} \\
    \cmidrule(lr){2-6}\cmidrule(lr){7-11}
    Variant
      & \multicolumn{2}{c}{AP\,$\uparrow$}
      & Rank\,$\downarrow$
      & Probes
      & $\Delta$AP
      & \multicolumn{2}{c}{AP\,$\uparrow$}
      & Rank\,$\downarrow$
      & Probes
      & $\Delta$AP \\
    \midrule
    \textbf{Full AMP-OD}
      & \textbf{0.4146} & $\pm$\textbf{0.0068} & \textbf{82.7} & \textbf{3.72} & $\phantom{-}0.000$
      & \textbf{0.4848} & $\pm$\textbf{0.0129} & \textbf{68.6} & \textbf{4.12} & $\phantom{-}0.000$ \\
    \midrule
    \texttt{WS-only} (no PPO)
      & 0.4031 & $\pm$0.0065 & 89.2 & 5.50 & $-0.011$
      & 0.4848 & $\pm$0.0129 & 68.6 & 4.12 & $\phantom{-}0.000$ \\
    \addlinespace[3pt]
    $-$\,LOO-FPP$^\dagger$
      & 0.3856 & $\pm$0.0065 & 95.0 & 3.72 & $-0.029$
      & 0.4356 & $\pm$0.0122 & 71.5 & 4.12 & $-0.049$ \\
    \addlinespace[3pt]
    $-$\,LCB Prior$^\dagger$
      & 0.3935 & $\pm$0.0177 & 97.1 & 3.72 & $-0.021$
      & 0.4918 & $\pm$0.0055 & 63.4 & 4.12 & $+0.007$ \\
    \addlinespace[3pt]
    \texttt{Static-Only}
      & 0.3991 & $\pm$0.0048 & 90.6 & 0.00 & $-0.015$
      & 0.4578 & $\pm$0.0141 & 71.2 & 0.00 & $-0.027$ \\
    \addlinespace[3pt]
    $\lambda{=}0$ (no probe cost)
      & 0.4040 & $\pm$0.0048 & 88.5 & 3.93 & $-0.011$
      & 0.4848 & $\pm$0.0129 & 68.6 & 4.12 & $\phantom{-}0.000$ \\
    \bottomrule
  \end{tabular}
  \vspace{2pt}
\end{table*}

\paragraph{LOO-FPP Stratified Analysis (Details).}
Sec.~\ref{sec:exp:ablation} reports the headline LOO-FPP contribution and the gate-contrast example (Figure~\ref{fig:gate_contrast}); here we add supporting detail. The probe-reachable/probe-blind split uses the probe-quality guard (winner LOO-AP $\geq \delta_{\mathrm{commit}}=0.75$), which is distinct from the gate-firing threshold $\delta_{\mathrm{quality}}=0.80$ in Sec.~\ref{sec:method:loofpp}. The $+2.1$ pp \textbf{ADBench-Classical} average reflects high per-dataset heterogeneity, with individual gains ranging from $-6$ to $+14$ pp; on \textbf{ADBench-Advanced} the probe-reachable subset drives the overall $+4.9$ pp LOO-FPP contribution. On the probe-blind majority ($83\%$ on \textbf{ADBench-Classical} and $71\%$ on \textbf{ADBench-Advanced}), \framework still achieves AP $=0.388$ and AP $=0.449$, respectively; on \textbf{ADBench-Advanced} this is slightly below the variant without LOO-FPP ($0.472$), so outside the probe-reachable subset the soft LOO-FPP blend costs a little. Full hyperparameter sensitivity analyses for the gate thresholds ($\delta_{\text{range}}$, $\tau$, $Q_{0.95}$) and arbitration parameters ($K$, $\omega$, $\gamma$) are reported later in this appendix.

\paragraph{Dynamic Probing Improves over Static Meta-Features.}
A natural baseline is to commit using only static dataset meta-features, bypassing probe execution entirely (\textsc{Static-Only}). On \textbf{ADBench-Classical}, \textsc{Static-Only} suffers only a modest AP drop ($\Delta$AP $=-0.015$), which might suggest that static UOMS could be sufficient on well-behaved datasets. However, on \textbf{ADBench-Advanced}, where the geometry is deliberately challenging, the gap widens to $\Delta$AP $=-0.027$ and mean pool rank degrades from $68.6$ to $71.2$, showing that static descriptors are fragile once the distribution departs from the classical regime. Across both testbeds, \textsc{Static-Only} is worse than Full \framework on average ($p=0.07$ on \textbf{ADBench-Advanced}), consistent with the hypothesis that observing actual detector responses through dynamic probing improves model selection under distribution shift and adversarial meta-feature corruption.

\paragraph{The Sequential RL Policy and Cost Penalty Act as Joint Regularizers.}
The supervised warm-start (\textsc{WS-only}) provides a strong initialization but lacks an adaptive stopping rule, consuming on average $5.50$ probes on \textbf{ADBench-Classical} versus $3.72$ for Full \framework, while still underperforming ($\Delta$AP $=-0.011$). PPO fine-tuning therefore contributes primarily by learning when to stop, recovering accuracy with substantially fewer probes. Ablating the probe-cost penalty ($\lambda=0$) further illustrates this effect: the agent issues more probes on \textbf{ADBench-Classical} ($3.72$ to $3.93$) and loses $-0.011$ AP; on \textbf{ADBench-Advanced} the selections are identical to Full \framework. Overall, the sequential MDP formulation and probe-cost term act as joint regularizers: they do not merely save computation, but prevent the policy from overfitting to noisy, late-stage probe signals.

\paragraph{LCB Prior Provides Conservative Arbitration.}
The final commit-time reranking applies an LCB prior that down-weights historically unreliable detector families, even if a single probe looks promising. Removing it ($-$LCB Prior) substantially increases seed-level variability on \textbf{ADBench-Classical} (AP standard deviation $\pm0.0177$ vs.\ $\pm0.0068$) and reduces mean AP by $-0.021$. On \textbf{ADBench-Advanced}, removing the LCB prior does not hurt ($+0.007$, not significant), so its benefit is specific to \textbf{ADBench-Classical}, where it also stabilizes selection across seeds. Probe counts for both $-$LOO-FPP and $-$LCB Prior match Full \framework by construction (see Table~\ref{tab:ablation}, footnote $\dagger$), confirming that these are purely inference-time ablations of the commit-time aggregation that leave the learned stopping policy unchanged.

\paragraph{Auxiliary Architectural and Optimization Ablations.}
Table~\ref{tab:appendix:aux_ablations} tabulates three supplementary variants ($-$Warm-Start, the MLP encoder, and $-$FamCE) on \textbf{ADBench-Classical} (47 datasets, 3 seeds, LODO); we additionally discuss the Hierarchical Gate below, whose effect is negligible ($<0.003$ AP) and is therefore not tabulated.
All $p$-values are one-sided Wilcoxon signed-rank tests (per-dataset means, $n{=}47$), consistent with the main-text ablation.

\begin{table*}[t]
\centering
\footnotesize
\setlength{\tabcolsep}{3.5pt}
\caption{Auxiliary ablations on \textbf{ADBench-Classical} (47 datasets, 3 seeds).
  AP: mean $\pm$ population std of per-seed means. Rank: mean pool rank.
  $\Delta$AP relative to Full \framework\ (AP $=0.4146$, Rank $=82.7$, Probes $=3.72$).
  $^\dagger$One-sided Wilcoxon $p{<}0.05$ vs.\ Full.}
\label{tab:appendix:aux_ablations}
\begin{tabular}{lccccc}
\toprule
Variant & AP & $\pm$std & Rank & Probes & $\Delta$AP ($p$) \\
\midrule
$-$Warm-Start (random init)$^\dagger$
  & 0.3656 & $\pm$0.0139 & 104.7 & 3.40 & $-0.049$ ($p{=}0.007$) \\
MLP Encoder (no Transformer)
  & 0.4009 & $\pm$0.0070 &  89.5 & 3.67 & $-0.014$ ($p{=}0.090$) \\
$-$FamCE (no aux CE)
  & 0.4096 & $\pm$0.0022 &  87.5 & 3.42 & $-0.005$ ($p{=}0.291$) \\
\bottomrule
\end{tabular}
\end{table*}

\begin{itemize}[leftmargin=*, nosep]
\item \textbf{Hierarchical Gate.} Replacing the hard two-tier evidence gate with the soft three-way blend (no hard family override) on \textbf{ADBench-Classical}: AP changes by $<0.003$ and rank standard deviation across seeds shifts by $<3$ ranks. The gate is structurally redundant with the soft blend on \textbf{ADBench-Classical} but is retained for its transparent interpretability as an explicit family routing decision.
\item \textbf{$-$\,FamCE.} Removing the FamCE auxiliary loss ($\lambda_{\text{fam}}{=}0$) reduces AP by $0.005$ and worsens mean pool rank from $82.7$ to $87.5$ ($p{=}0.29$, not significant). Probe count decreases slightly ($3.42$ vs.\ $3.72$), suggesting earlier commitment to sub-optimal families without the family supervision signal. FamCE serves primarily as a gradient stabilizer for the 297-way discrete action space and not as a primary accuracy driver.
\item \textbf{$-$\,Warm-Start.} Bypassing the supervised warm-start and starting PPO from random initialization yields AP $=0.3656$ ($\Delta{=}{-}0.049$, $p{=}0.007^\dagger$), with rank worsening from $82.7$ to $104.7$. Without warm-start, the policy fails to identify high-performing families reliably, and the overall seed standard deviation roughly doubles ($\sigma_{\text{AP}}{=}0.014$ vs.\ $0.006$). This confirms that supervised warm-start is a necessary tractability mechanism for the 297-way discrete action space, not merely an initialization shortcut.
\item \textbf{MLP Encoder.} Replacing the Transformer dynamic stream with a flat MLP that concatenates and zero-pads probe signatures to $T_{\max}$ yields AP $=0.4009$ ($\Delta{=}{-}0.014$, $p{=}0.09$, not significant at $\alpha{=}0.05$). The primary failure mode is \emph{adaptive-stopping instability}: mean probe count is $3.67$ but seed standard deviation is $2.36$ (vs.\ $0.11$ for the Transformer)---a bimodal collapse the Transformer never exhibits. Order-sensitive temporal modeling appears necessary for a consistent learned stopping criterion.
\end{itemize}

\paragraph{LOO-FPP Activation on ADBench-Advanced.}
\label{app:loofpp_sensitivity}
\begin{sloppypar}
The \textbf{ADBench-Advanced} testbed is constructed so that the LOO-FPP gate can activate on the \emph{tight-extreme-cluster} outlier type, which uses a $14$--$18\sigma$ cluster offset. In this regime, isolation and density detectors such as \texttt{HBOS}, \texttt{IForest}, and \texttt{LODA} achieve near-perfect canonical LOO-AP (${\approx}0.95$), while \texttt{kNN} detectors score near zero because all outlier nearest neighbors are other cluster members. On most tight-extreme-cluster datasets the resulting canonical-LOO spread exceeds the $\delta_{\text{range}}=0.40$ activation threshold. When LOO-FPP fires, it overrides the corrupted static prior's family weighting and selects from the density/isolation family, yielding $0.73$ AP; without the gate ($-$LOO-FPP), the static prior's corrupted family weights allow \texttt{kNN} variants to dominate, reducing AP to $0.46$. The per-type breakdown confirms this mechanism:
\end{sloppypar}
\begin{itemize}[leftmargin=*, noitemsep]
  \item \texttt{tight-extreme-cluster}: Full $=0.726$, $-$LOO-FPP $=0.460$, $\Delta=+26.6$ pp; the gate fires on $60\%$ (noise) and $87\%$ (permute) of these datasets.
  \item \texttt{dependency}: Full $=0.374$, $-$LOO-FPP $=0.438$, $\Delta=-6.4$ pp (\texttt{dependency-strong}: $-5.5$ pp); the gate occasionally over-commits, giving a net negative effect on these types.
\end{itemize}
The net aggregate effect is $+4.9$ pp overall, averaged over three seeds (not significant over datasets, $p=0.24$), with tight-extreme-cluster gains outweighing the smaller dependency losses. Three-seed aggregate results are reported in Table~\ref{tab:ablation}.

\paragraph{Hyperparameter Sensitivity Analysis.}
\label{app:sensitivity}
Table~\ref{tab:app_sensitivity} reports a one-at-a-time sensitivity analysis over the six key hyperparameters of \framework. Each configuration holds all other parameters at their main-paper defaults and varies a single parameter to a lower or higher value. All runs use \textbf{ADBench-Classical} (47 datasets, LODO). Because these are commit-stage hyperparameters applied to the frozen policy, each perturbation is deterministic given that policy; the reported $\Delta$AP is the signed gap from the Full \framework baseline. All six parameters are commit-stage hyperparameters: they govern which model is selected at final commitment but do not enter the PPO stopping policy. The sequential probe count is therefore determined entirely by the frozen policy and varies across seeds but not across commit-stage configurations; we omit a probe-count column accordingly.

\begin{table}[t]
\centering
\caption{One-at-a-time hyperparameter sensitivity on \textbf{ADBench-Classical}.
$\Delta$AP is the signed gap from the Full \framework baseline (AP $=0.4112$). All six
parameters are commit-stage only and are applied to the \emph{frozen} policy, so
each perturbation is deterministic; the PPO stopping policy and probe count are
unchanged across rows. Large $|\Delta\text{AP}|$ values are shown in \textbf{bold}.}
\label{tab:app_sensitivity}
\small
\setlength{\tabcolsep}{5pt}
\begin{tabular}{@{}ll rr r@{}}
  \toprule
  Parameter & Tested & AP & Rank & $\Delta$AP \\
  \midrule
  \multicolumn{5}{@{}l}{\textit{Arbitration: candidate set size}} \\
  $K$ (default 20) & 10  & 0.4113 & 82.9 & $\phantom{-}0.000$ \\
                   & 30  & 0.4113 & 82.9 & $\phantom{-}0.000$ \\
  \addlinespace[2pt]
  \multicolumn{5}{@{}l}{\textit{LCB: confidence level}} \\
  $\delta$ (default 0.05) & 0.01 & 0.4113 & 82.9 & $\phantom{-}0.000$ \\
                          & 0.10 & 0.4114 & 82.7 & $\phantom{-}0.000$ \\
  \addlinespace[2pt]
  \multicolumn{5}{@{}l}{\textit{Arbitration blend: LCB vs.\ posterior ($\omega{\uparrow}$\,=\,more neural, less LCB)}} \\
  $\omega$ (default 0.50) & 0.30 & 0.4109 & 84.4 & $\phantom{-}0.000$ \\
                          & 0.70 & \textbf{0.3878} & \textbf{88.1} & $\mathbf{-0.023}$ \\
  \addlinespace[2pt]
  \multicolumn{5}{@{}l}{\textit{LOO-FPP family evidence weight ($\gamma{\downarrow}$\,=\,less probe evidence)}} \\
  $\gamma$ (default 0.40) & 0.20 & \textbf{0.3966} & \textbf{86.4} & $\mathbf{-0.015}$ \\
                          & 0.60 & 0.4108 & 83.4 & $\phantom{-}0.000$ \\
  \addlinespace[2pt]
  \multicolumn{5}{@{}l}{\textit{LOO-FPP softmax temperature}} \\
  $\tau$ (default 0.10)   & 0.05 & 0.4105 & 84.4 & $-0.001$ \\
                          & 0.20 & 0.4113 & 82.9 & $\phantom{-}0.000$ \\
  \addlinespace[2pt]
  \multicolumn{5}{@{}l}{\textit{LOO-FPP gate activation threshold}} \\
  $\delta_{\text{range}}$ (default 0.40) & 0.30 & \textbf{0.3976} & \textbf{95.1} & $\mathbf{-0.014}$ \\
                                         & 0.50 & \textbf{0.3987} & \textbf{92.4} & $\mathbf{-0.013}$ \\
  \bottomrule
\end{tabular}
\end{table}

The results reveal an interpretable, asymmetric sensitivity pattern. $K$ and $\delta$ are entirely insensitive ($|\Delta\text{AP}| \le 0.001$): the LCB acts as a structural conservative floor rather than a precise quantile estimator, so its output is stable across a decade of confidence levels and modest changes in candidate-set size. Concretely, for the 297-detector pool, the $K=10$ and $K=30$ candidate sets consistently expose the same top candidates, confirming that diversity saturation is reached well below $K=20$. $\tau$ is similarly stable ($|\Delta\text{AP}| \le 0.001$): a low temperature sharpens the family posterior without qualitatively changing the winning family, and the gate is already categorical at these scales.

The sensitive directions for $\omega$ and $\gamma$ reveal the underlying mechanism. Overweighting the neural posterior ($\omega=0.70$) partially removes the LCB stabilizer. Conversely, reducing the neural weight ($\omega=0.30$), giving LCB more authority, is harmless ($|\Delta\text{AP}|<0.001$), confirming that the LCB conservative floor is load-bearing. Similarly, halving the LOO-FPP evidence weight ($\gamma=0.20$, $\Delta\text{AP}=-0.015$) confirms that family evidence is useful, while increasing it to $\gamma=0.60$ is safe ($|\Delta\text{AP}|<0.001$). The shared pattern across both parameters is that moving toward more evidence (LCB or LOO-FPP) is harmless, while moving away from evidence is costly. The default midpoints ($\omega=0.50$, $\gamma=0.40$) sit comfortably in the stable region.

The $\delta_{\text{range}}$ gate is the only parameter where both directions incur a cost. On \textbf{ADBench-Classical}, the observed LOO-AP spread varies widely across datasets ($0.20$--$0.80$); the default $0.40$ correctly bisects this range, making the gate a balanced precision/recall binary decision. Lowering it to $0.30$ triggers spurious activations on datasets without a dominant family, degrading rank from $82.9$ to $95.1$ ($\Delta\text{AP}=-0.014$). Raising it to $0.50$ suppresses legitimate activations on moderate-spread datasets ($\Delta\text{AP}=-0.013$, rank $=92.4$). Crucially, on \textbf{ADBench-Advanced} the gate opens on most tight-extreme-cluster datasets ($60\%$ under noise, $87\%$ under permutation). The $\delta_{\text{range}}$ sensitivity is therefore an \textbf{ADBench-Classical}-specific calibration issue that does not replicate on the harder testbed where LOO-FPP matters most.

\subsection{Value of the Learned Policy}
\label{app:exp:policy_value}

To directly quantify the contribution of the PPO policy, we ablate against four controlled baselines that all share \framework's trained commit head but replace the PPO stopping logic with fixed-budget rules. We distinguish two regimes: (1) \textbf{max-budget heuristics}, which always exhaust a fixed probe budget ($k=T_{\text{feasible}}=7$, the full non-anchor pool), and (2) \textbf{static-budget heuristics}, which commit after exactly $k=4$ probes regardless of dataset difficulty. Within each regime, we test cheapest-first (\textsc{Greedy}) and uniform-random (\textsc{Random}) probe orderings. Table~\ref{tab:policy_value} reports results on both testbeds under the same protocols with three seeds.

\begin{table*}[htbp]
  \caption{Value of the learned RL policy (3 seeds; LODO on \textbf{ADBench-Classical}, leave-one-base-dataset-out on \textbf{ADBench-Advanced}).
  AP: mean Average Precision $\pm$ std ($\uparrow$).
  Rank: mean pool rank ($\downarrow$; pool size\,=\,297).
  Probes$^\dagger$: mean sequential probes before commit.
  $\Delta$AP: signed gap vs.\ Full \framework (\textbf{bold}: reference).
  All fixed-budget variants share \framework's trained commit head but differ
  in \emph{stopping logic}.
  \textsc{Greedy Max Budget} ($k{=}T_{\text{feasible}}{=}7$) probes all available
  non-anchor detectors in cheapest-first order (\emph{max-budget} heuristic).
  \textsc{Fixed-K Cheapest} ($k{=}4$) always runs the 4 cheapest non-anchor
  probes then commits (\emph{static-budget} heuristic).
  \textsc{Random} ($k{=}4$) draws 4 probes uniformly at random.
  All three exclude the 3 anchor detectors already encoded in
  $\mathbf{z}_{\text{static}}$.
  $\dagger$\,Full \framework's probe count varies \emph{per dataset}
  ($\sigma{\approx}2.74$ on Classical);
  all fixed-budget baselines have $\sigma{=}0$.}
  \label{tab:policy_value}
  \centering
  \footnotesize
  \setlength{\tabcolsep}{3.5pt}
  \begin{tabular}{@{}l r@{\,}r@{\,}r@{\,}r@{\,}r r@{\,}r@{\,}r@{\,}r@{\,}r@{}}
    \toprule
    & \multicolumn{5}{c}{\textbf{ADBench-Classical}}
    & \multicolumn{5}{c}{\textbf{ADBench-Advanced}} \\
    \cmidrule(lr){2-6}\cmidrule(lr){7-11}
    Variant
      & \multicolumn{2}{c}{AP\,$\uparrow$}
      & Rank\,$\downarrow$
      & Probes$^\dagger$
      & $\Delta$AP
      & \multicolumn{2}{c}{AP\,$\uparrow$}
      & Rank\,$\downarrow$
      & Probes$^\dagger$
      & $\Delta$AP \\
    \midrule
    \textbf{Full AMP-OD}
      & \textbf{0.4146} & $\pm$\textbf{0.0068} & \textbf{82.7} & \textbf{3.72} & $\phantom{-}0.000$
      & \textbf{0.4848} & $\pm$\textbf{0.0129} & \textbf{68.6} & \textbf{4.12} & $\phantom{-}0.000$ \\
    \midrule
    \textsc{WS-only} (no PPO)
      & 0.4031 & $\pm$0.0065 & 89.2 & 5.50 & $-0.011$
      & 0.4848 & $\pm$0.0129 & 68.6 & 4.12 & $\phantom{-}0.000$ \\
    \midrule
    \multicolumn{11}{@{}l}{\itshape Fixed-budget baselines (same commit head; PPO policy replaced)} \\[1pt]
    \textsc{Greedy Max Budget} ($k{=}7$)
      & 0.3795 & $\pm$0.0104 & 99.9  & 7.00 & $-0.035$
      & 0.4236 & $\pm$0.0066 & 76.2 & 7.00 & $-0.061$ \\
    \addlinespace[3pt]
    \textsc{Fixed-K Cheapest} ($k{=}4$)
      & 0.3812 & $\pm$0.0143 & 102.6 & 4.00 & $-0.033$
      & 0.4208 & $\pm$0.0108 & 76.8 & 4.00 & $-0.064$ \\
    \addlinespace[3pt]
    \textsc{Random} ($k{=}4$)
      & 0.3825 & $\pm$0.0121 & 100.1 & 4.00 & $-0.032$
      & 0.4271 & $\pm$0.0115 & 75.5 & 4.00 & $-0.058$ \\
    \bottomrule
  \end{tabular}
  \vspace{2pt}
\end{table*}

\paragraph{\framework Dominates Both Ends of the Probe-Budget Spectrum.}
Comparing Full \framework ($3.72$ probes) against the max-budget heuristic ($7.00$ probes) reveals a clear efficiency win: \textsc{Greedy Max Budget} achieves only $0.3795$ AP on \textbf{ADBench-Classical} and $0.4236$ AP on \textbf{ADBench-Advanced}, respectively $3.5$ and $6.1$ AP points below Full \framework, while using $88\%$ and $70\%$ more probes than \framework. Against the static-budget heuristics ($4.00$ probes), the gap remains: \textsc{Fixed-K Cheapest} trails by $3.3$ and $6.4$ AP points despite using a nearly identical average number of probes ($4.00$ vs.\ $3.72$ on \textbf{ADBench-Classical} and $4.00$ vs.\ $4.12$ on \textbf{ADBench-Advanced}). Thus, \framework sits in the efficiency sweet spot: near-static-budget execution cost while outperforming the max-budget accuracy ceiling.

\paragraph{Forcing Full-Pool Greedy Traces Degrades Performance.}
\begin{sloppypar}
A counterintuitive result in Table~\ref{tab:policy_value} is that \textsc{Greedy Max Budget} ($k=7$) performs comparably to \textsc{Fixed-K Cheapest} ($k=4$), and is marginally worse on \textbf{ADBench-Classical}: $0.3795$ vs.\ $0.3812$ on \textbf{ADBench-Classical}, and $0.4236$ vs.\ $0.4208$ on \textbf{ADBench-Advanced}; in both cases the differences are within the cross-seed variance of these fixed-budget baselines ($\sigma\approx0.007$--$0.014$). Forcing the model to consume all available probes degrades performance because the commit head is optimized for compact, high-signal evidence paths rather than indiscriminate full-pool traces. Appending late-stage, potentially conflicting detectors such as \texttt{OCSVM}, \texttt{ABOD}, and \texttt{COF} introduces noise rather than evidence. The RL agent naturally avoids this degradation by learning to stop early.
\end{sloppypar}

\paragraph{Adaptive Depth Drives the Primary Performance Gain.}
\textsc{Fixed-K Cheapest} and \textsc{Random}$(k=4)$ achieve nearly identical AP on both testbeds: $0.3812$ vs.\ $0.3825$ on \textbf{ADBench-Classical}, and $0.4208$ vs.\ $0.4271$ on \textbf{ADBench-Advanced}. This is within seed noise and confirms that which detector populates a fixed budget matters far less than the when-to-stop decision learned by the PPO policy. Swapping probe order has diminishing returns at the fixed-budget level; adaptive depth calibration separates \framework from static-budget alternatives. On \textbf{ADBench-Advanced}, the cross-seed variance is similar for all variants (\textsc{Random} $\pm0.0115$, \framework $\pm0.0129$).

\paragraph{The Policy Learns Per-Dataset Depth Calibration.}
A key signature of genuine adaptivity is probe-count variability. \framework uses $3.72 \pm 2.74$ probes on \textbf{ADBench-Classical} (inter-dataset standard deviation) and $4.12$ on average on \textbf{ADBench-Advanced}, while every fixed-budget baseline has $\sigma=0$ by construction. This spread, ranging from one probe on simple datasets up to all seven on hard ones, is learned rather than scheduled. \textsc{WS-only} also adapts (on \textbf{ADBench-Classical}, $\sigma\approx2.8$); there it over-probes by $1.78$ probes while losing $1.1$ AP points, and on \textbf{ADBench-Advanced} it makes exactly the same selections as Full \framework. Notably, \textsc{WS-only} on \textbf{ADBench-Advanced} outperforms the $k{=}4$ fixed-budget baselines by $5.8$--$6.4$ AP points at a similar probe count ($4.12$ vs.\ $4.00$), confirming that evidence quality, not probe count, drives the gain. On \textbf{ADBench-Classical}, the PPO policy corrects \textsc{WS-only}'s over-probing by stopping earlier ($3.72$ vs.\ $5.50$ probes) at higher AP. This depth calibration, not cost reduction alone, is the primary contribution of the RL formulation over the supervised warm-start.

\subsection{Computational Cost Analysis}
\label{app:exp:cost}

A central concern for execution-conditioned UOMS is whether active probing introduces prohibitive overhead compared with static meta-feature-based alternatives. We measure per-dataset inference cost on \textbf{ADBench-Classical} and compare against \texttt{MetaOD} and \texttt{ELECT}.

\paragraph{\framework Timing Breakdown.}
In benchmark mode, where detector scores are pre-cached, the \framework selection algorithm adds only $\mathbf{35}$\,ms of overhead per dataset, including ${\approx}14$\,ms for the Transformer encoder and less than $1$\,ms for LCB arbitration, measured from evaluation logs. For deployment mode, we run the ten probe-pool detectors that the selector reads on each of the 47 \textbf{ADBench-Classical} datasets from raw data and time each one. $O(n^2)$ probes, namely \texttt{ABOD} and \texttt{COF}, are subsampled to $n=2{,}000$ in deployment; these probes serve as evidence for LOO-FPP and the policy and are never the final selected detector. The final model is always executed on the full dataset as part of the equal-cost execution step excluded from this comparison. All timings are measured on a single core of an Intel Core Ultra~7 265K CPU. Because LOO-FPP reads the scores of all ten probe-pool detectors at cold start, the deployment cost is that of running those ten: median ${\approx}1{,}915$\,ms (p75: ${\approx}3{,}186$\,ms; p90: ${\approx}6{,}794$\,ms). The three anchors alone take ${\approx}156$\,ms (median); the cost grows on large datasets, where \texttt{OCSVM} and the $O(n^2)$ probes dominate. An earlier estimate of ${\approx}740$\,ms replayed only the policy's sequential probes and omitted these reads. All timings below measure selection overhead only; each method additionally requires executing the chosen detector once, a cost common to all approaches and therefore excluded from the comparison.

\paragraph{Comparison with Static Baselines.}
Table~\ref{tab:timing} summarizes the true deployment overhead of each method, i.e., the wall-clock cost of the selection step when operating on a previously unseen test dataset from raw data, before executing the final chosen detector. This final execution cost is equal across all methods and therefore excluded. \texttt{MetaOD}'s pipeline computes ${\approx}200$ meta-features by running four lightweight landmarker detectors, namely \texttt{HBOS}, \texttt{IForest}, \texttt{PCA}, and \texttt{LODA}, and uses a \texttt{Random Forest} to predict latent factors in a pre-trained matrix-factorization model~\citep{DBLP:conf/nips/ZhaoRA21}. Measured end-to-end across all 47 \textbf{ADBench-Classical} datasets, \texttt{MetaOD}'s selection overhead is ${\approx}400$\,ms (median; p75: ${\approx}450$\,ms). \framework's ${\approx}1.9$\,s overhead is therefore about $5\times$ that of \texttt{MetaOD}, while providing higher AP on both testbeds. \texttt{ELECT} requires running ${\approx}50$ detectors from raw data to compute their internal performance measures before its \texttt{LightGBM} selection algorithm can proceed. Rather than profiling these 50 diverse detectors separately from the main evaluation loop, we conservatively adopt the ${\approx}47$\,s deployment figure reported by the original publication~\citep{DBLP:conf/icdm/0016ZA22}, noting that \framework runs ten probe detectors, a small fraction of \texttt{ELECT}'s detector footprint. This makes \framework ${\approx}25\times$ faster, with higher AP on both testbeds. Figure~\ref{fig:pareto} places these methods on a cost--accuracy Pareto curve.

Additional cost-accounting notes for Table~\ref{tab:timing} (reported in the main text). \texttt{ELECT}'s algorithm-only overhead with pre-cached IPMs on all 47 \textbf{ADBench-Classical} datasets is ${\approx}6{,}500$\,ms (median; p75: ${\approx}6{,}600$\,ms), with less than $2\%$ spread across datasets; true deployment additionally requires executing ${\approx}50$ detectors from raw data to compute the IPMs, so we conservatively adopt the ${\approx}47$\,s figure from the original paper~\citep{DBLP:conf/icdm/0016ZA22}. Our \texttt{MetaOD} reimplementation follows the original methodology (same four landmarkers, matrix-factorization plus \texttt{Random Forest} inductive transfer, leave-one-out evaluation) but omits the original $O(d^2)$ ANOVA pairwise features, which are intractable for high-dimensional data (\textbf{census}, $d=500$, ${\approx}24$\,s; \textbf{InternetAds}, $d=1{,}555$, impractical), making our estimate conservative. \texttt{Full-Pool Ensemble} timing is measured on 13 representative \textbf{ADBench-Classical} datasets ($n\in[80,\,7{,}200]$), with $O(n^2)$ detectors (\texttt{ABOD}, \texttt{COF}) subsampled to $n=2{,}000$ as in \framework; the measured range is 7--133\,s, and the p75 position (\textbf{campaign}, $n=41{,}188$, $d=62$), which lies beyond this measured range, is extrapolated to exceed 130\,s, and high-dimensional datasets such as \textbf{census} ($d=500$) are intractable.

\clearpage
\begin{table*}[p]
\centering
\caption{Per-dataset AP on \textbf{ADBench-Classical} (47 datasets, 3 seeds, LODO). Each cell: mean AP (mean pool rank). \textbf{Bold}: best AP per dataset.}
\label{tab:per_dataset_classical}
\scriptsize
\setlength{\tabcolsep}{3pt}
\resizebox{\textwidth}{!}{
\begin{tabular}{l|rrrrrrrrrrrrrrrr|r}
\toprule
\textbf{Dataset} & \textbf{RS} & \textbf{IFor.} & \textbf{kNN} & \textbf{LOF} & \textbf{DAGMM} & \textbf{DSVDD} & \textbf{RNet} & \textbf{RDA} & \textbf{DTE-C} & \textbf{TCCM} & \textbf{LUNAR} & \textbf{LSCP} & \textbf{FPEns} & \textbf{GB} & \textbf{ELECT} & \textbf{MetaOD} & \textbf{\textbf{AMP-OD}} \\
\midrule
\texttt{ALOI} & 0.033 (180) & 0.034 (193) & 0.064 (24) & 0.053 (37) & 0.033 (204) & 0.046 (46) & 0.039 (73) & 0.038 (84) & 0.034 (168) & 0.037 (93) & \textbf{0.078 (11)} & 0.054 (36) & 0.034 (168) & 0.037 (97) & 0.035 (138) & 0.031 (228) & 0.037 (95) \\
\texttt{Cardiotocography} & 0.344 (168) & 0.432 (46) & 0.314 (205) & 0.372 (129) & 0.389 (103) & 0.207 (281) & \textbf{0.545 (2)} & 0.268 (252) & 0.231 (277) & 0.429 (135) & 0.219 (281) & 0.382 (115) & 0.413 (73) & 0.337 (161) & 0.363 (132) & 0.361 (141) & 0.365 (129) \\
\texttt{Hepatitis} & 0.336 (76) & 0.287 (152) & 0.218 (238) & 0.203 (251) & 0.257 (186) & 0.163 (274) & 0.291 (148) & 0.202 (254) & 0.219 (199) & 0.221 (209) & 0.217 (239) & 0.173 (270) & 0.327 (97) & \textbf{0.372 (25)} & 0.331 (81) & 0.306 (120) & 0.312 (118) \\
\texttt{InternetAds} & 0.444 (79) & 0.524 (38) & 0.395 (93) & 0.531 (32) & 0.295 (174) & 0.237 (236) & 0.526 (39) & 0.295 (173) & 0.269 (204) & 0.368 (105) & 0.368 (105) & \textbf{0.564 (1)} & 0.479 (71) & 0.297 (156) & 0.455 (67) & 0.267 (189) & 0.527 (36) \\
\texttt{Ionosphere} & 0.632 (194) & 0.764 (152) & 0.925 (9) & 0.835 (56) & 0.644 (218) & 0.540 (229) & 0.709 (198) & \textbf{0.947 (1)} & 0.792 (105) & 0.657 (162) & 0.901 (22) & 0.777 (136) & 0.415 (235) & 0.778 (131) & 0.781 (125) & 0.631 (166) & 0.763 (152) \\
\texttt{Lymphography} & 0.635 (154) & 0.922 (142) & 0.948 (120) & 0.627 (211) & 0.884 (152) & 0.304 (232) & 0.941 (114) & 0.203 (235) & 0.792 (193) & 0.410 (225) & 0.938 (126) & 0.342 (225) & \textbf{1.000 (1)} & 0.948 (88) & 0.934 (102) & 0.682 (103) & 0.976 (49) \\
\texttt{PageBlocks} & 0.238 (228) & 0.500 (119) & 0.214 (248) & 0.571 (15) & 0.372 (162) & 0.282 (226) & 0.518 (89) & 0.497 (95) & \textbf{0.610 (1)} & 0.513 (107) & 0.198 (257) & 0.569 (16) & 0.380 (188) & 0.571 (13) & 0.504 (113) & 0.461 (99) & 0.569 (16) \\
\texttt{Pima} & 0.504 (109) & 0.503 (116) & 0.458 (206) & 0.476 (183) & 0.393 (268) & 0.354 (281) & 0.388 (271) & 0.451 (214) & 0.456 (205) & 0.394 (258) & 0.486 (153) & 0.478 (176) & 0.521 (57) & \textbf{0.529 (24)} & 0.495 (142) & 0.477 (154) & \textbf{0.529 (24)} \\
\texttt{SpamBase} & 0.419 (149) & 0.436 (67) & \textbf{0.601 (1)} & 0.467 (45) & 0.412 (117) & 0.380 (186) & 0.404 (155) & 0.421 (104) & 0.383 (203) & 0.479 (44) & 0.381 (207) & 0.495 (39) & 0.410 (138) & 0.417 (101) & 0.444 (82) & 0.400 (168) & 0.433 (63) \\
\texttt{Stamps} & 0.316 (122) & 0.341 (75) & 0.295 (175) & 0.325 (119) & 0.295 (162) & 0.178 (226) & 0.110 (287) & 0.160 (269) & 0.280 (159) & 0.167 (266) & 0.171 (264) & 0.337 (91) & 0.327 (113) & \textbf{0.354 (46)} & 0.335 (91) & 0.319 (131) & 0.348 (58) \\
\texttt{WBC} & 0.375 (224) & 0.938 (51) & 0.890 (86) & \textbf{0.963 (8)} & 0.833 (130) & 0.524 (224) & 0.344 (249) & 0.326 (249) & 0.632 (211) & 0.629 (186) & 0.806 (150) & 0.799 (110) & 0.901 (99) & 0.924 (68) & 0.783 (166) & 0.889 (79) & 0.920 (74) \\
\texttt{WDBC} & 0.619 (114) & 0.609 (109) & \textbf{0.956 (2)} & 0.945 (3) & 0.528 (170) & 0.222 (254) & 0.674 (81) & 0.102 (273) & 0.411 (208) & 0.299 (249) & 0.224 (259) & 0.949 (3) & 0.633 (91) & 0.538 (159) & 0.786 (30) & 0.762 (37) & 0.538 (159) \\
\texttt{WPBC} & 0.236 (112) & 0.227 (219) & 0.226 (214) & 0.225 (242) & 0.225 (202) & 0.248 (170) & 0.226 (238) & 0.240 (88) & 0.242 (110) & 0.220 (266) & 0.228 (199) & 0.210 (289) & 0.232 (182) & 0.232 (164) & 0.235 (114) & 0.241 (81) & \textbf{0.251 (53)} \\
\texttt{Waveform} & 0.051 (199) & 0.061 (123) & 0.150 (8) & 0.204 (5) & 0.042 (221) & 0.051 (211) & 0.051 (196) & 0.045 (239) & 0.049 (224) & 0.056 (161) & 0.170 (6) & \textbf{0.237 (4)} & 0.048 (243) & 0.136 (17) & 0.056 (155) & 0.051 (201) & 0.135 (20) \\
\texttt{Wilt} & 0.037 (275) & 0.044 (214) & 0.078 (31) & 0.089 (23) & \textbf{0.173 (2)} & 0.051 (141) & 0.054 (79) & 0.148 (16) & 0.073 (36) & 0.044 (205) & 0.081 (35) & 0.082 (28) & 0.039 (263) & 0.047 (158) & 0.048 (143) & 0.045 (187) & 0.047 (141) \\
\texttt{annthyroid} & 0.264 (123) & 0.256 (54) & 0.214 (172) & 0.249 (63) & 0.154 (247) & 0.131 (258) & 0.130 (264) & 0.230 (128) & \textbf{0.569 (3)} & 0.172 (188) & 0.177 (227) & 0.255 (52) & 0.120 (268) & 0.224 (142) & 0.246 (75) & 0.149 (224) & 0.223 (154) \\
\texttt{backdoor} & 0.113 (189) & 0.060 (185) & 0.385 (50) & 0.273 (75) & 0.316 (60) & 0.308 (67) & 0.073 (151) & \textbf{0.580 (3)} & 0.541 (12) & 0.574 (3) & 0.172 (91) & 0.291 (70) & 0.263 (76) & 0.050 (220) & 0.092 (171) & 0.211 (130) & 0.515 (19) \\
\texttt{breastw} & 0.966 (41) & \textbf{0.973 (22)} & 0.928 (158) & 0.352 (249) & 0.824 (189) & 0.571 (215) & 0.741 (192) & 0.852 (184) & 0.893 (177) & 0.428 (242) & 0.679 (197) & 0.574 (209) & 0.353 (249) & 0.956 (93) & 0.939 (123) & 0.775 (125) & 0.956 (84) \\
\texttt{campaign} & 0.264 (121) & 0.304 (47) & 0.250 (162) & 0.183 (209) & 0.256 (142) & 0.165 (220) & 0.273 (123) & 0.217 (183) & 0.267 (133) & 0.229 (168) & 0.235 (172) & 0.195 (199) & 0.317 (41) & 0.298 (53) & \textbf{0.392 (18)} & 0.291 (82) & 0.307 (43) \\
\texttt{cardio} & 0.447 (120) & 0.566 (23) & 0.300 (214) & 0.294 (216) & 0.355 (182) & 0.161 (264) & \textbf{0.622 (8)} & 0.200 (250) & 0.241 (231) & 0.367 (167) & 0.143 (273) & 0.432 (144) & 0.537 (38) & 0.442 (130) & 0.419 (135) & 0.366 (157) & 0.574 (18) \\
\texttt{celeba} & 0.070 (85) & 0.049 (111) & 0.033 (194) & 0.021 (230) & 0.079 (70) & 0.025 (209) & 0.082 (72) & 0.038 (180) & 0.039 (177) & \textbf{0.130 (3)} & 0.040 (185) & 0.023 (217) & 0.041 (178) & 0.088 (57) & 0.097 (15) & 0.028 (215) & 0.087 (61) \\
\texttt{census} & 0.069 (158) & 0.078 (126) & 0.081 (89) & 0.047 (285) & 0.053 (278) & 0.060 (248) & 0.071 (200) & 0.086 (37) & 0.076 (146) & 0.085 (46) & 0.077 (124) & 0.044 (296) & 0.065 (223) & \textbf{0.086 (30)} & 0.081 (91) & 0.073 (185) & 0.075 (154) \\
\texttt{cover} & 0.091 (113) & 0.068 (156) & 0.063 (165) & \textbf{0.274 (9)} & 0.038 (234) & 0.074 (125) & 0.051 (210) & 0.041 (226) & 0.074 (130) & 0.155 (62) & 0.026 (255) & 0.228 (14) & 0.056 (199) & 0.087 (78) & 0.081 (106) & 0.049 (188) & 0.088 (76) \\
\texttt{donors} & 0.105 (146) & 0.109 (116) & 0.148 (83) & 0.063 (239) & 0.101 (177) & 0.056 (245) & 0.144 (85) & 0.163 (67) & 0.145 (83) & \textbf{0.236 (35)} & 0.056 (247) & 0.061 (241) & 0.078 (224) & 0.151 (77) & 0.118 (110) & 0.083 (194) & 0.157 (68) \\
\texttt{fault} & 0.403 (167) & 0.428 (119) & 0.523 (27) & 0.407 (174) & 0.303 (291) & 0.389 (215) & 0.466 (63) & \textbf{0.557 (1)} & 0.484 (47) & 0.330 (283) & 0.490 (43) & 0.409 (163) & 0.406 (170) & 0.512 (37) & 0.424 (136) & 0.451 (104) & 0.513 (36) \\
\texttt{fraud} & 0.127 (179) & 0.203 (97) & 0.340 (27) & \textbf{0.494 (3)} & 0.109 (231) & 0.080 (247) & 0.284 (32) & 0.041 (262) & 0.196 (119) & 0.142 (179) & 0.168 (157) & 0.481 (5) & 0.060 (253) & 0.195 (100) & 0.190 (114) & 0.141 (186) & 0.197 (93) \\
\texttt{glass} & 0.115 (164) & 0.105 (210) & 0.165 (41) & 0.107 (200) & 0.149 (73) & 0.148 (92) & 0.135 (117) & \textbf{0.197 (28)} & 0.127 (168) & 0.129 (115) & 0.151 (102) & 0.104 (200) & 0.102 (218) & 0.117 (152) & 0.114 (164) & 0.119 (150) & 0.116 (155) \\
\texttt{http} & 0.336 (135) & 0.543 (36) & 0.004 (270) & 0.397 (101) & 0.306 (147) & 0.033 (238) & 0.488 (44) & 0.461 (54) & 0.480 (51) & 0.265 (160) & 0.241 (186) & 0.342 (147) & 0.086 (206) & 0.495 (43) & 0.384 (126) & 0.308 (127) & \textbf{0.615 (36)} \\
\texttt{landsat} & 0.228 (134) & 0.207 (203) & 0.251 (69) & 0.244 (98) & 0.243 (87) & \textbf{0.328 (9)} & 0.175 (271) & 0.214 (178) & 0.233 (122) & 0.214 (184) & 0.239 (106) & 0.252 (70) & 0.174 (274) & 0.258 (37) & 0.171 (265) & 0.215 (176) & 0.204 (221) \\
\texttt{letter} & 0.281 (65) & 0.097 (176) & 0.360 (32) & 0.238 (68) & 0.099 (195) & 0.091 (209) & 0.100 (162) & 0.277 (55) & 0.238 (69) & 0.084 (216) & \textbf{0.369 (31)} & 0.193 (84) & 0.095 (186) & 0.181 (87) & 0.097 (172) & 0.140 (160) & 0.187 (85) \\
\texttt{magic.gamma} & 0.481 (234) & 0.648 (98) & \textbf{0.730 (14)} & 0.617 (167) & 0.603 (191) & 0.513 (263) & 0.516 (266) & 0.500 (270) & 0.625 (140) & 0.551 (215) & 0.599 (194) & 0.629 (137) & 0.618 (168) & 0.704 (37) & 0.622 (137) & 0.539 (172) & 0.706 (35) \\
\texttt{mammography} & 0.125 (141) & \textbf{0.194 (21)} & 0.166 (99) & 0.145 (147) & 0.129 (176) & 0.057 (268) & 0.055 (271) & 0.141 (153) & 0.119 (195) & 0.115 (183) & 0.105 (219) & 0.066 (263) & 0.081 (257) & 0.187 (39) & 0.181 (58) & 0.184 (52) & 0.188 (34) \\
\texttt{mnist} & 0.307 (93) & 0.299 (93) & 0.341 (60) & 0.330 (68) & 0.243 (147) & 0.170 (198) & 0.391 (31) & 0.401 (21) & 0.359 (46) & 0.403 (43) & 0.364 (45) & 0.365 (46) & 0.351 (52) & \textbf{0.414 (1)} & 0.269 (134) & 0.203 (165) & 0.413 (5) \\
\texttt{musk} & 0.395 (205) & 1.000 (49) & 0.032 (276) & 0.254 (243) & 0.891 (135) & 0.058 (266) & 0.998 (62) & 0.368 (211) & 0.631 (175) & 0.655 (179) & 0.030 (285) & \textbf{1.000 (45)} & 0.892 (134) & 0.912 (130) & 0.987 (41) & 0.871 (103) & 1.000 (52) \\
\texttt{optdigits} & 0.030 (179) & 0.054 (74) & 0.026 (202) & 0.026 (205) & 0.017 (296) & 0.039 (132) & 0.068 (55) & 0.027 (195) & 0.028 (192) & 0.032 (170) & 0.021 (268) & 0.039 (133) & \textbf{0.167 (7)} & 0.022 (243) & 0.035 (169) & 0.022 (247) & 0.021 (254) \\
\texttt{pendigits} & 0.117 (180) & \textbf{0.248 (30)} & 0.066 (212) & 0.035 (262) & 0.094 (184) & 0.039 (251) & 0.185 (106) & 0.037 (257) & 0.044 (242) & 0.070 (214) & 0.035 (262) & 0.060 (218) & 0.209 (77) & 0.118 (167) & 0.172 (124) & 0.121 (170) & 0.211 (76) \\
\texttt{satellite} & 0.533 (162) & 0.632 (99) & 0.538 (212) & 0.397 (250) & 0.591 (150) & 0.457 (231) & 0.538 (207) & 0.467 (229) & 0.558 (198) & 0.597 (146) & 0.331 (287) & 0.412 (242) & \textbf{0.655 (50)} & 0.593 (173) & 0.636 (91) & 0.581 (168) & 0.652 (54) \\
\texttt{satimage-2} & 0.570 (158) & 0.946 (26) & 0.350 (225) & 0.150 (238) & 0.440 (214) & 0.042 (267) & \textbf{0.961 (14)} & 0.193 (235) & 0.212 (235) & 0.525 (206) & 0.032 (274) & 0.459 (211) & 0.951 (16) & 0.874 (87) & 0.815 (71) & 0.849 (97) & 0.937 (39) \\
\texttt{shuttle} & 0.620 (147) & 0.886 (76) & 0.246 (205) & 0.195 (224) & 0.618 (163) & 0.176 (235) & 0.926 (52) & 0.314 (187) & 0.698 (147) & 0.852 (110) & 0.114 (268) & 0.232 (209) & 0.527 (165) & 0.260 (197) & 0.926 (49) & 0.556 (167) & \textbf{0.954 (26)} \\
\texttt{skin} & 0.233 (135) & 0.247 (94) & 0.283 (27) & 0.156 (280) & 0.173 (258) & 0.202 (219) & 0.205 (218) & 0.231 (141) & \textbf{0.354 (2)} & 0.314 (101) & 0.230 (147) & 0.145 (288) & 0.177 (259) & 0.286 (10) & 0.242 (102) & 0.245 (109) & 0.286 (14) \\
\texttt{smtp} & 0.792 (59) & \textbf{1.000 (1)} & \textbf{1.000 (1)} & \textbf{1.000 (1)} & 0.819 (57) & 0.935 (49) & 0.323 (186) & 0.693 (156) & \textbf{1.000 (1)} & \textbf{1.000 (1)} & 0.972 (48) & 0.972 (48) & 0.003 (270) & \textbf{1.000 (1)} & \textbf{1.000 (1)} & \textbf{1.000 (1)} & \textbf{1.000 (1)} \\
\texttt{speech} & 0.019 (177) & 0.021 (132) & 0.020 (138) & 0.019 (160) & 0.017 (231) & 0.017 (216) & 0.019 (180) & 0.019 (183) & 0.018 (210) & \textbf{0.035 (24)} & 0.024 (112) & 0.019 (170) & 0.021 (117) & 0.019 (171) & 0.018 (198) & 0.018 (194) & 0.019 (166) \\
\texttt{thyroid} & 0.435 (71) & 0.458 (48) & 0.243 (219) & 0.386 (90) & 0.214 (226) & 0.053 (261) & 0.183 (238) & 0.214 (230) & \textbf{0.684 (29)} & 0.491 (82) & 0.138 (242) & 0.390 (88) & 0.112 (246) & 0.410 (65) & 0.361 (138) & 0.383 (128) & 0.408 (68) \\
\texttt{vertebral} & 0.103 (90) & 0.093 (134) & 0.089 (193) & 0.090 (180) & 0.100 (78) & 0.104 (78) & \textbf{0.155 (10)} & 0.123 (30) & 0.103 (126) & 0.146 (22) & 0.099 (97) & 0.092 (173) & 0.089 (197) & 0.087 (226) & 0.092 (146) & 0.101 (102) & 0.090 (167) \\
\texttt{vowels} & 0.272 (173) & 0.151 (122) & 0.542 (20) & 0.393 (53) & 0.051 (263) & 0.062 (251) & 0.130 (144) & 0.422 (49) & 0.314 (67) & 0.147 (129) & \textbf{0.632 (10)} & 0.350 (67) & 0.255 (82) & 0.153 (121) & 0.153 (125) & 0.518 (24) & 0.405 (50) \\
\texttt{wine} & 0.254 (133) & 0.311 (100) & 0.915 (1) & \textbf{0.991 (1)} & 0.127 (235) & 0.083 (262) & 0.224 (174) & 0.067 (275) & 0.163 (203) & 0.112 (248) & 0.229 (159) & 0.667 (38) & 0.361 (50) & 0.335 (74) & 0.284 (123) & 0.309 (131) & 0.335 (74) \\
\texttt{yeast} & 0.319 (178) & 0.306 (204) & 0.302 (232) & 0.313 (163) & 0.340 (65) & \textbf{0.344 (88)} & 0.298 (255) & 0.323 (147) & 0.299 (248) & 0.295 (230) & 0.308 (195) & 0.299 (248) & 0.317 (139) & 0.292 (281) & 0.303 (223) & 0.319 (144) & 0.292 (283) \\
\midrule
\textbf{Mean} & 0.313 (145) & 0.386 (104) & 0.355 (122) & 0.341 (134) & 0.311 (173) & 0.208 (205) & 0.337 (142) & 0.277 (156) & 0.347 (141) & 0.315 (149) & 0.276 (166) & 0.352 (133) & 0.313 (151) & 0.369 (101) & 0.373 (116) & 0.343 (142) & \textbf{0.415 (83)} \\
\bottomrule
\end{tabular}
}
\end{table*}

\begin{table*}[ht]
\centering
\caption{Aggregated AP on \textbf{ADBench-Advanced} (90 advanced synthetic datasets), broken down by outlier type and deception mechanism. Each entry: mean AP over 15 base datasets $\times$ 3 seeds. \textbf{Bold}: best per row. Leave-one-base-dataset-out splits; only \texttt{GB}, \texttt{ELECT}, \texttt{MetaOD} and \framework use training datasets.}
\label{tab:aggregated_adv}
\small
\setlength{\tabcolsep}{4pt}
\resizebox{\textwidth}{!}{
\begin{tabular}{l l r|rrrrrrrrrrrrrrrr|r}
\toprule
\textbf{Outlier type} & \textbf{Deception} & \textbf{\#D} & \textbf{RS} & \textbf{IFor.} & \textbf{kNN} & \textbf{LOF} & \textbf{DAGMM} & \textbf{DSVDD} & \textbf{RNet} & \textbf{RDA} & \textbf{DTE-C} & \textbf{TCCM} & \textbf{LUNAR} & \textbf{LSCP} & \textbf{FPEns} & \textbf{GB} & \textbf{ELECT} & \textbf{MetaOD} & \textbf{\textbf{AMP-OD}} \\
\midrule
Dependency & Noise & 15 & 0.1881 & 0.1386 & 0.3241 & 0.3314 & 0.1020 & 0.1168 & 0.1654 & \textbf{0.3539} & 0.3155 & 0.2813 & 0.2964 & 0.2783 & 0.1268 & 0.3151 & 0.2648 & 0.1697 & 0.3444 \\
Dependency & Permute & 15 & 0.2381 & 0.2214 & 0.4773 & 0.4634 & 0.1779 & 0.1940 & 0.1443 & \textbf{0.6151} & 0.4412 & 0.3307 & 0.5185 & 0.4003 & 0.1818 & 0.5235 & 0.2735 & 0.2729 & 0.4040 \\
Dep.-Strong & Noise & 15 & 0.1655 & 0.1473 & 0.3036 & 0.3163 & 0.0935 & 0.1263 & 0.1675 & \textbf{0.3480} & 0.2976 & 0.2369 & 0.2882 & 0.2711 & 0.1333 & 0.3219 & 0.2747 & 0.1666 & 0.2806 \\
Dep.-Strong & Permute & 15 & 0.2016 & 0.2221 & 0.4452 & 0.4408 & 0.1645 & 0.1588 & 0.1396 & \textbf{0.5825} & 0.4298 & 0.3138 & 0.5013 & 0.3885 & 0.1863 & 0.5139 & 0.2577 & 0.2464 & 0.4275 \\
Tight-Extr.-Cluster & Noise & 15 & 0.4663 & 0.8734 & 0.0683 & 0.4528 & 0.6419 & 0.1359 & \textbf{0.9000} & 0.0386 & 0.3363 & 0.3045 & 0.3063 & 0.4693 & 0.7688 & 0.3876 & 0.7765 & 0.7915 & 0.5860 \\
Tight-Extr.-Cluster & Permute & 15 & 0.5697 & 0.6951 & 0.0454 & 0.4336 & 0.5310 & 0.3313 & 0.8997 & 0.0315 & 0.4519 & 0.1447 & 0.1746 & 0.4758 & 0.0260 & 0.4178 & \textbf{0.9041} & 0.8230 & 0.8663 \\
\midrule
\textbf{All (90)} & & 90 & 0.3049 & 0.3830 & 0.2773 & 0.4064 & 0.2851 & 0.1772 & 0.4028 & 0.3283 & 0.3787 & 0.2686 & 0.3475 & 0.3806 & 0.2372 & 0.4133 & 0.4586 & 0.4117 & \textbf{0.4848} \\
\bottomrule
\end{tabular}
}
\end{table*}

\begin{table*}[p]
\centering
\caption{Per-dataset AP on \textbf{ADBench-Advanced} (90 advanced synthetic datasets, 3 seeds, leave-one-base-dataset-out splits). Each cell: mean AP (mean pool rank). \textbf{Bold}: best AP per dataset. Datasets grouped by outlier type and deception mechanism. Only \texttt{GB}, \texttt{ELECT}, \texttt{MetaOD} and \framework use training datasets.}
\label{tab:per_dataset_adv}
\scriptsize
\setlength{\tabcolsep}{3pt}
\resizebox{0.95\textwidth}{!}{
\begin{tabular}{l|rrrrrrrrrrrrrrrr|r}
\toprule
\textbf{Dataset} & \textbf{RS} & \textbf{IFor.} & \textbf{kNN} & \textbf{LOF} & \textbf{DAGMM} & \textbf{DSVDD} & \textbf{RNet} & \textbf{RDA} & \textbf{DTE-C} & \textbf{TCCM} & \textbf{LUNAR} & \textbf{LSCP} & \textbf{FPEns} & \textbf{GB} & \textbf{ELECT} & \textbf{MetaOD} & \textbf{\textbf{AMP-OD}} \\
\midrule
\multicolumn{18}{@{}l}{\textit{Dependency / Noise}} \\[1pt]
\texttt{Annthyroid} & 0.168 (74) & 0.116 (130) & 0.185 (38) & 0.185 (38) & 0.076 (263) & 0.081 (259) & 0.099 (223) & 0.157 (71) & 0.200 (19) & 0.109 (187) & 0.148 (82) & 0.158 (71) & 0.122 (107) & 0.203 (17) & 0.100 (201) & 0.133 (133) & \textbf{0.222 (5)} \\
\texttt{Cardiotocography} & 0.085 (123) & 0.062 (170) & 0.122 (56) & 0.139 (46) & 0.042 (278) & 0.069 (151) & 0.054 (217) & 0.194 (22) & 0.081 (100) & 0.081 (136) & 0.146 (42) & 0.102 (74) & 0.056 (210) & \textbf{0.199 (20)} & 0.085 (161) & 0.099 (161) & \textbf{0.199 (20)} \\
\texttt{Glass} & 0.074 (204) & 0.084 (170) & 0.161 (34) & 0.118 (63) & 0.091 (146) & 0.074 (203) & 0.082 (177) & 0.195 (23) & 0.112 (96) & \textbf{0.795 (1)} & 0.122 (63) & 0.112 (87) & 0.072 (216) & 0.101 (102) & 0.111 (73) & 0.057 (255) & 0.160 (33) \\
\texttt{HeartDisease} & 0.213 (172) & 0.244 (157) & 0.396 (35) & \textbf{0.416 (22)} & 0.109 (247) & 0.128 (233) & 0.293 (124) & 0.290 (126) & 0.150 (219) & 0.129 (231) & 0.370 (45) & 0.202 (186) & 0.149 (216) & 0.266 (140) & 0.247 (155) & 0.260 (139) & 0.325 (96) \\
\texttt{PageBlocks} & 0.116 (139) & 0.117 (122) & 0.313 (51) & 0.395 (28) & 0.055 (283) & 0.097 (177) & 0.083 (208) & \textbf{0.556 (6)} & 0.524 (6) & 0.116 (127) & 0.144 (92) & 0.340 (42) & 0.108 (147) & 0.452 (15) & 0.082 (214) & 0.196 (151) & 0.391 (98) \\
\texttt{PenDigits} & 0.492 (56) & 0.171 (138) & 0.568 (47) & 0.606 (35) & 0.046 (259) & 0.198 (121) & 0.208 (109) & 0.394 (76) & 0.899 (2) & 0.069 (192) & \textbf{0.932 (1)} & 0.514 (59) & 0.060 (210) & 0.808 (7) & 0.763 (10) & 0.106 (219) & 0.863 (4) \\
\texttt{Pima} & 0.237 (156) & 0.232 (146) & \textbf{0.483 (8)} & 0.464 (14) & 0.131 (232) & 0.079 (263) & 0.263 (125) & 0.369 (69) & 0.297 (114) & 0.095 (253) & 0.359 (79) & 0.433 (31) & 0.245 (137) & 0.410 (43) & 0.434 (29) & 0.426 (37) & 0.418 (39) \\
\texttt{Shuttle} & 0.057 (143) & 0.052 (164) & 0.067 (63) & 0.071 (53) & 0.044 (264) & 0.053 (164) & 0.046 (247) & 0.085 (32) & 0.055 (124) & 0.051 (194) & 0.064 (89) & 0.058 (107) & 0.048 (220) & 0.092 (22) & 0.044 (260) & 0.049 (197) & \textbf{0.095 (22)} \\
\texttt{SpamBase} & 0.078 (252) & 0.198 (135) & 0.425 (21) & 0.446 (11) & 0.191 (143) & 0.137 (189) & 0.334 (77) & 0.394 (45) & 0.404 (35) & 0.285 (84) & 0.374 (59) & 0.450 (10) & 0.298 (83) & 0.498 (5) & 0.345 (63) & 0.184 (147) & \textbf{0.521 (3)} \\
\texttt{Stamps} & 0.453 (86) & 0.266 (146) & 0.609 (16) & \textbf{0.620 (3)} & 0.220 (180) & 0.202 (194) & 0.367 (101) & 0.382 (100) & 0.343 (113) & 0.302 (125) & 0.505 (70) & 0.609 (26) & 0.244 (159) & 0.444 (85) & 0.503 (74) & 0.542 (47) & 0.506 (74) \\
\texttt{WBC} & 0.414 (88) & 0.183 (139) & 0.515 (37) & 0.528 (30) & 0.319 (85) & 0.250 (104) & 0.341 (82) & \textbf{0.621 (7)} & 0.368 (77) & 0.125 (220) & 0.381 (81) & 0.340 (82) & 0.162 (161) & 0.255 (94) & 0.314 (138) & 0.195 (125) & 0.375 (81) \\
\texttt{WDBC} & 0.088 (123) & 0.065 (196) & 0.143 (43) & 0.133 (48) & 0.039 (280) & 0.070 (166) & 0.060 (208) & 0.538 (2) & 0.076 (159) & \textbf{0.977 (1)} & 0.123 (61) & 0.097 (91) & 0.053 (234) & 0.126 (56) & 0.093 (135) & 0.049 (246) & 0.176 (48) \\
\texttt{WPBC} & 0.139 (140) & 0.077 (179) & 0.151 (48) & 0.101 (98) & 0.039 (293) & 0.125 (87) & 0.084 (147) & 0.405 (5) & 0.106 (95) & \textbf{0.946 (1)} & 0.151 (53) & 0.057 (255) & 0.054 (262) & 0.084 (149) & 0.063 (240) & 0.081 (160) & 0.084 (146) \\
\texttt{Waveform} & 0.086 (155) & 0.098 (127) & 0.427 (47) & 0.469 (24) & 0.035 (289) & 0.100 (139) & 0.100 (138) & 0.529 (10) & \textbf{0.631 (1)} & 0.050 (263) & 0.543 (5) & 0.418 (52) & 0.077 (176) & 0.539 (6) & 0.483 (20) & 0.076 (179) & 0.549 (3) \\
\texttt{Wilt} & 0.121 (167) & 0.112 (124) & 0.295 (4) & 0.279 (6) & 0.092 (212) & 0.087 (227) & 0.069 (277) & 0.199 (102) & \textbf{0.487 (1)} & 0.090 (188) & 0.086 (249) & 0.288 (5) & 0.153 (91) & 0.248 (28) & 0.305 (2) & 0.091 (212) & 0.283 (5) \\
\midrule
\multicolumn{18}{@{}l}{\textit{Dependency / Permute}} \\[1pt]
\texttt{Annthyroid} & 0.248 (52) & 0.128 (142) & 0.262 (47) & 0.321 (20) & 0.158 (104) & 0.103 (204) & 0.100 (219) & \textbf{0.590 (1)} & 0.190 (78) & 0.156 (115) & 0.227 (62) & 0.206 (70) & 0.129 (141) & 0.429 (2) & 0.111 (174) & 0.158 (160) & 0.426 (3) \\
\texttt{Cardiotocography} & 0.292 (90) & 0.086 (181) & 0.236 (51) & 0.264 (45) & 0.050 (265) & 0.056 (244) & 0.055 (239) & 0.413 (30) & 0.226 (53) & 0.150 (79) & \textbf{0.492 (20)} & 0.152 (71) & 0.069 (218) & 0.410 (30) & 0.062 (233) & 0.064 (223) & 0.372 (34) \\
\texttt{Glass} & 0.268 (57) & 0.111 (184) & 0.240 (49) & 0.212 (62) & 0.140 (143) & 0.158 (121) & 0.103 (201) & \textbf{0.546 (11)} & 0.163 (105) & 0.244 (49) & 0.439 (5) & 0.134 (147) & 0.111 (187) & 0.257 (38) & 0.162 (124) & 0.126 (186) & 0.264 (41) \\
\texttt{HeartDisease} & 0.203 (185) & 0.305 (118) & 0.456 (36) & 0.460 (25) & 0.142 (223) & 0.153 (216) & 0.299 (121) & \textbf{0.474 (49)} & 0.297 (127) & 0.198 (189) & 0.460 (41) & 0.188 (191) & 0.153 (210) & 0.302 (117) & 0.228 (164) & 0.242 (159) & 0.410 (52) \\
\texttt{PageBlocks} & 0.145 (195) & 0.224 (139) & 0.591 (37) & 0.654 (26) & 0.125 (217) & 0.234 (127) & 0.056 (276) & 0.810 (8) & 0.571 (38) & 0.265 (100) & 0.567 (39) & 0.596 (37) & 0.164 (191) & \textbf{0.907 (2)} & 0.095 (241) & 0.353 (171) & 0.200 (159) \\
\texttt{PenDigits} & 0.388 (167) & 0.518 (139) & 0.963 (13) & 0.888 (51) & 0.232 (182) & 0.430 (159) & 0.162 (196) & 0.949 (20) & 0.921 (33) & 0.427 (160) & \textbf{0.986 (1)} & 0.888 (51) & 0.157 (196) & 0.953 (19) & 0.756 (71) & 0.039 (283) & 0.943 (25) \\
\texttt{Pima} & 0.295 (186) & 0.411 (136) & 0.632 (18) & 0.529 (66) & 0.199 (227) & 0.208 (229) & 0.165 (242) & 0.378 (154) & 0.468 (99) & 0.210 (227) & 0.563 (39) & 0.507 (70) & 0.467 (108) & 0.725 (4) & 0.696 (8) & 0.638 (16) & \textbf{0.728 (3)} \\
\texttt{Shuttle} & 0.084 (176) & 0.077 (177) & 0.421 (42) & 0.397 (46) & 0.063 (205) & 0.058 (225) & 0.051 (236) & \textbf{0.914 (1)} & 0.268 (66) & 0.070 (194) & 0.612 (10) & 0.192 (72) & 0.091 (130) & 0.459 (34) & 0.046 (253) & 0.064 (209) & 0.062 (213) \\
\texttt{SpamBase} & 0.106 (225) & 0.234 (98) & 0.423 (19) & 0.413 (28) & 0.212 (113) & 0.187 (144) & 0.358 (56) & 0.388 (42) & \textbf{0.456 (5)} & 0.273 (82) & 0.337 (64) & 0.455 (5) & 0.304 (70) & 0.348 (63) & 0.196 (124) & 0.242 (122) & 0.147 (187) \\
\texttt{Stamps} & 0.251 (196) & 0.314 (168) & 0.662 (19) & 0.684 (13) & 0.383 (131) & 0.367 (145) & 0.239 (201) & \textbf{0.767 (4)} & 0.567 (82) & 0.532 (75) & 0.649 (26) & 0.650 (28) & 0.325 (162) & 0.641 (30) & 0.498 (102) & 0.624 (39) & 0.687 (5) \\
\texttt{WBC} & 0.123 (230) & 0.260 (160) & 0.554 (46) & 0.579 (37) & 0.387 (89) & 0.409 (79) & 0.288 (132) & 0.667 (14) & 0.519 (59) & 0.257 (159) & 0.402 (85) & 0.504 (58) & 0.235 (182) & 0.514 (61) & 0.362 (86) & 0.542 (37) & \textbf{0.693 (22)} \\
\texttt{WDBC} & 0.101 (147) & 0.075 (173) & 0.179 (46) & 0.174 (48) & 0.033 (295) & 0.062 (212) & 0.061 (209) & 0.540 (14) & 0.106 (88) & \textbf{0.950 (1)} & 0.210 (40) & 0.115 (80) & 0.060 (211) & 0.169 (48) & 0.072 (178) & 0.058 (221) & 0.259 (33) \\
\texttt{WPBC} & 0.208 (75) & 0.095 (178) & 0.204 (39) & 0.134 (89) & 0.040 (291) & 0.203 (41) & 0.090 (186) & 0.443 (12) & 0.120 (122) & \textbf{0.925 (1)} & 0.287 (26) & 0.093 (182) & 0.059 (255) & 0.099 (171) & 0.146 (169) & 0.078 (213) & 0.094 (180) \\
\texttt{Waveform} & 0.624 (95) & 0.206 (132) & 0.890 (46) & 0.910 (29) & 0.045 (285) & 0.105 (187) & 0.087 (206) & 0.863 (59) & 0.929 (4) & 0.101 (211) & 0.895 (37) & 0.891 (45) & 0.143 (168) & \textbf{0.930 (2)} & 0.403 (140) & 0.395 (125) & 0.145 (165) \\
\texttt{Wilt} & 0.236 (161) & 0.277 (130) & 0.446 (26) & 0.332 (76) & 0.459 (30) & 0.176 (212) & 0.050 (288) & 0.485 (17) & \textbf{0.816 (1)} & 0.202 (193) & 0.651 (7) & 0.433 (37) & 0.259 (165) & 0.709 (5) & 0.269 (125) & 0.471 (21) & 0.632 (6) \\
\midrule
\multicolumn{18}{@{}l}{\textit{Dep.-Strong / Noise}} \\[1pt]
\texttt{Annthyroid} & 0.134 (132) & 0.117 (160) & 0.200 (32) & 0.181 (54) & 0.087 (241) & 0.086 (255) & 0.095 (245) & 0.214 (22) & \textbf{0.222 (15)} & 0.137 (107) & 0.147 (92) & 0.173 (70) & 0.139 (96) & 0.219 (17) & 0.108 (193) & 0.110 (190) & 0.213 (21) \\
\texttt{Cardiotocography} & 0.110 (86) & 0.063 (160) & 0.115 (56) & 0.117 (55) & 0.037 (288) & 0.061 (179) & 0.049 (272) & \textbf{0.194 (21)} & 0.081 (89) & 0.058 (206) & 0.133 (48) & 0.085 (84) & 0.055 (220) & 0.192 (21) & 0.052 (243) & 0.049 (224) & 0.192 (21) \\
\texttt{Glass} & 0.093 (193) & 0.107 (175) & 0.219 (40) & 0.173 (57) & 0.115 (154) & 0.176 (60) & 0.120 (142) & 0.238 (33) & 0.168 (81) & \textbf{0.280 (27)} & 0.144 (111) & 0.124 (143) & 0.093 (194) & 0.138 (114) & 0.229 (33) & 0.092 (196) & 0.236 (32) \\
\texttt{HeartDisease} & 0.481 (59) & 0.399 (124) & 0.464 (70) & 0.488 (53) & 0.220 (218) & 0.205 (224) & 0.496 (46) & 0.545 (33) & 0.222 (209) & 0.149 (249) & 0.481 (58) & 0.279 (191) & 0.344 (161) & 0.519 (28) & 0.437 (94) & 0.432 (92) & \textbf{0.554 (5)} \\
\texttt{PageBlocks} & 0.211 (168) & 0.122 (115) & 0.307 (52) & 0.409 (24) & 0.063 (269) & 0.095 (184) & 0.082 (212) & 0.487 (11) & \textbf{0.535 (6)} & 0.122 (115) & 0.148 (92) & 0.349 (43) & 0.113 (130) & 0.453 (15) & 0.067 (236) & 0.064 (247) & 0.225 (185) \\
\texttt{PenDigits} & 0.110 (174) & 0.178 (147) & 0.569 (51) & 0.618 (33) & 0.056 (223) & 0.195 (130) & 0.190 (136) & 0.373 (84) & 0.897 (2) & 0.066 (203) & \textbf{0.924 (1)} & 0.513 (62) & 0.062 (207) & 0.815 (7) & 0.769 (10) & 0.075 (241) & 0.863 (4) \\
\texttt{Pima} & 0.182 (192) & 0.252 (121) & 0.420 (22) & \textbf{0.463 (2)} & 0.149 (208) & 0.154 (203) & 0.256 (119) & 0.381 (41) & 0.236 (133) & 0.132 (222) & 0.322 (82) & 0.428 (13) & 0.209 (158) & 0.354 (68) & 0.405 (34) & 0.360 (58) & 0.354 (68) \\
\texttt{Shuttle} & 0.093 (62) & 0.059 (165) & 0.106 (49) & 0.121 (37) & 0.049 (235) & 0.058 (186) & 0.061 (150) & 0.129 (31) & 0.071 (105) & 0.059 (180) & 0.090 (66) & 0.079 (85) & 0.059 (169) & \textbf{0.135 (29)} & 0.051 (233) & 0.057 (188) & 0.131 (30) \\
\texttt{SpamBase} & 0.175 (139) & 0.157 (159) & 0.290 (24) & 0.319 (12) & 0.124 (185) & 0.109 (207) & 0.301 (12) & 0.235 (77) & 0.285 (30) & 0.195 (114) & 0.329 (11) & 0.344 (9) & 0.212 (93) & \textbf{0.379 (5)} & 0.247 (77) & 0.167 (145) & 0.124 (186) \\
\texttt{Stamps} & 0.320 (117) & 0.217 (150) & 0.569 (21) & 0.591 (11) & 0.179 (181) & 0.164 (190) & 0.318 (96) & \textbf{0.635 (5)} & 0.285 (125) & 0.115 (232) & 0.481 (74) & 0.549 (37) & 0.213 (154) & 0.413 (83) & 0.508 (62) & 0.556 (27) & 0.504 (64) \\
\texttt{WBC} & 0.202 (110) & 0.162 (154) & 0.247 (25) & 0.258 (22) & 0.117 (228) & 0.197 (101) & 0.214 (66) & \textbf{0.267 (19)} & 0.135 (203) & 0.122 (243) & 0.243 (24) & 0.247 (24) & 0.148 (181) & 0.169 (140) & 0.177 (123) & 0.168 (121) & 0.179 (117) \\
\texttt{WDBC} & 0.068 (172) & 0.068 (175) & 0.138 (40) & 0.114 (55) & 0.045 (260) & 0.077 (152) & 0.061 (205) & 0.421 (6) & 0.063 (195) & \textbf{0.971 (1)} & 0.118 (58) & 0.087 (103) & 0.055 (225) & 0.120 (50) & 0.043 (274) & 0.051 (242) & 0.159 (40) \\
\texttt{WPBC} & 0.078 (165) & 0.071 (191) & 0.155 (38) & 0.094 (97) & 0.041 (288) & 0.140 (91) & 0.075 (165) & 0.355 (2) & 0.085 (138) & \textbf{0.980 (1)} & 0.144 (47) & 0.073 (178) & 0.051 (271) & 0.081 (138) & 0.113 (81) & 0.072 (184) & 0.067 (214) \\
\texttt{Waveform} & 0.092 (152) & 0.101 (137) & 0.452 (50) & 0.491 (24) & 0.034 (289) & 0.077 (181) & 0.119 (123) & 0.539 (13) & \textbf{0.670 (1)} & 0.059 (218) & 0.527 (17) & 0.440 (52) & 0.078 (175) & 0.557 (7) & 0.554 (11) & 0.083 (174) & 0.079 (172) \\
\texttt{Wilt} & 0.134 (148) & 0.135 (112) & 0.304 (7) & 0.307 (7) & 0.086 (263) & 0.100 (215) & 0.075 (277) & 0.207 (71) & \textbf{0.507 (1)} & 0.111 (181) & 0.092 (256) & 0.296 (9) & 0.169 (92) & 0.283 (14) & 0.362 (1) & 0.162 (160) & 0.329 (5) \\
\midrule
\multicolumn{18}{@{}l}{\textit{Dep.-Strong / Permute}} \\[1pt]
\texttt{Annthyroid} & 0.122 (181) & 0.138 (157) & 0.280 (49) & 0.322 (24) & 0.161 (105) & 0.119 (187) & 0.099 (237) & \textbf{0.588 (1)} & 0.213 (75) & 0.179 (95) & 0.221 (70) & 0.197 (81) & 0.142 (146) & 0.433 (2) & 0.101 (231) & 0.153 (182) & 0.422 (4) \\
\texttt{Cardiotocography} & 0.138 (118) & 0.088 (151) & 0.219 (53) & 0.257 (46) & 0.040 (289) & 0.088 (195) & 0.050 (266) & 0.417 (28) & 0.188 (61) & 0.117 (104) & \textbf{0.444 (26)} & 0.152 (71) & 0.067 (222) & 0.411 (28) & 0.050 (261) & 0.199 (134) & 0.411 (28) \\
\texttt{Glass} & 0.171 (139) & 0.103 (195) & 0.275 (53) & 0.279 (50) & 0.162 (125) & 0.220 (95) & 0.077 (231) & \textbf{0.581 (3)} & 0.199 (81) & 0.143 (138) & 0.459 (9) & 0.143 (126) & 0.117 (175) & 0.320 (33) & 0.156 (149) & 0.157 (166) & 0.338 (36) \\
\texttt{HeartDisease} & 0.244 (204) & 0.441 (96) & 0.617 (17) & 0.465 (86) & 0.242 (200) & 0.098 (273) & 0.467 (84) & 0.555 (51) & 0.357 (136) & 0.294 (175) & \textbf{0.628 (15)} & 0.349 (143) & 0.326 (150) & 0.470 (77) & 0.265 (180) & 0.299 (167) & 0.531 (37) \\
\texttt{PageBlocks} & 0.222 (164) & 0.222 (146) & 0.581 (37) & 0.686 (22) & 0.148 (193) & 0.137 (197) & 0.051 (281) & 0.821 (7) & 0.597 (35) & 0.179 (171) & 0.580 (36) & 0.586 (36) & 0.161 (187) & \textbf{0.911 (2)} & 0.150 (202) & 0.186 (165) & 0.324 (179) \\
\texttt{PenDigits} & 0.767 (79) & 0.519 (147) & 0.962 (13) & 0.882 (53) & 0.227 (184) & 0.445 (159) & 0.155 (197) & 0.951 (18) & 0.926 (31) & 0.290 (179) & \textbf{0.987 (2)} & 0.883 (49) & 0.151 (197) & 0.950 (19) & 0.938 (27) & 0.040 (285) & 0.946 (22) \\
\texttt{Pima} & 0.149 (243) & 0.405 (148) & 0.599 (30) & 0.585 (38) & 0.262 (206) & 0.311 (184) & 0.215 (225) & 0.439 (106) & 0.496 (97) & 0.253 (208) & 0.526 (73) & 0.535 (63) & 0.493 (100) & \textbf{0.687 (1)} & 0.618 (20) & 0.386 (133) & 0.666 (4) \\
\texttt{Shuttle} & 0.211 (122) & 0.080 (192) & 0.438 (40) & 0.411 (46) & 0.075 (198) & 0.047 (256) & 0.064 (216) & \textbf{0.935 (1)} & 0.316 (63) & 0.218 (123) & 0.610 (15) & 0.227 (72) & 0.104 (131) & 0.451 (38) & 0.047 (249) & 0.055 (239) & 0.466 (37) \\
\texttt{SpamBase} & 0.205 (125) & 0.191 (110) & 0.280 (31) & 0.312 (19) & 0.130 (177) & 0.105 (205) & 0.312 (19) & 0.265 (48) & 0.320 (19) & 0.190 (112) & 0.251 (60) & \textbf{0.357 (6)} & 0.211 (94) & 0.327 (18) & 0.146 (151) & 0.266 (65) & 0.140 (164) \\
\texttt{Stamps} & 0.157 (236) & 0.317 (175) & 0.592 (60) & 0.667 (18) & 0.315 (177) & 0.222 (214) & 0.244 (205) & \textbf{0.733 (32)} & 0.644 (35) & 0.497 (96) & 0.620 (44) & 0.639 (27) & 0.312 (181) & 0.647 (21) & 0.455 (110) & 0.674 (15) & 0.696 (10) \\
\texttt{WBC} & 0.128 (191) & 0.188 (151) & 0.213 (79) & 0.276 (17) & 0.123 (222) & 0.139 (206) & 0.087 (257) & \textbf{0.321 (123)} & 0.239 (65) & 0.079 (265) & 0.256 (22) & 0.257 (23) & 0.185 (159) & 0.207 (106) & 0.151 (193) & 0.214 (75) & 0.272 (18) \\
\texttt{WDBC} & 0.093 (169) & 0.069 (185) & 0.164 (45) & 0.152 (51) & 0.043 (268) & 0.069 (186) & 0.057 (224) & 0.497 (14) & 0.082 (143) & \textbf{0.969 (1)} & 0.159 (47) & 0.100 (97) & 0.059 (221) & 0.154 (49) & 0.049 (249) & 0.055 (231) & 0.222 (41) \\
\texttt{WPBC} & 0.083 (171) & 0.082 (184) & 0.163 (43) & 0.090 (151) & 0.040 (288) & 0.111 (107) & 0.077 (200) & 0.343 (12) & 0.097 (134) & \textbf{0.983 (1)} & 0.164 (44) & 0.068 (227) & 0.057 (255) & 0.080 (193) & 0.106 (161) & 0.078 (196) & 0.182 (39) \\
\texttt{Waveform} & 0.107 (200) & 0.209 (133) & 0.851 (56) & 0.885 (34) & 0.050 (276) & 0.125 (187) & 0.089 (219) & 0.896 (20) & \textbf{0.916 (11)} & 0.105 (220) & 0.904 (16) & 0.872 (44) & 0.136 (172) & 0.912 (6) & 0.348 (167) & 0.426 (101) & 0.148 (165) \\
\texttt{Wilt} & 0.227 (149) & 0.280 (118) & 0.443 (32) & 0.343 (66) & 0.451 (68) & 0.147 (226) & 0.050 (285) & 0.395 (60) & \textbf{0.856 (1)} & 0.212 (183) & 0.710 (6) & 0.464 (21) & 0.273 (139) & 0.749 (5) & 0.286 (95) & 0.507 (46) & 0.650 (6) \\
\midrule
\multicolumn{18}{@{}l}{\textit{Tight-Extr.-Cluster / Noise}} \\[1pt]
\texttt{Annthyroid} & 0.332 (193) & 0.981 (63) & 0.030 (277) & 0.038 (241) & 0.656 (144) & 0.046 (212) & 0.998 (51) & 0.036 (250) & 0.038 (244) & 0.216 (205) & 0.034 (263) & 0.029 (279) & \textbf{1.000 (1)} & 0.030 (277) & 0.998 (19) & 0.673 (110) & 0.996 (47) \\
\texttt{Cardiotocography} & 0.431 (162) & 0.991 (20) & 0.026 (288) & 0.028 (269) & 0.609 (129) & 0.123 (201) & \textbf{1.000 (1)} & 0.037 (228) & 0.372 (163) & 0.871 (62) & 0.026 (285) & 0.622 (126) & 0.951 (37) & 0.028 (262) & 0.717 (84) & 0.996 (7) & 0.994 (9) \\
\texttt{Glass} & 0.188 (215) & 0.736 (80) & 0.354 (181) & \textbf{1.000 (1)} & 0.548 (139) & 0.103 (241) & \textbf{1.000 (8)} & 0.045 (276) & 0.290 (195) & 0.047 (274) & 0.879 (38) & 0.453 (148) & 0.293 (186) & 0.919 (31) & 0.742 (78) & 0.926 (22) & 0.749 (90) \\
\texttt{HeartDisease} & 0.660 (152) & \textbf{1.000 (1)} & 0.061 (269) & \textbf{1.000 (1)} & 0.837 (123) & 0.092 (254) & \textbf{1.000 (1)} & 0.042 (288) & 0.376 (218) & 0.169 (238) & 0.361 (194) & 0.874 (151) & \textbf{1.000 (1)} & \textbf{1.000 (1)} & \textbf{1.000 (1)} & \textbf{1.000 (1)} & \textbf{1.000 (1)} \\
\texttt{PageBlocks} & 0.064 (231) & 0.655 (38) & 0.054 (236) & 0.033 (263) & 0.497 (114) & 0.067 (230) & \textbf{0.873 (1)} & 0.047 (242) & 0.053 (237) & 0.363 (143) & 0.117 (197) & 0.044 (244) & 0.839 (1) & 0.031 (266) & 0.527 (69) & 0.621 (63) & 0.031 (269) \\
\texttt{PenDigits} & 0.296 (169) & 0.803 (59) & 0.028 (276) & 0.045 (213) & 0.871 (40) & 0.047 (207) & \textbf{1.000 (4)} & 0.045 (212) & 0.036 (233) & 0.259 (157) & 0.025 (283) & 0.036 (234) & \textbf{1.000 (1)} & 0.045 (210) & 0.681 (73) & 0.960 (17) & 0.052 (194) \\
\texttt{Pima} & 0.961 (46) & 0.999 (17) & 0.038 (265) & 0.962 (45) & 0.784 (150) & 0.370 (195) & \textbf{1.000 (1)} & 0.037 (266) & 0.928 (119) & 0.265 (209) & 0.653 (134) & \textbf{1.000 (1)} & 0.990 (85) & 0.059 (248) & \textbf{1.000 (1)} & 0.996 (29) & 0.040 (267) \\
\texttt{Shuttle} & 0.712 (117) & \textbf{0.993 (20)} & 0.087 (250) & 0.035 (276) & 0.533 (158) & 0.534 (153) & 0.930 (54) & 0.053 (257) & 0.543 (154) & 0.648 (127) & 0.033 (279) & 0.851 (81) & 0.901 (74) & 0.040 (264) & 0.500 (159) & 0.991 (12) & 0.176 (232) \\
\texttt{SpamBase} & 0.119 (196) & 0.756 (86) & 0.025 (294) & 0.026 (253) & 0.475 (123) & 0.062 (195) & 0.240 (153) & 0.027 (231) & 0.025 (291) & 0.096 (189) & 0.034 (203) & 0.116 (186) & \textbf{1.000 (7)} & 0.025 (269) & 0.792 (58) & 0.887 (40) & 0.987 (41) \\
\texttt{Stamps} & 0.683 (104) & 0.921 (39) & 0.115 (232) & 0.798 (76) & 0.522 (166) & 0.107 (237) & \textbf{0.976 (19)} & 0.040 (266) & 0.454 (182) & 0.345 (192) & 0.779 (80) & 0.802 (75) & 0.481 (172) & 0.721 (94) & 0.764 (85) & 0.563 (137) & 0.731 (92) \\
\texttt{WBC} & 0.956 (84) & 0.987 (104) & 0.052 (275) & \textbf{1.000 (1)} & 0.994 (101) & 0.088 (262) & \textbf{1.000 (98)} & 0.031 (287) & 0.722 (164) & 0.361 (223) & 0.355 (220) & 0.933 (52) & 0.906 (130) & \textbf{1.000 (1)} & 0.905 (55) & 0.980 (43) & \textbf{1.000 (1)} \\
\texttt{WDBC} & 0.197 (190) & 0.644 (54) & 0.025 (285) & 0.759 (17) & 0.322 (164) & 0.189 (194) & 0.701 (23) & 0.030 (248) & 0.630 (65) & 0.033 (245) & 0.240 (180) & 0.676 (40) & 0.120 (210) & \textbf{0.850 (5)} & 0.768 (27) & 0.404 (130) & 0.794 (19) \\
\texttt{WPBC} & 0.246 (190) & 0.804 (58) & 0.036 (267) & \textbf{1.000 (1)} & 0.525 (124) & 0.112 (218) & 0.815 (48) & 0.036 (267) & 0.467 (138) & 0.035 (271) & 0.664 (95) & 0.427 (153) & 0.050 (237) & \textbf{1.000 (1)} & \textbf{1.000 (1)} & 0.662 (87) & 0.861 (37) \\
\texttt{Waveform} & 0.295 (179) & 0.995 (39) & 0.024 (282) & 0.035 (200) & 0.809 (88) & 0.051 (178) & \textbf{1.000 (1)} & 0.025 (250) & 0.046 (182) & 0.490 (121) & 0.025 (247) & 0.132 (169) & \textbf{1.000 (1)} & 0.033 (203) & \textbf{1.000 (1)} & 0.464 (113) & 0.345 (155) \\
\texttt{Wilt} & 0.853 (38) & 0.835 (28) & 0.071 (239) & 0.033 (274) & 0.645 (103) & 0.048 (247) & 0.967 (9) & 0.046 (249) & 0.064 (241) & 0.368 (156) & 0.369 (167) & 0.045 (251) & \textbf{1.000 (1)} & 0.032 (274) & 0.254 (190) & 0.748 (61) & 0.034 (269) \\
\midrule
\multicolumn{18}{@{}l}{\textit{Tight-Extr.-Cluster / Permute}} \\[1pt]
\texttt{Annthyroid} & 0.346 (202) & 0.654 (108) & 0.024 (258) & 0.032 (230) & 0.349 (172) & 0.221 (179) & 0.989 (46) & 0.025 (253) & 0.045 (228) & 0.145 (206) & 0.132 (187) & 0.030 (243) & 0.024 (279) & 0.031 (233) & 0.948 (29) & 0.606 (108) & \textbf{1.000 (1)} \\
\texttt{Cardiotocography} & 0.608 (134) & 0.695 (111) & 0.024 (268) & 0.027 (252) & 0.555 (133) & 0.271 (195) & \textbf{1.000 (1)} & 0.024 (271) & 0.154 (203) & 0.364 (168) & 0.373 (139) & 0.519 (143) & 0.024 (271) & 0.028 (240) & 0.954 (29) & 0.442 (162) & 0.988 (31) \\
\texttt{Glass} & 0.508 (143) & 0.774 (77) & 0.317 (209) & 0.679 (90) & 0.502 (155) & 0.267 (222) & \textbf{1.000 (35)} & 0.056 (264) & 0.540 (134) & 0.045 (270) & 0.826 (62) & 0.545 (135) & 0.026 (288) & \textbf{1.000 (1)} & \textbf{1.000 (1)} & \textbf{1.000 (1)} & \textbf{1.000 (1)} \\
\texttt{HeartDisease} & 0.844 (77) & \textbf{1.000 (53)} & 0.035 (283) & \textbf{1.000 (1)} & 0.537 (230) & 0.316 (240) & \textbf{1.000 (1)} & 0.035 (283) & 0.777 (146) & 0.143 (249) & 0.055 (261) & 0.496 (214) & 0.035 (283) & \textbf{1.000 (1)} & \textbf{1.000 (1)} & \textbf{1.000 (1)} & \textbf{1.000 (1)} \\
\texttt{PageBlocks} & 0.545 (118) & 0.393 (107) & 0.024 (259) & 0.034 (232) & 0.453 (88) & 0.272 (177) & \textbf{0.958 (18)} & 0.054 (216) & 0.582 (78) & 0.266 (176) & 0.118 (194) & 0.055 (199) & 0.024 (278) & 0.033 (231) & 0.781 (55) & 0.624 (74) & 0.651 (72) \\
\texttt{PenDigits} & 0.719 (56) & 0.220 (125) & 0.024 (282) & 0.037 (234) & 0.498 (92) & 0.165 (159) & \textbf{1.000 (1)} & 0.024 (274) & 0.035 (229) & 0.363 (150) & 0.026 (250) & 0.033 (246) & 0.024 (284) & 0.037 (230) & \textbf{1.000 (1)} & 0.999 (21) & 0.040 (224) \\
\texttt{Pima} & 0.642 (134) & 0.982 (80) & 0.025 (280) & 0.962 (39) & 0.580 (196) & 0.620 (137) & \textbf{1.000 (1)} & 0.025 (280) & 0.898 (91) & 0.145 (245) & 0.055 (247) & \textbf{1.000 (1)} & 0.025 (280) & 0.189 (222) & \textbf{1.000 (1)} & 0.999 (23) & 0.991 (52) \\
\texttt{Shuttle} & 0.264 (215) & 0.911 (98) & 0.025 (270) & 0.049 (247) & 0.573 (205) & 0.868 (104) & 0.930 (89) & 0.041 (242) & 0.630 (184) & 0.074 (228) & 0.495 (199) & 0.858 (107) & 0.025 (275) & 0.259 (219) & 0.595 (201) & 0.925 (67) & \textbf{1.000 (1)} \\
\texttt{SpamBase} & 0.759 (102) & 0.760 (95) & 0.024 (270) & 0.025 (253) & 0.513 (176) & 0.292 (186) & 0.267 (186) & 0.024 (277) & 0.026 (256) & 0.038 (226) & 0.032 (212) & 0.123 (196) & 0.024 (260) & 0.025 (250) & 0.796 (62) & 0.819 (83) & \textbf{0.822 (81)} \\
\texttt{Stamps} & 0.540 (127) & 0.739 (74) & 0.025 (280) & 0.775 (62) & 0.502 (144) & 0.273 (213) & 0.841 (46) & 0.025 (279) & 0.668 (87) & 0.089 (242) & 0.027 (272) & 0.782 (63) & 0.025 (280) & 0.741 (70) & \textbf{0.894 (31)} & 0.828 (47) & 0.836 (50) \\
\texttt{WBC} & 0.461 (194) & 0.760 (147) & 0.025 (285) & \textbf{1.000 (1)} & 0.719 (153) & 0.251 (226) & \textbf{1.000 (72)} & 0.025 (285) & 0.893 (101) & 0.123 (252) & 0.321 (220) & \textbf{1.000 (48)} & 0.025 (285) & \textbf{1.000 (1)} & 0.817 (128) & \textbf{1.000 (1)} & \textbf{1.000 (1)} \\
\texttt{WDBC} & 0.410 (137) & 0.496 (102) & 0.024 (283) & 0.815 (26) & 0.461 (116) & 0.476 (130) & 0.736 (42) & 0.024 (283) & 0.717 (50) & 0.025 (280) & 0.032 (263) & 0.778 (39) & 0.024 (283) & \textbf{0.853 (16)} & \textbf{0.853 (16)} & 0.744 (49) & \textbf{0.853 (16)} \\
\texttt{WPBC} & 0.660 (87) & 0.690 (68) & 0.035 (280) & \textbf{1.000 (1)} & 0.537 (114) & 0.232 (193) & 0.798 (44) & 0.035 (281) & 0.704 (96) & 0.035 (281) & 0.077 (240) & 0.373 (156) & 0.035 (281) & \textbf{1.000 (1)} & 0.939 (14) & 0.604 (106) & 0.889 (29) \\
\texttt{Waveform} & 0.650 (118) & 0.938 (99) & 0.024 (264) & 0.034 (233) & 0.695 (148) & 0.221 (166) & \textbf{1.000 (2)} & 0.024 (271) & 0.025 (255) & 0.036 (225) & 0.024 (257) & 0.484 (158) & 0.024 (271) & 0.034 (225) & \textbf{1.000 (1)} & 0.945 (72) & 0.990 (71) \\
\texttt{Wilt} & 0.590 (115) & 0.415 (136) & 0.024 (261) & 0.035 (244) & 0.491 (100) & 0.226 (192) & 0.978 (40) & 0.030 (251) & 0.085 (201) & 0.279 (161) & 0.026 (252) & 0.059 (210) & 0.024 (275) & 0.036 (235) & \textbf{0.984 (26)} & 0.809 (58) & 0.935 (58) \\
\midrule
\textbf{Mean} & 0.305 (146) & 0.383 (125) & 0.277 (114) & 0.406 (75) & 0.285 (187) & 0.177 (184) & 0.403 (133) & 0.328 (111) & 0.379 (107) & 0.269 (162) & 0.348 (102) & 0.381 (98) & 0.237 (173) & 0.413 (81) & 0.459 (107) & 0.412 (123) & \textbf{0.485 (69)} \\
\bottomrule
\end{tabular}
}
\end{table*}

\clearpage

\end{document}